\documentclass[sn-mathphys,Numbered]{sn-jnl}% Math and Physical Sciences Reference Style
\usepackage{graphicx}%
\graphicspath{ {./images/} }
\usepackage{multirow}%
\usepackage{amsmath,amssymb,amsfonts}%
\usepackage{amsthm}%
\usepackage{mathrsfs}%
\usepackage[title]{appendix}%
\usepackage{xcolor}%
\usepackage{textcomp}%
\usepackage{manyfoot}%
\usepackage{booktabs}%
\usepackage{algorithm}%
\usepackage{algorithmicx}%
\usepackage{algpseudocode}%
\usepackage{listings}%

\usepackage{cleveref}
\usepackage{subcaption}
\usepackage{multirow}
\usepackage{braket}
\usepackage{placeins}

\begin{document}

\title[]{Conformalised data synthesis}


\author*[1]{\fnm{Julia A.} \sur{Meister}}\email{J.Meister@brighton.ac.uk}

\author[2]{\fnm{Khuong An} \sur{Nguyen}}\email{Khuong.Nguyen@rhul.ac.uk}

\affil[1]{\orgdiv{Computing and Maths Division}, \orgname{University of Brighton}, \orgaddress{\street{Lewes Road}, \city{Brighton}, \postcode{BN2 4AT}, \state{East Sussex}, \country{UK}}}

\affil[2]{\orgdiv{Department of Computer Science}, \orgname{Royal Holloway, University of London}, \orgaddress{\street{Egham Hill}, \city{Egham}, \postcode{TW20 0EX}, \state{Surrey}, \country{UK}}}

\abstract{
With the proliferation of increasingly complicated Deep Learning architectures, data synthesis is a highly promising technique to address the demand of data-hungry models. However, reliably assessing the quality of a `synthesiser' model's output is an open research question with significant associated risks for high-stake domains. To address this challenge, we propose a unique synthesis algorithm that generates data from high-confidence feature space regions based on the Conformal Prediction framework. We support our proposed algorithm with a comprehensive exploration of the core parameter's influence, an in-depth discussion of practical advice, and an extensive empirical evaluation of five benchmark datasets. To show our approach's versatility on ubiquitous real-world challenges, the datasets were carefully selected for their variety of difficult characteristics: low sample count, class imbalance, and non-separability. In all trials, training sets extended with our confident synthesised data performed at least as well as the original set and frequently significantly improved Deep Learning performance by up to 61 percentage points F$_1$-score.
}

\keywords{Conformal Prediction, uncertainty quantification, statistical confidence, synthetic data, data generation}
\pacs[MSC Classification]{68T37}

\maketitle

\newpage
\section{Introduction}\label{sec:intro}
    Specialised and data-hungry Deep Learning implementations are increasingly faced with small and unrepresentative datasets, a well-established challenge in the literature~\cite{Brigato2021,Moreno-Barea2020,Sarker2021}.
    Unfortunately, increasing the sample size by collecting more data is challenging in many real-world applications. Prohibitors include high data collection costs, low data availability, and the lack of expertise in ground-truth labelling.

    Data synthesis is a highly sophisticated approach to combat small datasets. Improving on techniques that modify and remix existing samples (e.g., data resampling, randomisation, and augmentation), data synthesis generates entirely new and unseen examples based on the original data~\cite{Zhuang2019}. Similarly to classification, synthesis relies on accurately modelling the data's distribution to extrapolate plausible new feature vectors~\cite{Liu2022dataSynthesis}. In Deep Learning, these synthesised samples may be included in the training set to improve model generalisation and, consequently, prediction performance.

    Generative Adversarial Networks (GANs) are a state-of-the-art synthesis technique, leveraging a zero-sum game between a Deep Learning generator and discriminator to synthesise realistic samples~\cite{Aggarwal2021}. However, a major challenge and open research question shared among most existing data synthesis techniques is how to quantify the produced synthetic data's quality. Evaluating a generator model's distributional fit is fundamentally difficult because there is no inherent quality metric for previously unseen data~\cite{Grnarova2019}.

\begin{figure}[!h]
        \centering
        \begin{subfigure}[b]{0.325\textwidth}
            \centering
            \includegraphics[width=\textwidth]{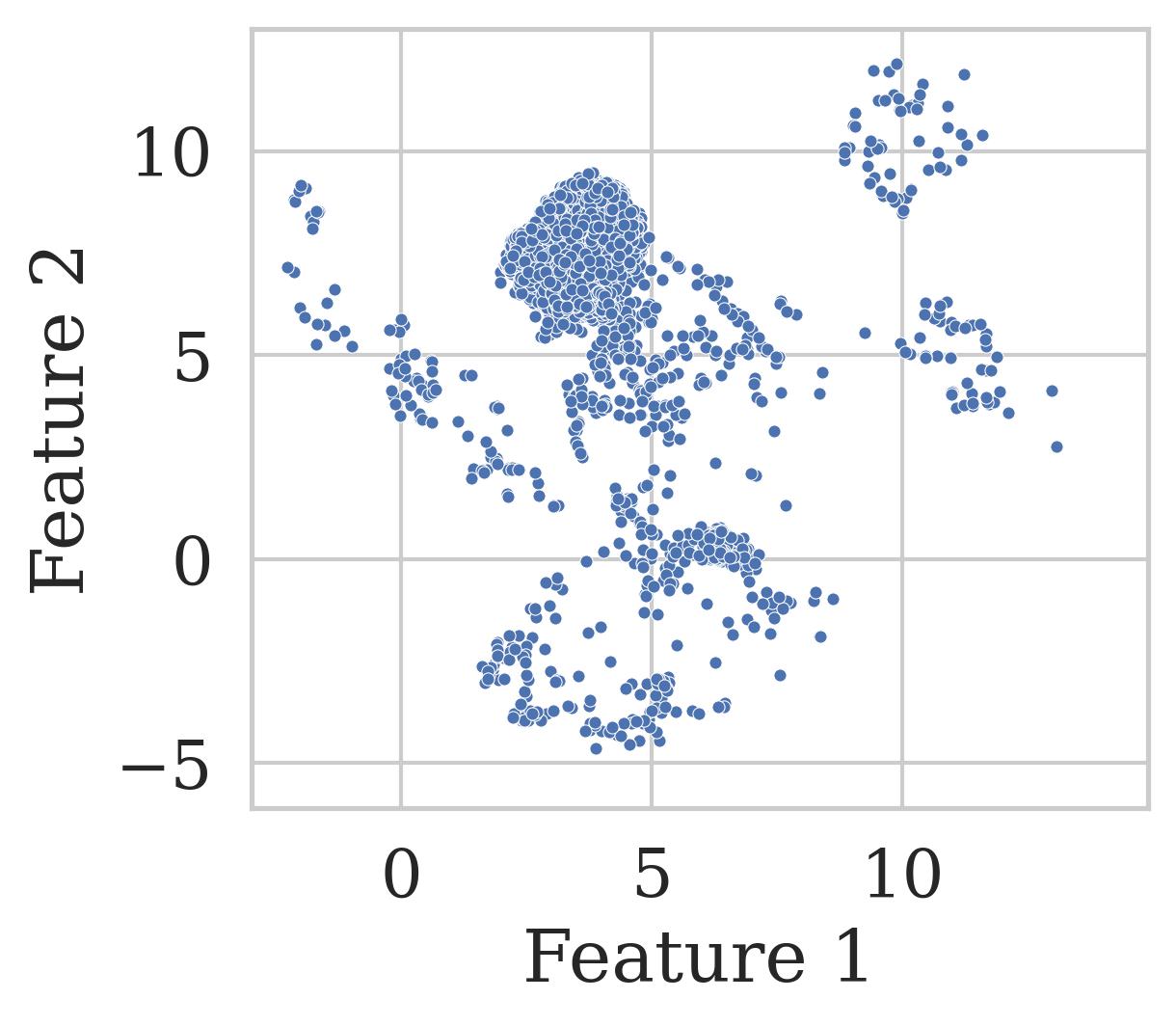}
            \caption{Original samples.}\label{fig:introOriginal}
        \end{subfigure}
        \hfill
        \begin{subfigure}[b]{0.325\textwidth}
            \centering
            \includegraphics[width=\textwidth]{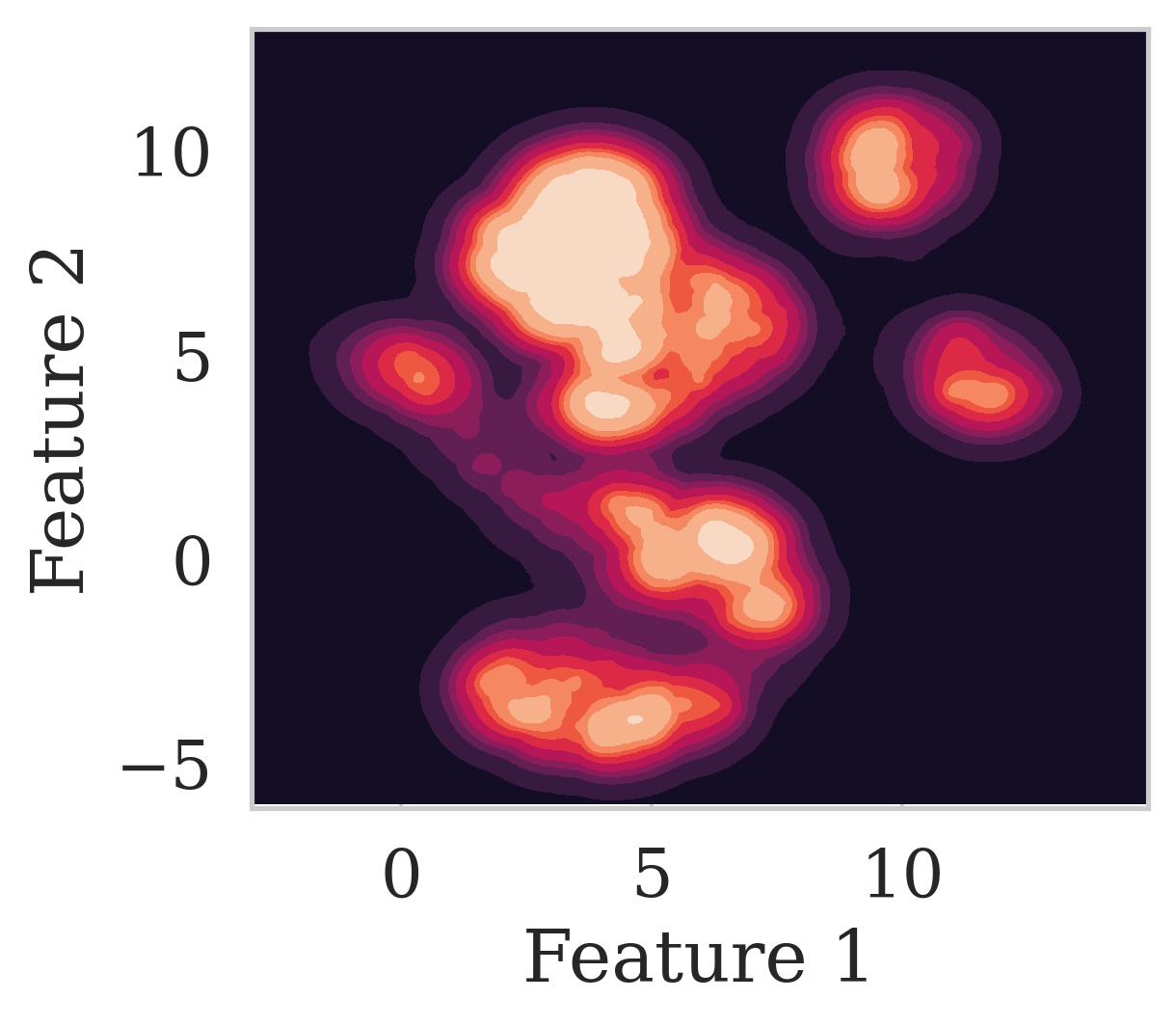}
            \caption{Confidence regions.}\label{fig:introConfidence}
        \end{subfigure}
        \hfill
        \begin{subfigure}[b]{0.325\textwidth}
            \centering
            \includegraphics[width=\textwidth]{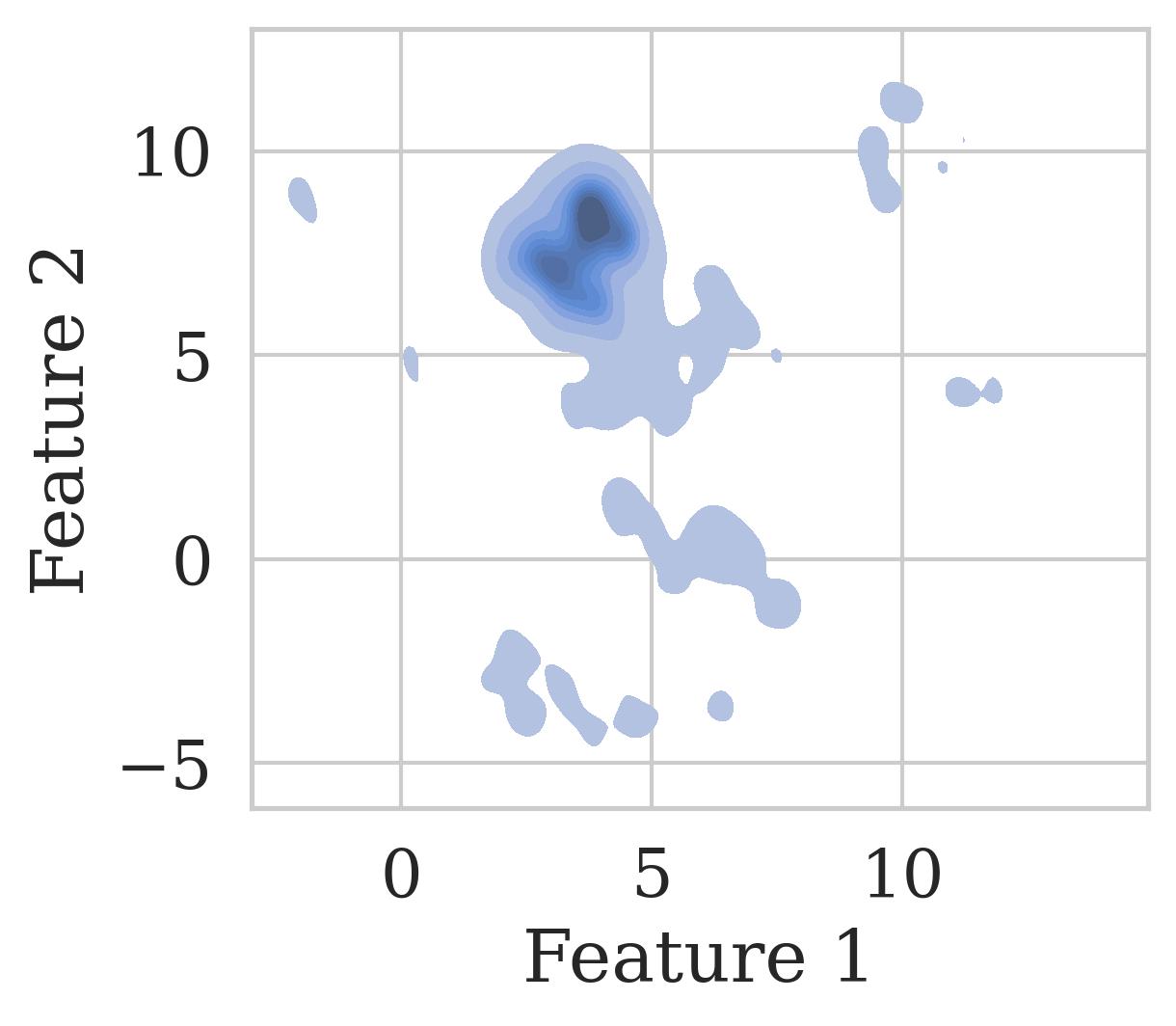}
            \caption{Density estimation.}\label{fig:introDensity}
        \end{subfigure}
        \caption{Visualisation of two techniques to identify feature space regions from which data is synthesised (\Cref{fig:introConfidence,fig:introDensity}). Compared to traditional density estimation of the samples, our proposed algorithm determines conformal high-confidence regions based on a user-selected threshold. The lower the threshold, the wider the confidence regions become from which samples are synthesised.}\label{fig:confidenceVsDensity}
    \end{figure}

    To address this challenge, we have designed a unique confident data synthesis algorithm based on a novel extension of the Conformal Prediction framework~\cite{shafer2008tutorial}.
    Inspired by Cherubin et al.'s innovative conformal clustering paper~\cite{Cherubin2015}, we rely on the confidence of feature space regions to guide our data generation. \Cref{fig:confidenceVsDensity} illustrates the effect of our confidence estimation approach on the feature space compared to traditional density estimation.
    \Cref{sec:contributions} details our contributions in more depth and describes the article's structure.  % chktex 38

    \subsection{Contributions}\label{sec:contributions}
        In this article, we propose conformal data synthesis to significantly improve Deep Learning prediction performance on small and imbalanced datasets. Our algorithm builds on Conformal Prediction, a foundational confidence framework founded on hypothesis testing. To the best of our knowledge, this is the first extension of Conformal Prediction to data generation.

        Inspired by the innovative conformal clustering work presented in~\cite{Cherubin2015}, our algorithm relies on the confidence of feature space regions to synthesise new data points. Our three key contributions are:
        \begin{itemize}
            \item We incorporate ground-truth labels, making the confidence modelling process \emph{supervised},
            \item We identify \emph{label-conditional} confidence regions in the feature space,
            \item And, most decisively, the confidence regions are an intermediate stage from which we \emph{synthesise new data points}.
        \end{itemize}

        In the following sections, we will discuss the context and motivation of our data synthesis solution (\Cref{sec:relatedWork}); Present its theoretical foundation and inspiration (\Cref{sec:cpBackground}); Introduce our proposed synthesis algorithm (\Cref{sec:proposal}); Systematically evaluate the empirical advantages on real-world datasets of varying difficulty (\Cref{sec:results}); And discuss practical advice and future directions of the proposed algorithm (\Cref{sec:discussion}).

\section{Related work in data synthesis}\label{sec:relatedWork}
    A model's performance is strongly dependent on its underlying data. Consequently, state-of-the-art Deep Learning models are frequently trained on vast datasets to optimise prediction performances. Although a large sample count is not necessarily required for high performance, it increases the likelihood of a representative collection of samples~\cite{Althnian2021}.
    Unfortunately, real-world applications frequently suffer under small and unrepresentative datasets due to the shortage of high-quality annotated samples~\cite{Alauthman2023}. However, artificially extended datasets have been shown to improve results. Data synthesis is the most sophisticated approach to extend datasets with entirely new data. The core concept is to generate previously unseen data points from the existing data automatically. The most popular use cases are to improve model inference with synthetic samples~\cite{Muramatsu2020,Salazar2021,Koshino2021} and to replace a dataset with synthetic samples~\cite{Li2020,Yoon2020,Thambawita2021}. The former is achieved by extending the training set with new samples in keeping with the original data to support model learning.

    In more detail, a successfully trained classification model $h_{\theta}$ will have learned a close distribution estimation of the classes $\mathbf{Y}$ in the feature space $\mathbf{X}$, P$(Y|X)$~\cite{Bashir2020}. In other words, it will map latent variable relationships, and therefore certain feature space regions, to a particular class $y \in \mathbf{Y}$:
    \begin{gather}
        h_{\theta}(\mathbf{X}) \approx \mathbf{Y}.
    \end{gather}
    The more densely a class $y$ is represented in a region during training, the more it will be reinforced in the model's feature space representation. However, if there are too few training samples, the model is underfit and will have learned a skewed representation of the dataset~\cite{Bejani2021}. Unfortunately, collecting more data to increase the sample size is difficult in many real-world situations. Prohibitors include the high data collection cost, low data availability, and lack of expertise for ground-truth labelling~\cite{Whang2023}.

    GANs are the most popular and ubiquitous synthesis technique to address this concern. Developed by Goodfellow et al., they encapsulate a zero-sum game between two models~\cite{Goodfellow2020}. The generator attempts to fool the discriminator with new samples synthesised from random noise. Simultaneously, the discriminator attempts to distinguish between original and generated samples. Over many training iterations, the generator's output becomes more and more realistic~\cite{Kammoun2023}. However, the final synthesised output is fundamentally difficult to evaluate because there is no inherent quality metric for the GAN's distributional fit~\cite{Borji2022,Navidan2021,Brophy2023}. We must rely on purely qualitative metrics and empirical results to estimate the samples' quality, which may be problematic for high-risk domains such as healthcare, finance, and security~\cite{Saxena2022}.

    Nonetheless, multiple approaches have been developed that show improved performance with empirical confidence measures. For example, Bhattarai et al.\ interpreted the probabilistic output of the GAN's generator and discriminator as a confidence measure~\cite{Bhattarai2020}. Based on this information, the synthetic samples were filtered for high scores, and empirical results for facial emotion recognition showed minor improvements. Similarly, Nie and Shen developed a difficulty-aware attention mechanism based on the model's confidence for medical image segmentation~\cite{Nie2020}. They aimed to improve overall performance by reducing the training weights of easy samples based on a `confidence' Convolutional Neural Network (CNN). The authors estimated the trustworthiness of image segments with the CNN's confidence maps and thereby took local confidence into account. In contrast, Du et al.\ developed a sophisticated confidence model to generate pseudo-healthy images of skin lesions~\cite{Du2022}. Using a GAN, the authors generated subject-specific healthy images paired with a pathological sample to support classification performance. As part of the framework, they included a confidence check such that only segments with a confidence greater than a pre-defined threshold were considered for prediction. More recent works combine evaluation metrics to quantify different desirable characteristics that a well-fitted GAN should portray. For example, Abdusalomov et al.\ evaluated the synthesised samples in two stages: their distributional similarity to the original data during training and their diversity and proximity to real images of different classes after training is complete~\cite{Abdusalomov2023}.
    While each of these examples improved prediction performance on specific datasets and tasks, there were no quality guarantees since the confidence measures were empirically derived from probabilistic neural network outputs.

    There is no question that GANs can be a powerful synthesis technique~\cite{Wang2023,Shi2023,zhao2022synthesizing}. Unfortunately, however, they can be notoriously difficult to train. In particular, they frequently inherit the Deep Learning requirement for relatively large and balanced datasets to enable strong generalisation~\cite{Huang2023}. Consequently, density-based generative models are a popular alternative for small datasets requiring extension~\cite{Plesovskaya2021}. For example, Kernel Density Estimation (KDE) is a state-of-the-art non-parametric approach to approximate the distribution of random variables~\cite{Park2024}. Part of KDE's versatility is the underlying kernel $K$, which may be defined to suit the underlying data~\cite{Bauer2024}. Consequently, using $n$ observations $x_i$, a random variable $x$'s probability density estimation $p(x)$ is defined as:
    \begin{equation}
        p(x) = \frac{1}{n}\sum^n_{i=1}K (x - x_i, w).
    \end{equation}
    The bandwidth $w$ acts as a smoothing parameter, regulating the estimation's bias-variance trade-off~\cite{Belhaj2024}. The fitted KDE model may be repeatedly sampled to synthesise new data points. For example, Pozi et al.\ developed a privacy-preserving data synthesis algorithm building on this technique~\cite{Pozi2020}. After learning the probability density function of the original features with a KDE-based generative model, the distribution is carefully and deliberately shifted to obscure personally identifiable information. Through extensive classification experiments, the authors found that the shifted data's utility remained equal to the original data, maintaining downstream prediction performances.

    Similar to GANs, empirical evaluations are a common occurrence for density-based generative models because they offer no inherent performance guarantees. Additionally, synthesis performance may be highly variable depending on the underlying data and user-selected parameters (e.g., the bandwidth)~\cite{Falxa2023}. In contrast, theoretically supported confidence frameworks such as Conformal Prediction have an enormous potential to increase trust in synthesised data. However, to the best of our knowledge, Conformal Prediction has not previously been utilised for data synthesis. Apart from conformal clustering (discussed in \Cref{sec:conformalClustering}), the closest related work is presented by Liu et al. The authors proposed the conformal framework for semi-supervised learning~\cite{Liu2021}. In particular, Inductive Conformal Prediction was used to measure the quality of augmented samples. As a first step, the highest-confidence original samples were pre-filtered for augmentation. After augmentation, the highest-credibility adjusted samples were retained to extend the training dataset. The proposal was tested on a small dataset for herbal medicine classification, and the resulting prediction performance improved on traditional augmentation techniques. The downside was the high computational inefficiency of the approach, which was further improved in~\cite{Liu2022}. As the foundation of our confident synthesis proposal, \Cref{sec:cpBackground} further discusses the Conformal Prediction confidence framework in more depth.

\section{Conformal Prediction background}\label{sec:cpBackground}
    \emph{How confident are we that a model’s prediction is correct?} This is the core question that uncertainty quantification techniques such as Conformal Prediction attempt to answer.
    To contextualise our conformal synthesis proposal (\Cref{sec:proposal}), we introduce the conformal framework for classification (\Cref{sec:cpBasics}) and summarise the conformal clustering work that inspired our approach (\Cref{sec:conformalClustering}).

    \subsection{The conformal framework}\label{sec:cpBasics}
        Conformal Prediction (CP) is a highly versatile confidence framework that acts as a wrapper to any underlying point prediction model~\cite{Zhang2021}.
        Based on hypothesis testing, CP's uncertainty measures for individual predictions are statistically supported to a user-selected significance level.
        This section focuses on conformal classification because of its relevance to the proposed synthesis algorithm. Interested readers are referred to~\cite{Johansson2014} for conformal regression.

        Under minimal exchangeability assumptions, the conformal validity property guarantees that prediction mistakes are made up to a maximum error rate~\cite{Angelopoulos2021}.
        To achieve this strong guarantee, an underlying model's point predictions are transformed into a prediction set $\Gamma$ including all plausible labels $y \in \mathbf{Y}$. Label inclusion is driven by the significance-level $\epsilon$, making the prediction sets $\epsilon$-dependent~\cite{Messoudi2020}. An error is defined as a prediction set missing a sample's true label $y^*$. Due to the hypothesis testing background, the probability of a mistake being made on each sample $i$ is capped at $\epsilon$, subject to statistical fluctuations (\Cref{eq:validity}). By the law of large numbers, the overall probability of an error occurring approaches $\epsilon$ with increasing predictions, while the individual error probability is unchanged~\cite{Zhan2020}:
        \begin{gather}
            P \left( y_i^* \notin \Gamma_i^{\epsilon} \right) \leq \epsilon, \label{eq:validity} \\
            \lim_{i\to\infty} \text{ avg } P \left( y_i^* \notin \Gamma_i^{\epsilon} \right) \approx \epsilon.
        \end{gather}

        Originally designed for an online setting, transductive CP requires retraining the conformal model for every new test sample. Therefore, an inductive variant (ICP) was proposed to remove the need for leave-one-out retraining~\cite{Papadopoulos08}. However, similar to the original approach, the guarantees are valid over all predictions but not necessarily per class. For example, `difficult' samples belonging to a minority class may average higher error rates than `easier' samples~\cite{Ashby2022a}. Therefore, the Mondrian variant (MICP) builds on ICP to address this challenge by extending the guarantees to label-conditional validity~\cite{Lofstrom2015}:
        \begin{equation}
            P \left( y^* \notin \Gamma^{\epsilon} \mid y^*=y \right) \approx \epsilon, \qquad \forall \quad y \in \mathbf{Y}.
        \end{equation}
        Due to its unique combination of advantages, the following section describes the MICP method of prediction set construction.

        The strong validity property is guaranteed through carefully constructing the prediction sets $\Gamma^\epsilon$. To prepare for inductive inference, we split the original training set $Z_{train}$ with samples $x \in \mathbf{X}$ and true labels $y^* \in \mathbf{Y}$ into the disjoint proper training set $Z_{prop}$ and calibration set $Z_{calib}$:
        \begin{align}
            Z_{train} &= \Set{i=1, \ldots, n | \left(x_i, \text{ } y^*_i \right)}, \\
            Z_{prop} &= \Set{i=1, \ldots, m | \left(x_i, \text{ } y^*_i \right)}, \\
            Z_{calib} &= \Set{i=m+1, \ldots, n | \left(x_i, \text{ } y^*_i \right)}.
        \end{align}
        To make a prediction, a test sample $x_{n+1}$'s plausibility is evaluated through extension with each possible label $y \in \mathbf{Y}$~\cite{Meister2023}. To measure the likelihood of the postulated label given the original data, a non-conformity measure $A$ evaluates a test extension's `strangeness' with an underlying point prediction model $U$ trained on $Z_{prop}$:
        \begin{gather}
            \alpha_{n+1}^y = A_U \left((x_{n+1}, y), \text{ } Z_{prop} \right), \qquad \forall \quad y \in \mathbf{Y}.
        \end{gather}
        The calibration non-conformity scores are calculated similarly to the test sample's. However, instead of extending each calibration point with every possible label, we only require values for their true labels $y^*$:
        \begin{equation}
            \alpha_i^{y^*} = A_U \left((x_{i}, y_i^*), \text{ } Z_{prop} \right), \qquad \forall \quad (x_i, y_i^*) \in Z_{calib}.
        \end{equation}
        An example of a straightforward non-conformity measure uses the K-Nearest Neighbour algorithm (KNN) as its underlying model:
        \begin{gather}
            A_{KNN} \quad : \quad \alpha^y_{j} = \sum \min_k \text{\ dist} \left( x_{j}, \Set{ x_i \in Z_{prop} | y_i^* = y } \right). \label{eq:ncm}
        \end{gather}
        Here, $\alpha$ quantifies an extended sample $x_{j}$'s similarity to the observed data by summing the distances to its $k$ nearest neighbours with the same true label $y^*$ as the postulated class $y$.
        The larger $\alpha$ is, the further away the test sample $x$ is from a class $y$, and the less likely it is that $y$ is a plausible label~\cite{Ndiaye2022}.

        Given a test sample's and the calibration set's non-conformity scores, we may calculate the probability of each label $y$ being the true test label $y_{n+1}^*$ via $p$-values~\cite{Meister2020}. Note that through the hypothesis testing background, all true $p$-values $p^{y^*}$ are automatically guaranteed to be uniformly distributed~\cite{Sesia2021}:
        \begin{gather}
            p^{y}_{n+1} = \frac{\# \Set{ i=m+1, \ldots, n | y^*_i = y, \text{ } \alpha^{y^*}_i \geq \alpha^y_{n+1} } + 1}{ \# \Set{ i = m+1, \ldots, n | y^*_i = y } + 1} \label{eq:micpPvalues}.
        \end{gather}
        Finally, the prediction set $\Gamma^{\epsilon}_{n+1}$ is constructed by including all labels $y$ whose confidence levels exceed the significance level $\epsilon$:
        \begin{gather}
            \Gamma^{\epsilon}_{n+1} = \Set{ y \in \mathbf{Y} | p_{n+1}^y > \epsilon }.
        \end{gather}
        The larger $\epsilon$ is, the narrower the prediction set becomes, and the more likely it is that a sample's true label $y^*$ is not included. Therefore, the trade-off between low error rates and precise prediction sets must be carefully balanced. The optimal classification prediction set includes exactly one label~\cite{Vovk2016}. So far, Conformal Prediction has been presented in the context of prediction tasks. \Cref{sec:conformalClustering} discusses an extension to clustering that inspired our proposed algorithm.

    \subsection{Conformal clustering}\label{sec:conformalClustering}
        Conformal Prediction was originally developed to measure confidence and provide performance guarantees for prediction tasks. However, the framework has since been extended to a variety of application domains, and the one most relevant to this article is unsupervised clustering.

        Introduced in~\cite{Cherubin2015}, conformal clustering is founded on measuring the conformal confidence of each point in a feature space. A grid of points represented the feature space with equal spacing to make the task discrete. By treating each grid point as a test sample, Cherubin et al.\ measured their unsupervised similarity to the observed training data with conformal $p$-values (\Cref{sec:cpBasics}). All grid points with $p>\epsilon$ were considered part of the clusters, where $\epsilon$ was a user-specified threshold between 0 and 1. The clusters themselves were defined via the neighbouring rule: two grid points were part of the same cluster if they were neighbours. Therefore, in addition to setting the confidence threshold, $\epsilon$ may be considered a regularisation factor of the clusters' connectedness. The smaller $\epsilon$ was, the more connected the high-confidence grid points were, and the fewer distinct clusters were formed.

        Building on this work, Nouretdinov et al.\ further underpinned the understanding of conformal clustering~\cite{Nouretdinov2020}. The authors extended the approach to develop multi-level conformal clustering, introducing a dendrogram construction similar to traditional hierarchical clustering methods. Additionally, a new technique for identifying out-of-distribution anomalies was established by testing whether new samples fell within a conformal cluster. More recently, Jung et al.\ developed a novel conformity measure for clustering that is applicable to circular variables~\cite{Jung2021}. The authors demonstrated their approach by performing clustering on a dataset containing torsion measurements of different proteins.

        In a further recent development, Ding et al.\ proposed Clustered Conformal Prediction to incorporate a clustering aspect into conformal classifiers to improve class-conditional coverage in the many-class scenario~\cite{Ding2023}. The authors cluster classes with similar conformal score distributions based on the Mondrian CP variant. Calibration is then carried out within each cluster, achieving stronger `cluster-conditional' coverage over marginal coverage.
        However, for the purpose of our proposed synthesis algorithm, the previously mentioned conformal clustering concept of measuring feature space confidence is particularly interesting. For a successful synthesis, we would expect the new samples to overlap with the original data in the feature space. By limiting synthesis to high-confidence regions, outliers should be minimised. With this inspiration, \Cref{sec:proposal} introduces our proposed confident data synthesis algorithm.

\section{Conformalised data synthesis}\label{sec:proposal}
    In this section, we propose a unique data synthesis algorithm that measures the feature space confidence during the generation process. The basis of our confidence measure is the foundational Conformal Prediction framework, traditionally used for prediction and distribution testing tasks (\Cref{sec:cpBackground}). Our work presents a novel extension of the conformal technique to data generation, a previously unexplored domain. After examining our proposal's implications for the synthesised data (\Cref{sec:proposalImplications}), we comprehensively discuss our algorithm's design and characteristics (\Cref{sec:proposedAlgorithm}).

    \subsection{Implications for data generation}\label{sec:proposalImplications}
        Inspired by the innovative conformal clustering work summarised in \Cref{sec:conformalClustering}, our novel conformal data synthesis algorithm relies on the confidence of feature space regions to generate synthetic datasets.
        Consequently, unlike traditional synthetic data generation techniques (\Cref{sec:relatedWork}), our conformal synthesis algorithm models a confidence-aware distribution of the original dataset in the feature space.
        
\begin{figure}[!h]
                \centering
                \begin{subfigure}[b]{0.495\textwidth}
                    \centering
                    \includegraphics[width=\textwidth]{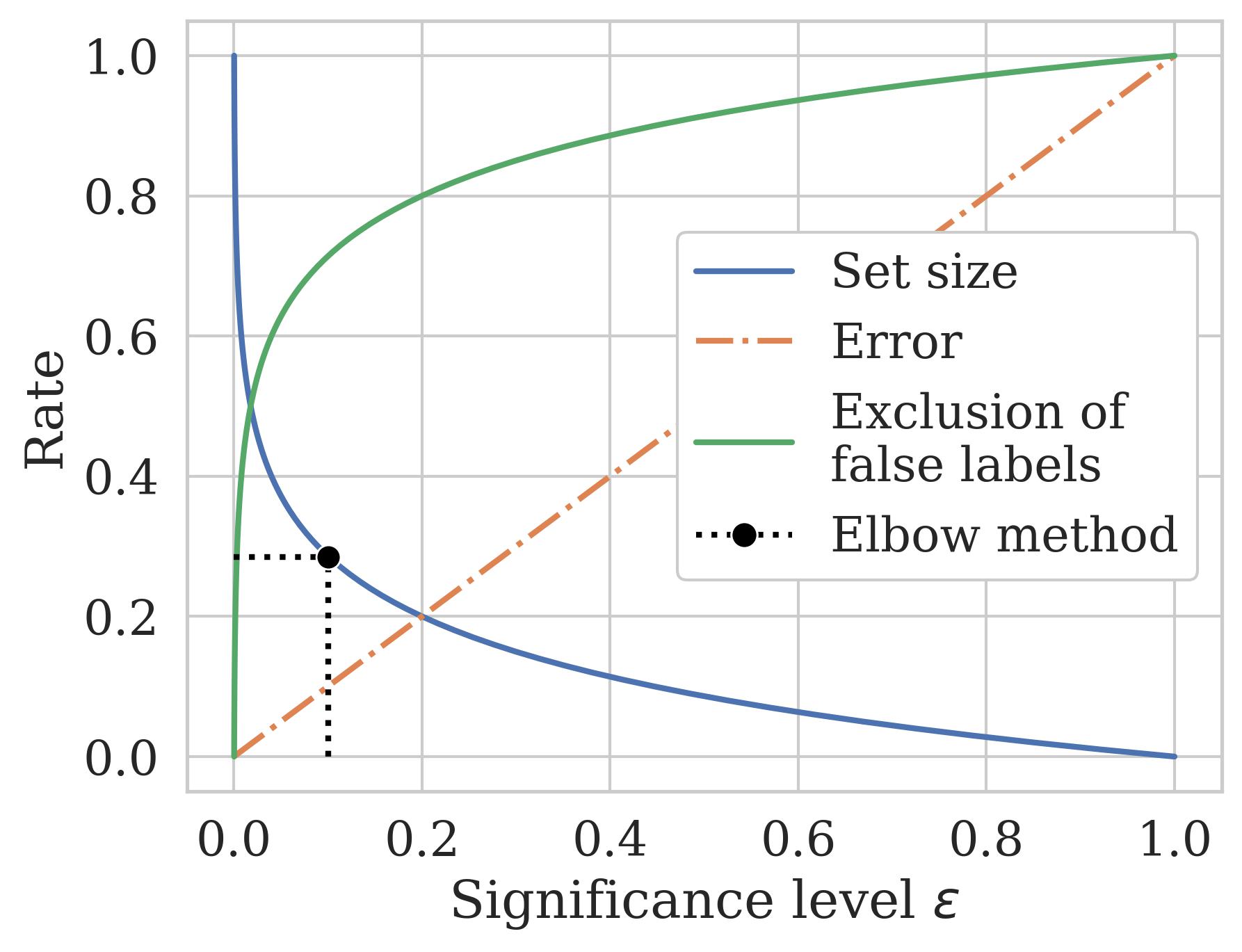}
                    \caption{Conformal classification.}\label{fig:epsilonTradeoffCp}
                \end{subfigure}
                \hfill
                \begin{subfigure}[b]{0.495\textwidth}
                    \centering
                    \includegraphics[width=\textwidth]{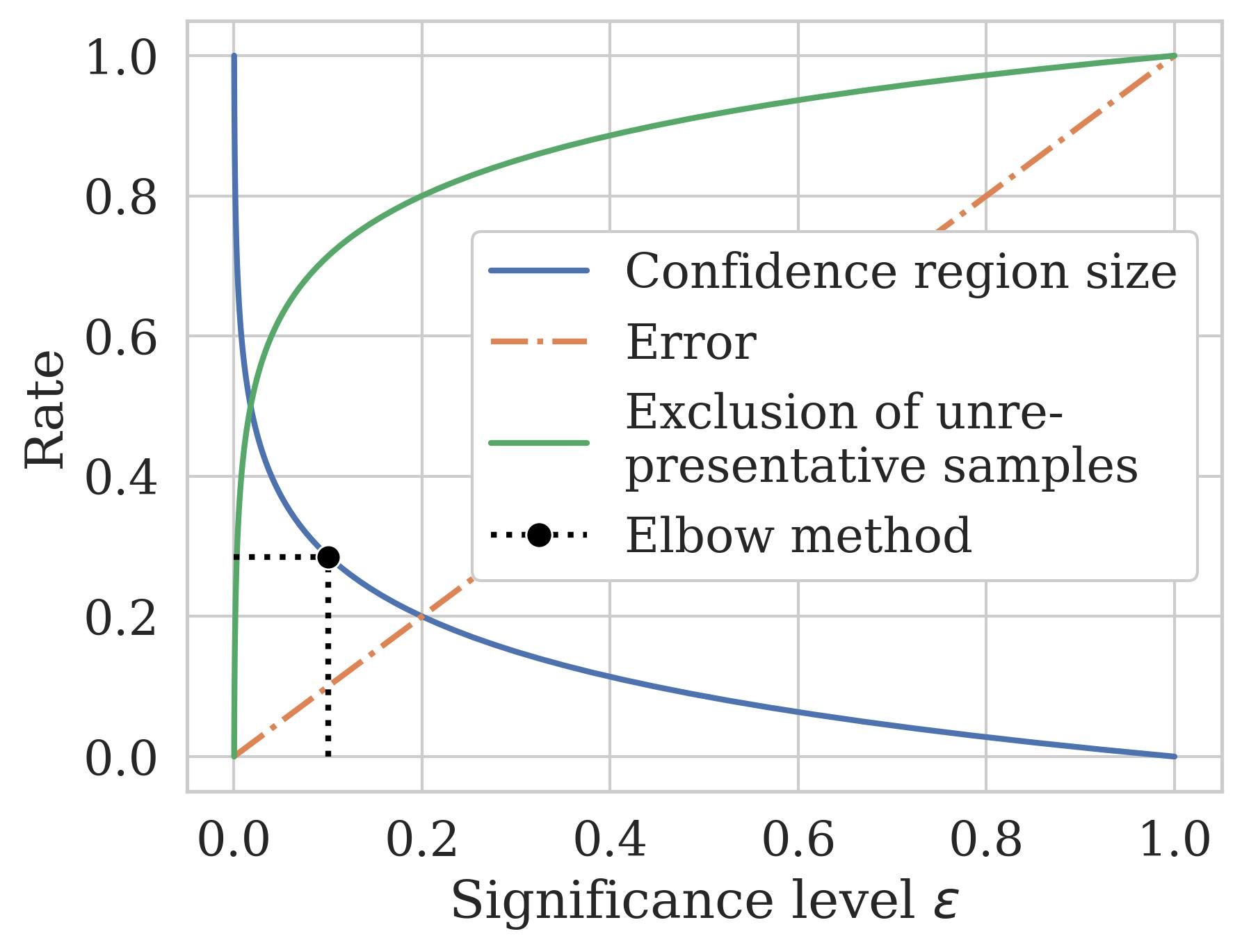}
                    \caption{Conformal synthesis.}\label{fig:epsilonTradeoffSynthesis}
                \end{subfigure}
                \caption{Intuition of the $\epsilon$ trade-off for traditional conformal classification and conformal synthesis. A small $\epsilon$ implies low error rates while increasing the inclusion of false labels or `unrepresentative' samples, respectively. Note that the latter and the graphs in (b) must be inferred from downstream model performances due to the data generation domain. The elbow method is a straightforward heuristic to select a value for $\epsilon$, balancing the opposing desires.}\label{fig:epsilonTradeoff}
            \end{figure}

        With traditional Conformal Prediction (CP), the significance level $\epsilon$ provides a statistically guaranteed error rate. Consequently, it directly regulates the trade-off between two opposing desires: low error rates and informative (i.e., narrow) prediction sets (\Cref{sec:cpBasics}). A common approach for optimising $\epsilon$ is via the elbow method heuristic~\cite{Jung2021}, identifying the point of diminishing return on improved set sizes at the cost of more frequent prediction errors (\Cref{fig:epsilonTradeoffCp}).
        For conformal synthesis, $\epsilon$ instead acts as a threshold to identify the high-confidence feature space regions from which samples are synthesised.
        However, it provides only limited guarantees due to the extension to data generation, discussed further in \Cref{sec:proposedAlgorithm}. Nonetheless, we assume and empirically investigate a close relationship between $\epsilon$ and the overall prediction performance on synthesised datasets in \Cref{sec:results}. Therefore, we must reinterpret the (approximated) $\epsilon$ trade off in the context of data generation.
        
        Our synthesis priority is to identify the high-confidence feature space regions, e.g., to support Deep Learning training. Consequently, an error is interpreted as the exclusion of `representative' synthetic samples. Intuitively, the smaller $\epsilon$ is, the wider the confidence regions must become to increase the likelihood of `representative' samples being included. However, the trade off is the inclusion of `unrepresentative' samples, reducing the synthetic dataset's effectiveness for model training (\Cref{fig:epsilonTradeoffSynthesis}). Similarly to traditional CP, we must balance the two desires by carefully optimising $\epsilon$.\@
        Note that because we cannot directly measure the rate of excluding `unrepresentative' samples, we assume a close relationship between excluding `unrepresentative' samples and a model's improved performance after training on the synthesised data. Therefore, we propose using the model's performance curve to identify $\epsilon$ via the elbow method to optimise downstream model performance, investigated in depth in \Cref{sec:toyDatasetSignificanceLevel}.

        The definition of the non-conformity measure (NCM) is an additional factor to consider for conformal synthesis. Feature space confidence regions are identified by comparing relative differences between non-conformity scores. As a core component to the calculations, the NCM may significantly impact the confidence regions' shapes and, consequently, how the original data's distribution is modelled for synthesis. For example, we could prioritise maintaining intra-class relationships by incorporating distances between classes or reducing noise by lowering outliers' importance. This flexibility is an inherent advantage of the CP framework and allows adjustments to be made that best suit the dataset for improved performance.
        With this understanding of our proposal's implications, \Cref{sec:proposedAlgorithm} details the algorithm's steps and characteristics.

    \subsection{Proposed algorithm}\label{sec:proposedAlgorithm}
        \Cref{algo:proposedAlgo} illustrates the logical flow of our proposed algorithm. Given a feature space $\mathbf{X} \in \mathcal{R}^d$, we construct a grid point representation $\mathbf{G}^{\gamma} \in \mathcal{R}^d$ of $\mathbf{X}$ with grid step $\gamma$ such that all observed samples $x \in \mathbf{X}$ fall within its boundaries. $\mathbf{X}_{grid}$ represents the collection of all grid points in $\mathbf{G}^{\gamma}$ (\Cref{line:gridConstruction}).
        To prepare for Mondrian Inductive Conformal Prediction (MICP), the training data $Z_{train} = \left(\left(x_1, y^*_1\right), \ldots, \left(x_n, y^*_n\right)\right)$ with $ x \in \mathbf{X}$ and $y^* \in \mathbf{Y}$ is split into the proper training and calibration subsets $Z_{prop} = \left(z_1, \ldots, z_m\right)$ and $Z_{calib} = \left(z_{m+1}, \ldots, z_{n}\right)$ in \Cref{line:icpDataSplit}. Then, we calculate the non-conformity scores $\alpha^{y^*}_{calib}$ of each point in the calibration set with their true label $y^*$ (\Cref{line:calibNCM}). In principle, any non-conformity measure $A_U$ may be used. For example, \Cref{eq:ncm} in \Cref{sec:cpBasics} defines a neighbour-based non-conformity measure $A_{KNN}$. Similar to the original CP framework, the choice may have a significant impact on the size of the prediction sets (i.e., the number of synthesised samples).

        \begin{algorithm}
            \caption{Logical flow of the proposed conformal synthesis algorithm based on Mondrian Inductive Conformal Prediction (MICP, \Cref{sec:cpBasics}). Each point in a discretised feature space is treated as a conformal test sample. Its class-conditional confidence dictates whether it is sampled as a synthetic point.}\label{algo:proposedAlgo}
            \begin{algorithmic}[1]
                \Require{A non-conformity measure $A$ with underlying predictor $U$, a grid step $\gamma$, a significance level $\epsilon$, and a labelled training set $Z_{train} = \left(x \in \mathbf{X}, \text{ } y^* \in \mathbf{Y}\right)$.}
                \Ensure{$\textbf{X} \in \mathcal{R}^d, \text{ } \textbf{Y} \in \mathcal{N}, \text{ } 0 \leq \epsilon \leq 1$.}
                \State{$\mathbf{X}_{grid} \Leftarrow \mathbf{G}^{\gamma} \in \mathbf{X}$}\label{line:gridConstruction}
                \State{$Z_{prop}, Z_{calib} \Leftarrow \text{split}\left(Z_{train}\right)$}\label{line:icpDataSplit}
                \State{$\alpha^{y^*}_{calib} \Leftarrow A_U\left(Z_{calib}, Z_{prop} \right)$}\label{line:calibNCM}  \Comment{For example, $A_{KNN}$ in \Cref{eq:ncm}.}
                \For{$y \in \textbf{Y}$}
                    \State{$\alpha^y_{grid} \Leftarrow A_U\left(\left(x \in \mathbf{X}_{grid},\text{ } y\right), Z_{prop}\right)$}\label{line:gridNCM}
                    \State{$p^y_{grid} \Leftarrow \text{calculate\_p\_values}\left(\alpha^{y^*}_{calib}, \text{ } \alpha^y_{grid}\right)$}\label{line:gridPvalues}  \Comment{Following \Cref{eq:micpPvalues}.}
                    \State{$\mathbf{R}_y^{\epsilon} \Leftarrow \Set{ x_{grid} | p^y_{grid} > \epsilon }$\label{line:highConfidenceRegions}  \Comment{Feature space confidence regions.}}
                \EndFor{}
                \State{$Z_{syn} \Leftarrow \bigcup_{y \in \mathbf{Y}} \Set{ \left( x, y \right) | x \in \mathbf{R}^{\epsilon}_y }$}\label{line:syntheticSamples}  \Comment{Sample synthesis from the confidence regions.}
            \end{algorithmic}
        \end{algorithm}

        \Cref{line:gridNCM,line:gridPvalues,line:highConfidenceRegions} contain the core of our confident data synthesis logic. To synthesise data points, we must first evaluate the confidence of each point in the feature space, represented by $\mathbf{X}_{grid}$. In Conformal Prediction terminology, we treat the grid points as test samples, extending them with each possible class label $y \in \mathbf{Y}$.
        Given the label-conditional non-conformity scores $\alpha^y_{grid}$, we calculate $p^y_{grid}$ following the Mondrian Inductive Conformal scheme (\Cref{eq:micpPvalues} in \Cref{sec:cpBasics}). The $p$-values represent each grid point's likelihood of being assigned to class $y$, assuming it represents class $y$. In other words, these $p$-values establish our confidence in a region's representation of a particular class. The threshold for the label-conditional confidence regions $\mathbf{R}_y^{\epsilon}$ is the user-selected significance-level $\epsilon$. With this information, all grid points falling into the label-conditional high-confidence regions $\mathbf{R}_y^{\epsilon}$ are sampled as synthetic data points with their matching class. Finally, the set of all synthetic points $Z_{syn}$ is constructed as the union of the label-conditional synthetic subsets (\Cref{line:syntheticSamples}).

        Derived from the validity property of CP (\Cref{eq:validityGuarantee}), we may draw some conclusions about the synthesised data's characteristics. In parallel with the probability of the true label $y^* \in \mathbf{Y}$ being included in the prediction set $\Gamma^{\epsilon}$, the synthesised data $\mathbf{R}_y^{\epsilon}$ will include a grid point with the true label ($y=y^*$) with a marginal probability of $1-\epsilon$:
        \begin{gather}
            \Gamma^{\epsilon} = \Set{ y \in \mathbf{Y} | p^y > \epsilon }, \text{ } \text{P} \left( y^* \notin \Gamma^{\epsilon} \right) \leq \epsilon, \label{eq:validityGuarantee} \\
            \mathbf{R}_y^{\epsilon} = \Set{ x_{grid} \in \mathbf{X}_{grid} | p^y_{grid} > \epsilon } = \Set{ x_{grid} \in \mathbf{X}_{grid} | y \in \Gamma^{\epsilon} }, \\
            \text{P}\left(y^* \notin \Gamma^{\epsilon} \right) = \text{P}\left(x^{y^*}_{grid} \notin \mathbf{R}_y^{\epsilon} \right), \label{eq:cpMarginalProbability} \\
            \text{P}\left(x^{y^*}_{grid} \notin \mathbf{R}_y^{\epsilon} \right) \leq \epsilon, \\
            \text{P}\left(x^{y^*}_{grid} \in \mathbf{R}_y^{\epsilon} \right) \geq 1 - \epsilon.  \label{eq:synMarginalProbability}
        \end{gather}
        However, inherited from the CP framework, no theoretically-founded conclusions can be drawn about the `false-class' predictions where $y\neq y^*$, or grid points being synthesised with a false label in conformal synthesis terms. Additionally, the focus lies on the feature space regions. Consequently, the synthesised dataset $\mathbf{R}_y^{\epsilon}$ does not necessarily follow the object distribution of the original data.
        
        Regarding practical properties, the algorithmic complexity is strongly influenced by the Inductive Conformal Prediction variant~\cite{Papadopoulos2007}. Assuming the underlying algorithm's training and application complexities $U_t$ and $U_a$, the number of original training samples $n$, the number of calibration samples $m$, the number of grid points $g$, and the number of classes $c$, the algorithmic complexity can be described as:
        \begin{gather}
            \Theta \left( m \cdot U_t + \left(n-m + c \cdot  g \right) \cdot U_a \right). \label{eq:synthesisComplexity}
        \end{gather}
        The largest individually contributing step is evaluating the per-class confidences of each point in the grid space (\Cref{line:gridNCM,line:gridPvalues} in \Cref{algo:proposedAlgo}). However, due to the inductive variant, this step is fully parallelisable both within and between classes. Furthermore, as long as the original data set is unchanged, the grid's $p$-values may be reused to generate confidence regions for any significance level $\epsilon$. Finally, the proposed algorithm's parameters may strongly influence the synthesised output and are investigated in depth in \Cref{sec:toyDataset}.

        In summary, our proposed conformal algorithm is a unique approach to employ feature space confidence for the data synthesis process. With this background, \Cref{sec:results} investigates our algorithm's empirical results on real-world datasets.

\section{Empirical results}\label{sec:results}

    In this section, we comprehensively evaluate our proposed algorithm's performance on real-world data and against a state-of-the-art density-based generative model. In particular, we investigated the following research questions:
    \begin{itemize}
        \item What influence do the conformal synthesis parameters have on the synthesised data? (\Cref{sec:toyDataset})
        \item To what degree does conformal synthesis improve Deep Learning performance on small and large datasets? (\Cref{sec:smallDataResults})
        \item How effective is conformal synthesis at equalising Deep Learning performance on highly imbalanced classes? (\Cref{sec:imbalancedDataResults})
        \item Can we successfully improve the quality of a dataset with overlapping classes by extending it with synthetic data points? (\Cref{sec:overlappingClassesResults})
        \item How accurately can we replace a real-world dataset with entirely synthetic examples? (\Cref{sec:synteticReplacementResults})
        \item How does our proposed synthesis algorithm perform compared to a state-of-the-art, density-based generative model? (\Cref{sec:kdeSynthesisComparison})
    \end{itemize}

    \subsection{Experimental setup}\label{sec:experimentalSetup}

        We conducted a rigorous empirical evaluation of the synthetic data points produced by our proposed conformal synthesis algorithm (\Cref{sec:proposedAlgorithm}).
        Because the expected use case is to bolster datasets with few samples for Machine Learning, we evaluated the effectiveness of our synthesised data by measuring the performance of a Deep Learning model trained on it.

        \subsubsection{Neural network architecture}
\begin{figure}[!h]
                \centering
                \includegraphics[width=0.7\linewidth]{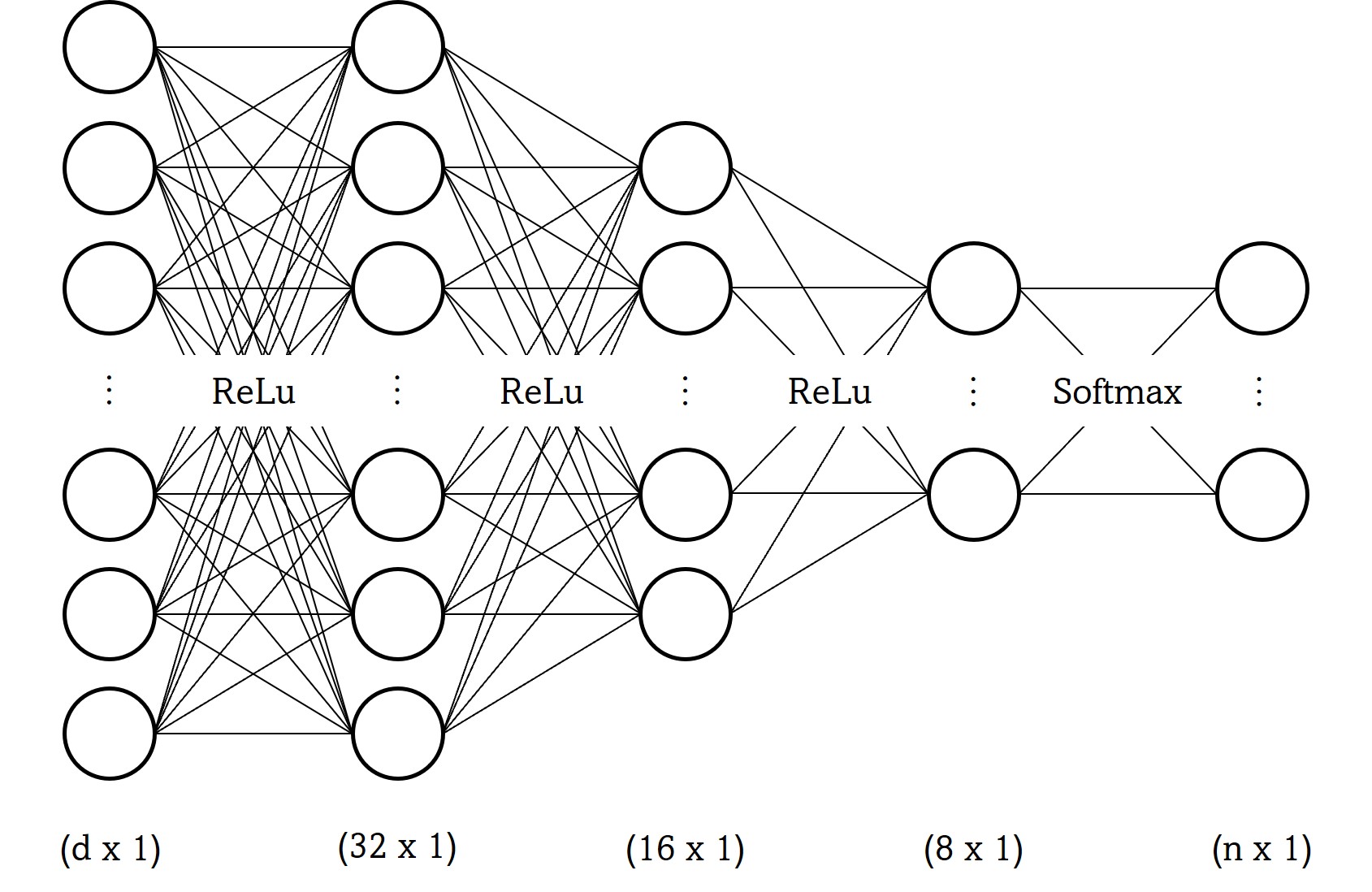}
                \caption{The Deep Learning model architecture used in all trials. Design decisions were made to improve the versatility on different real-world datasets and the robustness of the results. The input layer size $d$ and output layer size $n$ are driven by the dataset's dimensionality and number of classes.}\label{fig:nnModel}
            \end{figure}

            We chose a feedforward neural network with three hidden layers for the model, as shown in \Cref{fig:nnModel}. The model's architecture and parameters were chosen for their versatility and robustness to ensure a good baseline performance on all five selected datasets. The activation functions are the popular ReLu and Softmax functions for a classification output. The number of output nodes was adjusted based on the evaluated dataset's number of classes. The model was trained for ten epochs with early stopping enabled in all trials.

        % \FloatBarrier{}
        \subsubsection{Data subsets and synthesis parameters}\label{sec:dataAndSynthesisParams}
\begin{figure}[!h]
                \centering
                \includegraphics[width=0.75\linewidth]{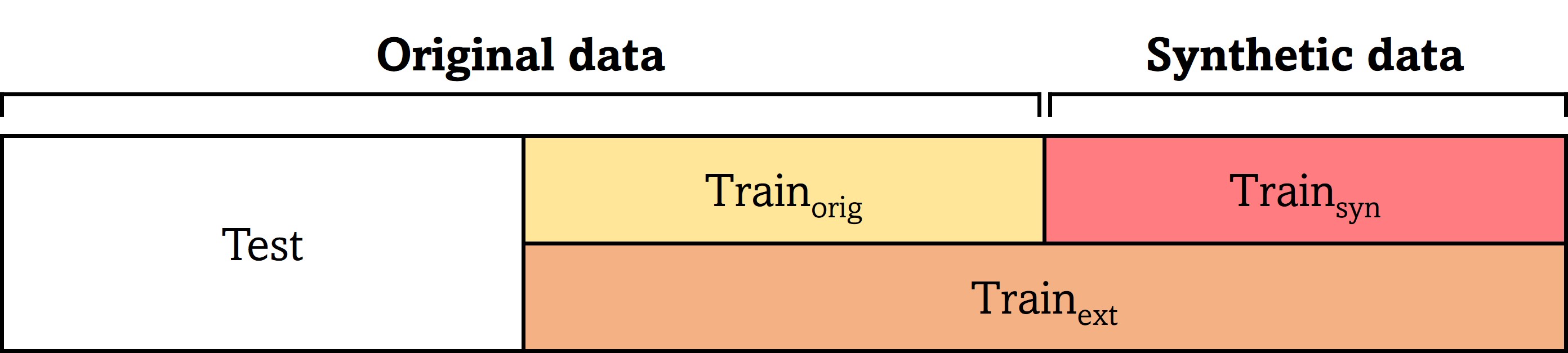}\caption{The original, synthesised, and extended training sets evaluated with Deep Learning on the same held-out test set. Initially, Train$_{\text{orig}}$ is temporarily divided into the proper training (60\%) and calibration sets (40\%) for conformal synthesis.}\label{fig:dataSubsets}
            \end{figure}

            Each trial was evaluated on the same test set for a meaningful comparison of the Deep Learning model's performance on the original data vs our synthesised data. \Cref{fig:dataSubsets} illustrates in more detail how the data was split and the evaluated training data configurations. To generate the synthetic samples, Train$_{\text{orig}}$ was temporarily divided into a calibration (40\%) and a proper training set (60\%). Each subset maintained the original class proportions. Primarily, one data split was tested and reported. However, the data split may affect the synthesis and model performance results. Therefore, three additional data splits were tested in some cases to support identifying high-level trends invariant to the randomness of how samples were assigned to the data subsets.

            Among a handful of other parameters, the conformal synthesis algorithm relies on defining a non-conformity measure (NCM) to identify the confidence regions of the feature space. We employed a KNN-based NCM that measures the sum of distances to the $k=5$ nearest neighbours using the generalised Minkowski distance (\Cref{eq:ncm}, \Cref{sec:cpBasics}). KNN was chosen as the underlying algorithm due to its simplicity, robustness, and popularity in the Conformal Prediction literature~\cite{Renkema2024,Hernandez2022,Liu2021}.
            A more common KNN-NCM additionally divides \Cref{eq:ncm} with the sum of $k$ minimum distances to instances of a different class, reducing confidence in regions with class overlap. However, our reasoning for employing the simplified NCM is to ensure that those overlapping regions are sufficiently represented in the synthesised datasets. Since feature space overlap can be common in real-world datasets, a larger volume of ``difficult'' synthesised samples may help Deep Learning models better generalise and distinguish them. \Cref{sec:discussion} discusses the potential of alternative NCMs on synthesis performance.
            The remaining synthesis parameters' intuition and influence on the generated data are investigated in depth in \Cref{sec:toyDataset}.

        \subsubsection{Performance metrics and statistical tests}
            Our primary performance metrics were the total and per-class F$_1$-scores. Unlike alternatives like accuracy and ROC-AUC, the F$_1$-score is robust against dataset imbalances. The per-class $F^c_1$ is defined as the harmonic mean of precision $P^c$ and recall $R^c$, which in turn are calculated from the true positive $TP$, the false positive $FP$, and the false negative $FN$ rates:
            \begin{align}
                P^c &= \frac{TP}{TP + FP}, \\
                R^c &= \frac{TP}{TP + FN}, \\
                \text{F}^c_1 &= 2 \cdot \frac{P^c \cdot R^c}{P^c + R^c}.
            \end{align}
            To ensure that the results were representative regardless of variance in the model's training, we repeated each experiment five times and reported the macro-average of all per-class scores:
            \begin{align}
                P &= \frac{\sum_{c \in C} P^c}{ | C | }, \\
                R &= \frac{\sum_{c \in C} R^c}{ | C | }, \\
                \text{F}_1 &= \frac{\sum_{c \in C} \text{F}^c_1}{ | C | }.
            \end{align}

            The Wilcoxon signed-rank test was employed to ascertain whether the Deep learning results improved, following~\cite{toccaceli2019combination,Liu2022,johansson2017model,Norinder2021,Campagner2024}. The non-parametric test compares two paired groups to investigate whether they are statistically different. The null hypothesis assumes that the median of differences of matched pairs is equal to 0~\cite{Rainio2024}. It was selected after the Kolmogorov-Smirnov test revealed that the results were overwhelmingly not normally distributed. The Wilcoxon test was carried out repeatedly, comparing a pair of models each time:
            \begin{itemize}
                \item Comparing parallel model test results after training on the original Train$_{\text{orig}}$ and the extended Train$_{\text{ext}}$ sets;
                \item And comparing two models trained on equivalent datasets generated from different splits of the original data (e.g., Train$_{\text{orig}}$ from two data splits).
            \end{itemize}

    \subsection{Illustrating the parameters' influence}\label{sec:toyDataset}
        We employed a simple toy dataset to demonstrate the influence of our proposed data synthesis algorithm's three parameters: the significance level $\epsilon$, the number of original training samples $n$, and the grid step $\gamma$.

\begin{figure}[!h]
            \centering
            \includegraphics[width=0.4\linewidth]{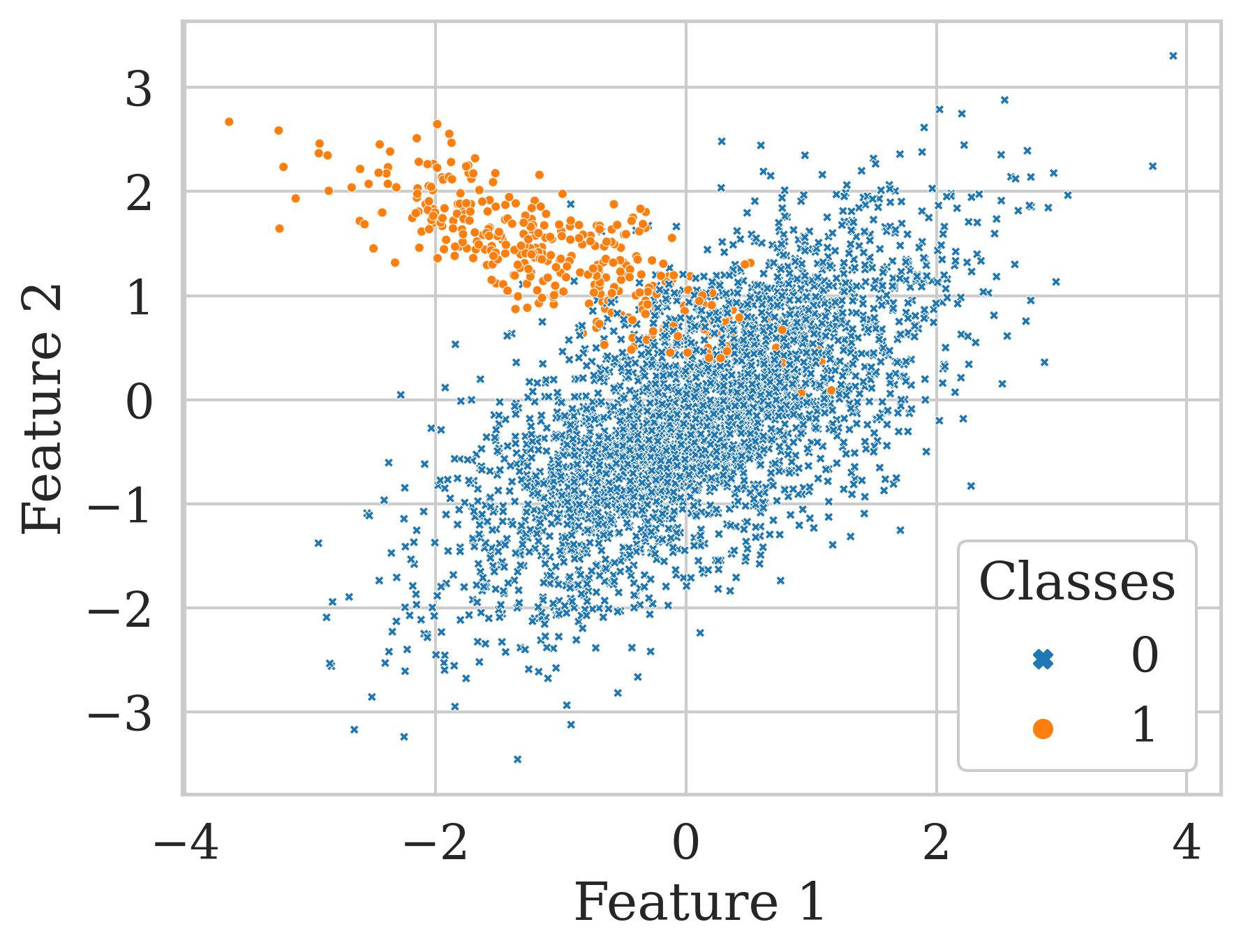}
            \caption{A straightforward 2-dimensional toy dataset used to illustrate the influence of the proposed algorithm's parameters.}\label{fig:toyOriginalDataset}
        \end{figure}

        Visualised in \Cref{fig:toyOriginalDataset}, the 2-dimensional classification dataset was generated with 2,000 samples and a 1:9 class imbalance. The two classes slightly overlap in the feature space, making the classification task more challenging but with an otherwise relatively clear class distinction. We limited the training set to 1,000 samples to simulate a small dataset, and the remaining 1,000 samples were used to evaluate the Deep Learning model's performance robustly. Each data subset maintained the original class distribution, including when the training set was artificially reduced in later sections (\Cref{tab:toyDatasetSampleCounts}).

       {\renewcommand{\arraystretch}{0.88}
       \begin{table}[!h]
            \centering
            \captionsetup{width=\textwidth}
            \caption{Toy dataset sample counts. The 1:9 class ratio is maintained in each data subset. 60\% of the training data is temporarily allocated to the proper training set and 40\% to the calibration set for synthesis.}\label{tab:toyDatasetSampleCounts}
            \begin{tabular}{llrrrr}
                \toprule
                 &  & \multicolumn{1}{c}{\textbf{Test}} & \multicolumn{3}{c}{\textbf{Train$\mathbf{_\text{orig}}$}} \\
                \cmidrule(lr){3-3} \cmidrule(lr){4-6}
                \multicolumn{1}{l}{\textbf{Subset}} & \textbf{Class} & \multicolumn{1}{c}{\textbf{All}} & \textbf{Prop.} & \multicolumn{1}{c}{\textbf{Calib.}} & \multicolumn{1}{c}{\textbf{All ($n$)}} \\
                \midrule\midrule
                \multirow{3}{*}{\textbf{$n$ = 1000}} & \textbf{Class 0} & 900 & 540 & 360 & 900 \\
                 & \textbf{Class 1} & 100 & 60 & 40 & 100 \\
                 & \textbf{All} & 1,000 & 600 & 400 & 1,000 \\
                \midrule
                \multirow{3}{*}{$n$ = \textbf{300}} & \textbf{Class 0} & 900 & 162 & 108 & 270 \\
                 & \textbf{Class 1} & 100 & 18 & 12 & 30 \\
                 & \textbf{All} & 1,000 & 180 & 120 & 300 \\
                \midrule
                \multirow{3}{*}{$n$ = \textbf{150}} & \textbf{Class 0} & 900 & 81 & 54 & 135 \\
                 & \textbf{Class 1} & 100 & 9 & 6 & 15 \\
                 & \textbf{All} & 1,000 & 90 & 60 & 150 \\
                \bottomrule
            \end{tabular}
        \end{table}}

% \FloatBarrier{}
        \subsubsection[Significance level]{Significance level $\epsilon$}\label{sec:toyDatasetSignificanceLevel}
\begin{figure}[!h]
                \centering
                \begin{subfigure}[b]{0.49\textwidth}
                    \centering
                    \includegraphics[width=\textwidth]{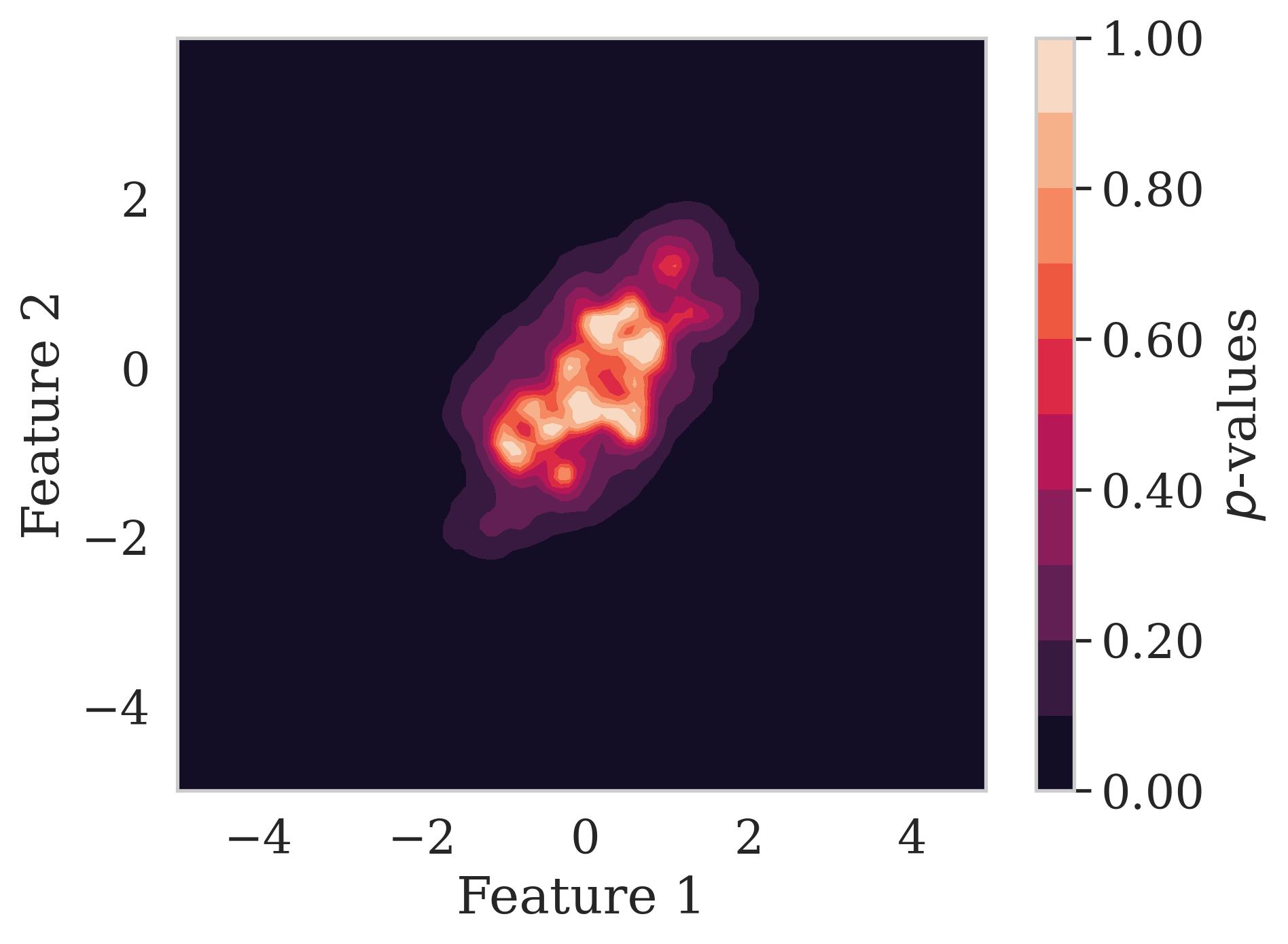}
                    \caption{Class 0 confidence regions.}
                \end{subfigure}
                \hfill
                \begin{subfigure}[b]{0.49\textwidth}
                    \centering
                    \includegraphics[width=\textwidth]{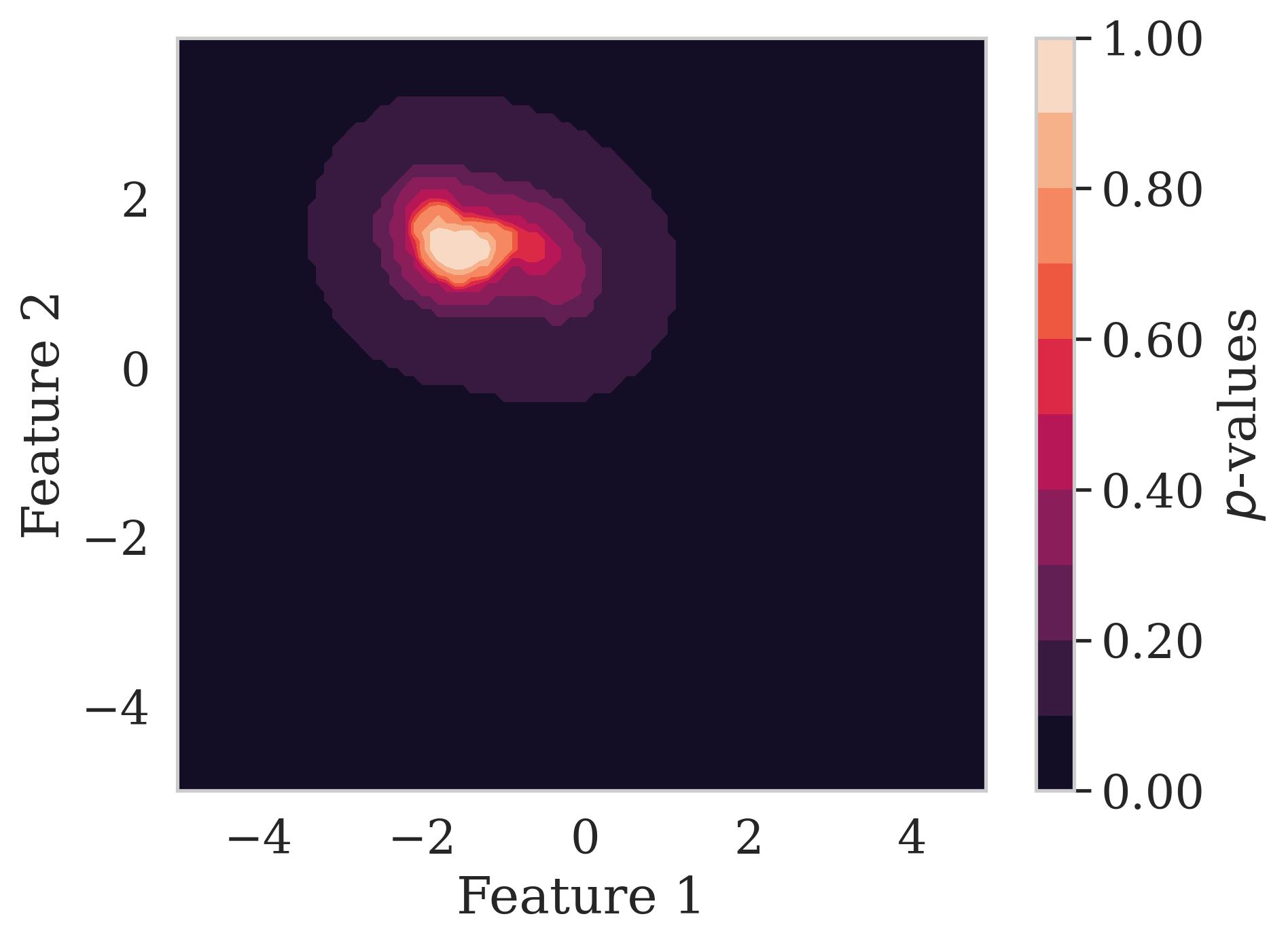}
                    \caption{Class 1 confidence regions.}
                \end{subfigure}
                \caption{Effects of the significance level $\epsilon$ on the feature space's confidence regions. The larger $\epsilon$ was, the more controlled the regions were around the original training samples. New samples were synthesised from the high-confidence regions, defined as $p > \epsilon$. }\label{fig:toyConfidenceRegions}
            \end{figure}

            Inherited from the Conformal Prediction framework, the significance level $\epsilon$ is a central parameter of our proposed algorithm. With it, a user may set the confidence threshold for data synthesis (\Cref{sec:proposedAlgorithm}). Given the $p$-values of each point in the feature space grid, we identified the per-class high-confidence regions where $p > \epsilon$. The larger $\epsilon$ was, the more carefully we controlled the high-confidence regions, tightening them around the existing data points (\Cref{fig:toyConfidenceRegions}). Consequently, we see a characteristic shape in the model's F$_1$-score performance. The following results were generated while holding the remaining synthesis parameters steady to isolate $\epsilon$'s effect ($n=300$, $\gamma=0.1$). Note that because fewer minority class 1 samples were available, the confidence regions are slightly broader around the original samples than class 0.

\begin{figure}[!h]
                \centering
                \begin{subfigure}[b]{0.49\textwidth}
                    \centering
                    \includegraphics[width=\textwidth]{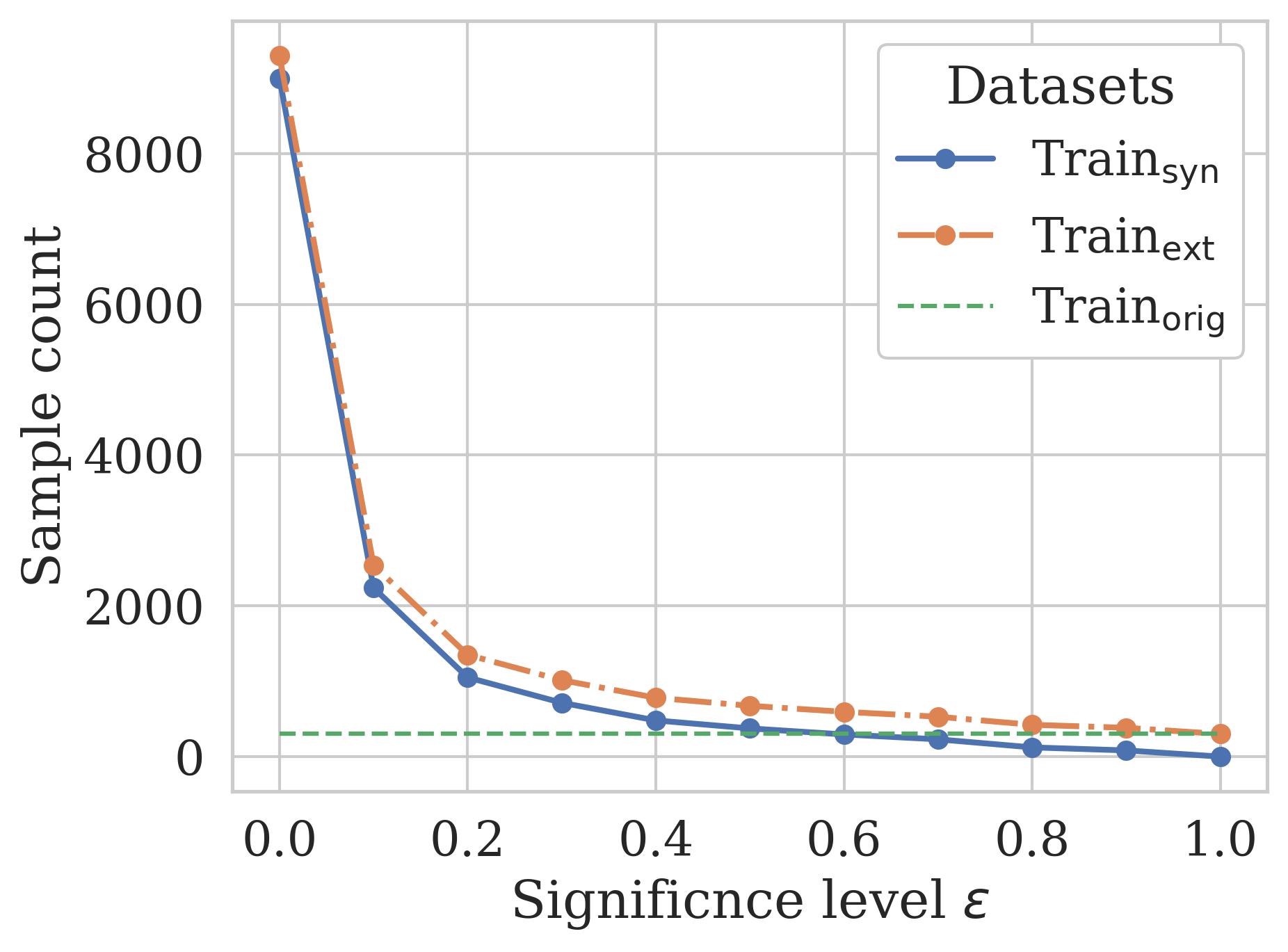}
                    \caption{Training set sizes.}
                \end{subfigure}
                \hfill
                \begin{subfigure}[b]{0.49\textwidth}
                    \centering
                    \includegraphics[width=\textwidth]{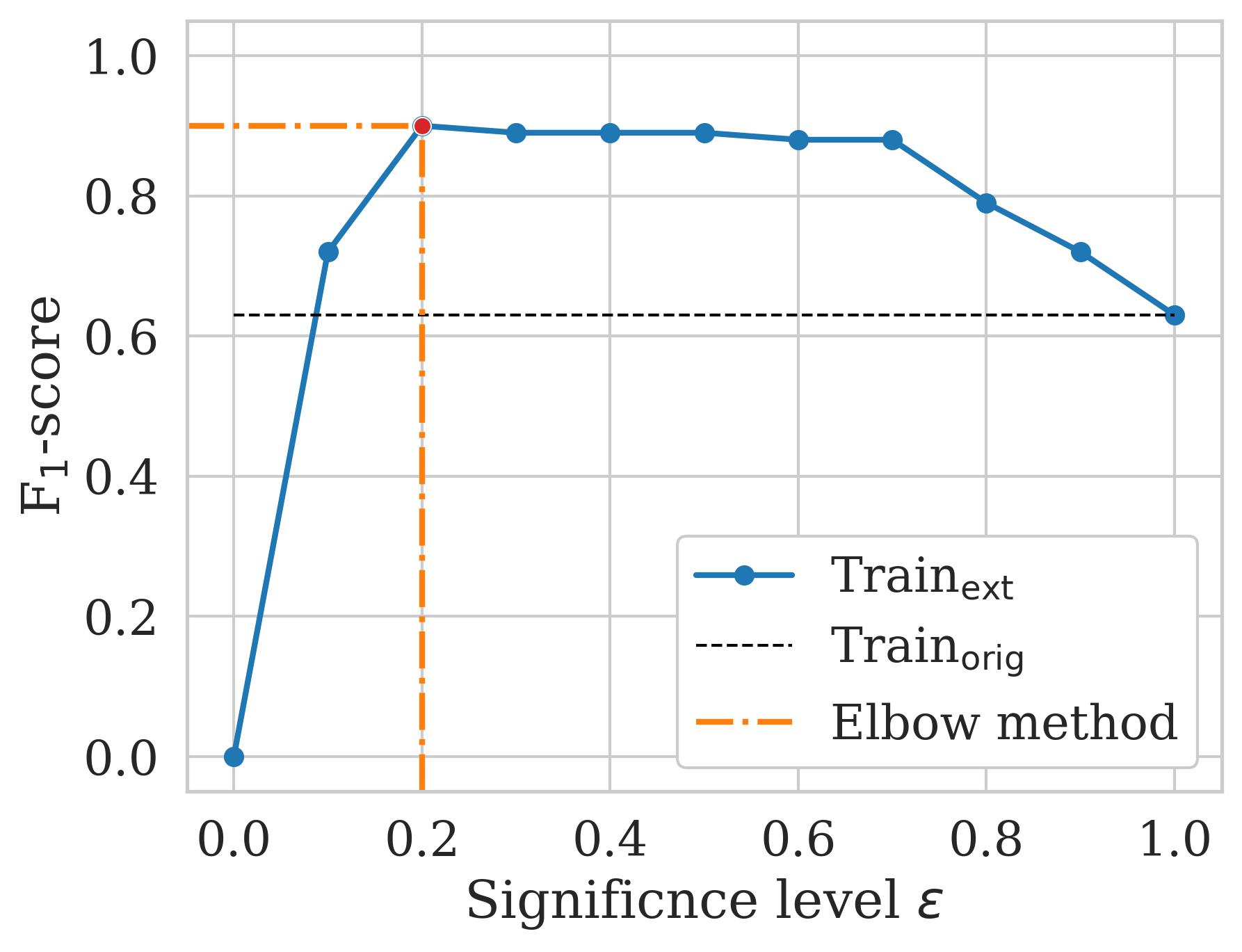}
                    \caption{Model performances.}\label{fig:toyEpsilonSelection}
                \end{subfigure}
                \caption{The effect of $\epsilon$ on the synthesised samples and a model's F$_1$-score after training on the extended dataset. A characteristic performance curve was revealed, significantly increasing model performance compared to the baseline for a large range of $\epsilon$. The elbow heuristic was employed to identify $\epsilon=0.2$ as optimal.}\label{fig:toyEpsilonEffect}
            \end{figure}

            Due to its significant impact on the synthesised data, $\epsilon$ also indirectly affected model performance, as visualised in \Cref{fig:toyEpsilonEffect}. As $\epsilon$ grew, fewer synthetic samples were generated because the confidence regions became narrower. This increased the likelihood that representative samples were not included in the extended set (i.e., the synthesis error rate increases). However, the likelihood of unrepresentative samples being falsely included in the extended set was also reduced. At $\epsilon=0$, we effectively had no confidence threshold, resulting in all possible points in the feature space being sampled to extend the training set for every class. Consequently, a model could understandably no longer distinguish the classes. However, as $\epsilon$ increased and the confidence regions became more distinguishing, we saw a sharp increase in modelling performance. Ideally, an equilibrium between the desire for improved model performance and low synthesis error rates is reached at low $\epsilon$. An intuitive heuristic is the elbow method (\Cref{sec:proposalImplications}), which identified $\epsilon=0.2$ in this case. Past this point, the F$_1$-score after training on the Train$_{\text{ext}}$ dataset plateaued, increasing model performance by about 30 percentage points compared to models trained on Train$_{\text{orig}}$. Finally, as the number of synthesised samples became negligible, the models' performance on both the original and extended datasets converged.

            Analysing the concrete performance results in \Cref{tab:toyEpsilonEffect}, we note that the majority of $\epsilon$-dependent extended training sets Train$_{\text{ext}}$ improved model performances with statistical significance according to the Wilcoxon test. The best performance was indeed achieved at $\epsilon=0.2$ with F$_1 = 90\%$ compared to the baseline of 63\%, as indicated by the previously visualised F$_1$ curve. Increasing the training set with data synthesis improved the model's ability to distinguish between classes reliably. Note that the conformal error guarantees do not automatically apply to the model's performance.

            \begin{table}[!h]
                \centering
                \setlength{\tabcolsep}{3.75pt}
                \captionsetup{width=\textwidth}
                \caption{Effects of the significance level $\epsilon$ on Deep Learning test results, the mean and standard deviation are reported. Training was carried out on the original and extended training sets. The optimal $\epsilon=0.2$ is in bold, selected in \Cref{fig:toyEpsilonSelection}. The majority of results ($0.2 \leq \epsilon \leq 0.7$) passed the Wilcoxon test ($p_W < 0.1$) and significantly improved results, marked with *.}\label{tab:toyEpsilonEffect}
                \begin{tabular}{ccrrcrrcrr}
                    \toprule
                    \textbf{} & \multicolumn{3}{c}{\textbf{F$_\mathbf{1}$-score}} & \multicolumn{3}{c}{\textbf{Precision}} & \multicolumn{3}{c}{\textbf{Recall}} \\
                    \cmidrule(lr){2-4} \cmidrule(lr){5-7} \cmidrule(lr){8-10}
                    \textbf{$\mathbf{\epsilon}$} & \textbf{Train$_\text{orig}$} & \multicolumn{1}{c}{\textbf{Train$_\text{ext}$}} & \multicolumn{1}{c}{\textbf{$\mathbf{p_W}$}} & \textbf{Train$_\text{orig}$} & \multicolumn{1}{c}{\textbf{Train$_\text{ext}$}} & \multicolumn{1}{c}{\textbf{$\mathbf{p_W}$}} & \textbf{Train$_\text{orig}$} & \multicolumn{1}{c}{\textbf{Train$_\text{ext}$}} & \multicolumn{1}{c}{\textbf{$\mathbf{p_W}$}} \\
                    \midrule\midrule
                    \textbf{0.1} & \multirow{4}{*}{''} & 0.72 (.16) & 0.46\:\: & \multirow{4}{*}{''} & 0.79 (.02) & 0.31\:\: & \multirow{4}{*}{''} & 0.91 (.00) & 0.01* \\  % chktex 21
                    \textbf{0.2} &  & \textbf{0.90} (.00) & 0.02* &  & \textbf{0.89} (.01) & 0.09* &  & \textbf{0.91} (.00) & 0.01* \\
                    \textbf{0.3} &  & 0.89 (.01) & 0.03* &  & 0.88 (.02) & 0.11\:\: &  & 0.91 (.01) & 0.01* \\  % chktex 21
                    \textbf{0.4} &  & 0.89 (.02) & 0.03* &  & 0.91 (.04) & 0.08* &  & 0.88 (.03) & 0.01* \\
                    \textbf{0.5} & \multicolumn{1}{l}{0.63 (.22)} & 0.89 (.02) & 0.03* & \multicolumn{1}{l}{0.66 (.28)} & 0.91 (.05) & 0.08* & \multicolumn{1}{l}{0.62 (.17)} & 0.87 (.04) & 0.01* \\
                    \textbf{0.6} & \multirow{4}{*}{''} & 0.88 (.02) & 0.03* & \multirow{4}{*}{''} & 0.91 (.05) & 0.08* & \multirow{4}{*}{''} & 0.84 (.05) & 0.01* \\
                    \textbf{0.7} &  & 0.88 (.03) & 0.03* &  & 0.93 (.05) & 0.07* &  & 0.84 (.05) & 0.02* \\
                    \textbf{0.8} &  & 0.79 (.15) & 0.21\:\: &  & 0.95 (.03) & 0.04* &  & 0.75 (.15) & 0.23\:\: \\  % chktex 21
                    \textbf{0.9} &  & 0.72 (.18) & 0.49\:\: &  & 0.85 (.23) & 0.26\:\: &  & 0.69 (.16) & 0.54\:\: \\  % chktex 21
                    \bottomrule
                \end{tabular}
            \end{table}

    \subsubsection[Original training sample count]{Original training sample count $n$}\label{sec:toyDatasetOriginalSampleCount}
\begin{figure}[!h]
                \centering
                \begin{subfigure}[b]{0.3\textwidth}
                    \centering
                    \includegraphics[width=\textwidth]{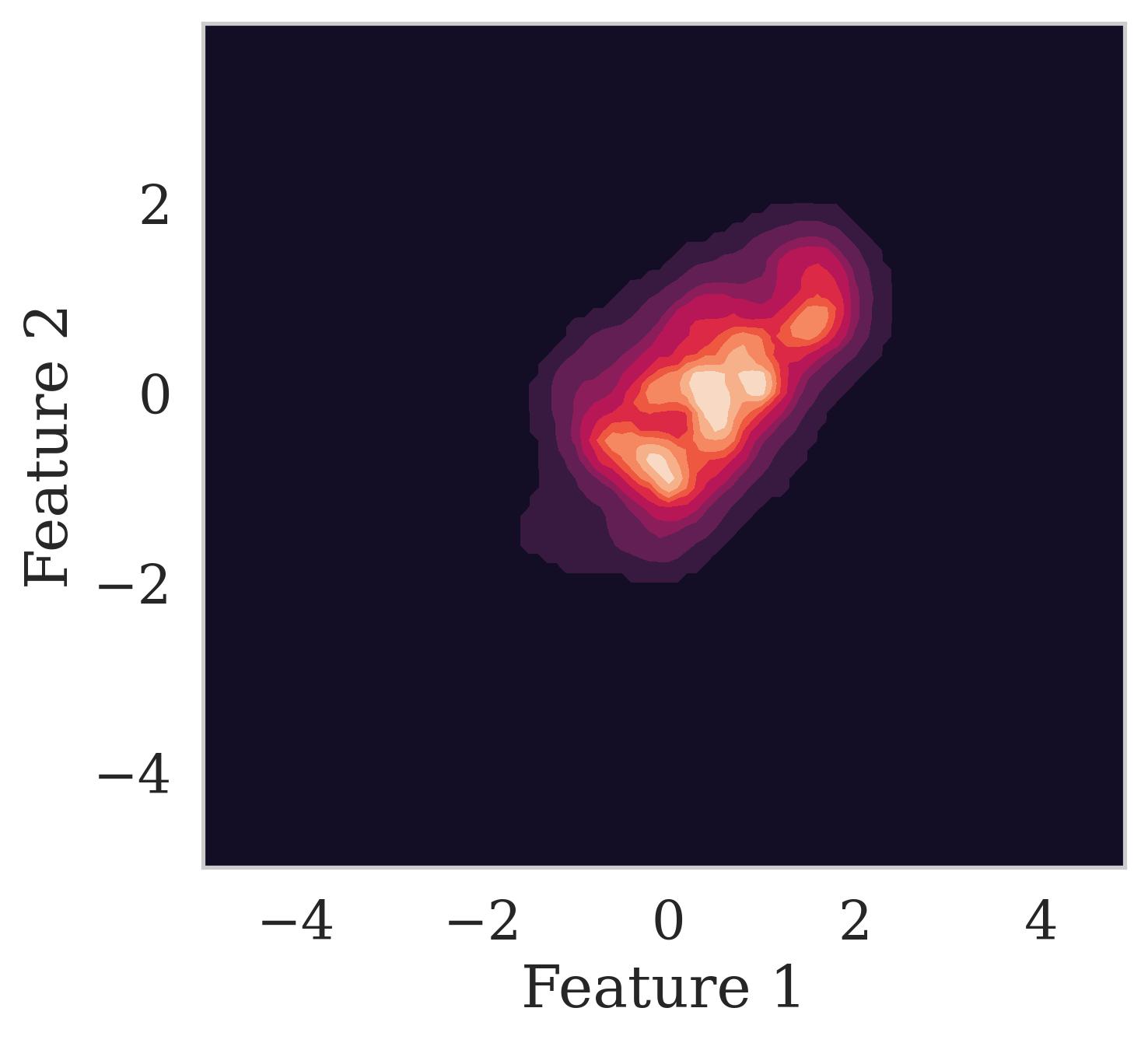}
                    \caption{$n = 150$.}
                \end{subfigure}
                \hfill
                \begin{subfigure}[b]{0.3\textwidth}
                    \centering
                    \includegraphics[width=\textwidth]{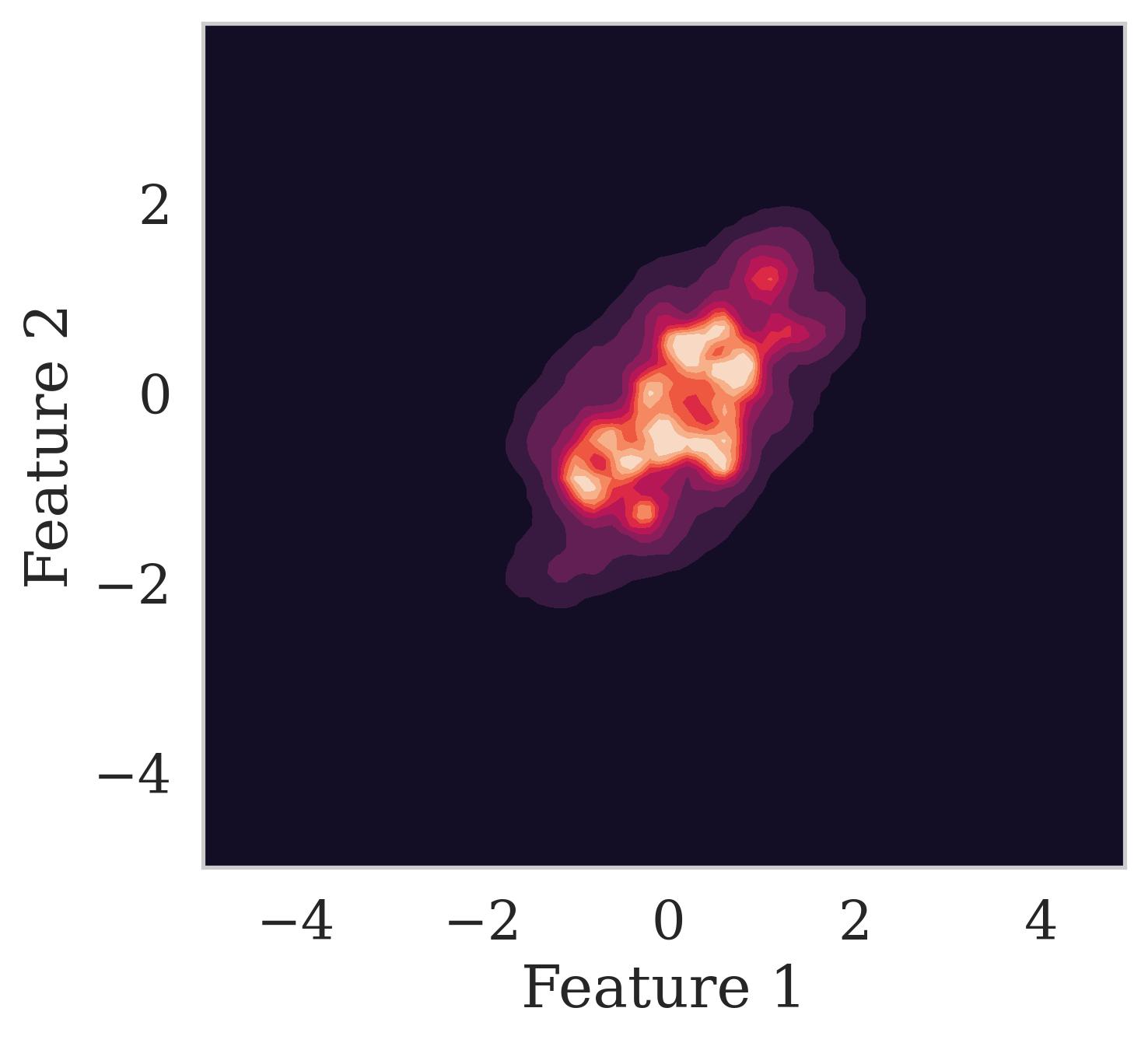}
                    \caption{$n = 300$.}
                \end{subfigure}
                \hfill
                \begin{subfigure}[b]{0.3744\textwidth}
                    \centering
                    \includegraphics[width=\textwidth]{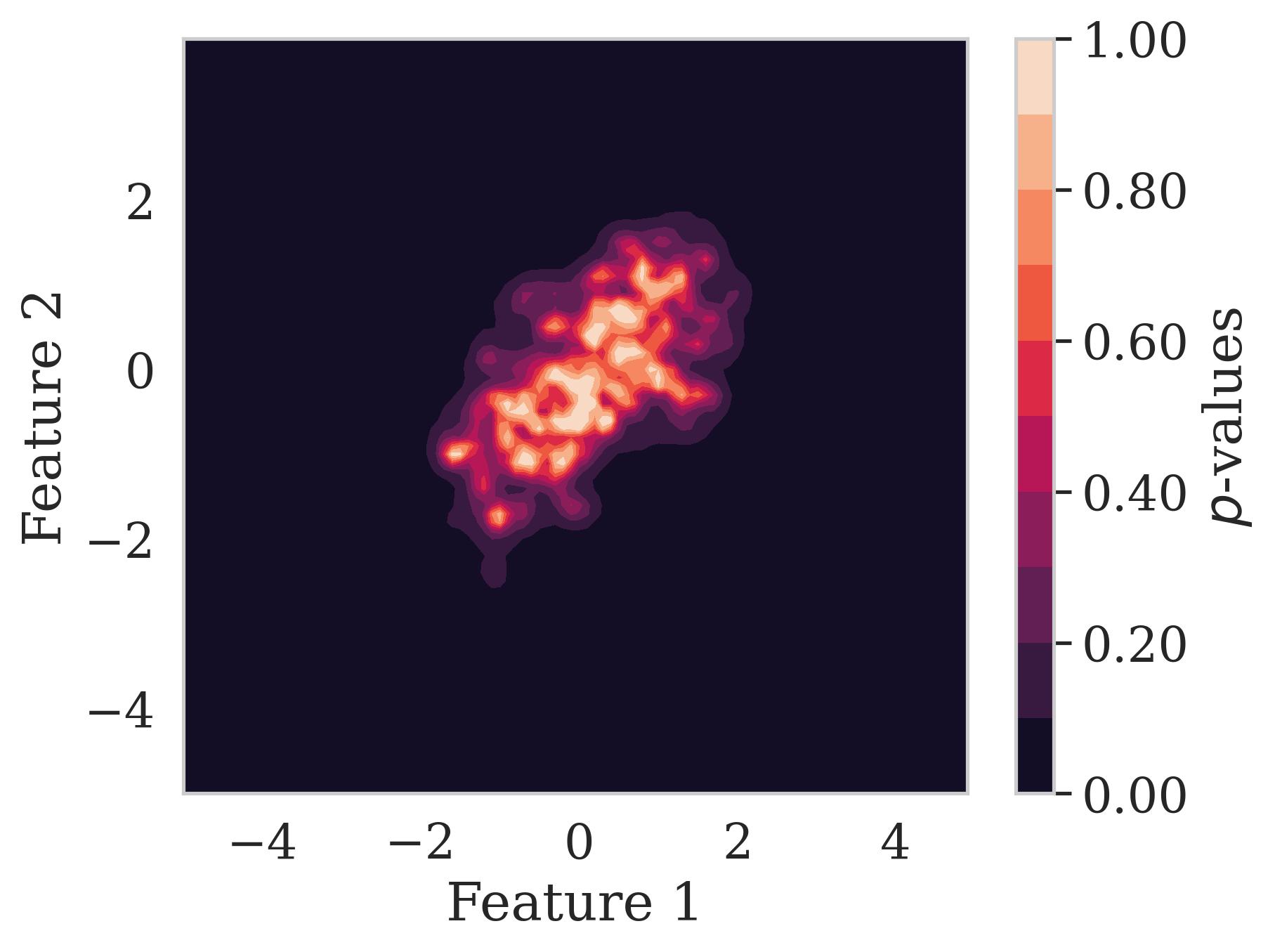}
                    \caption{$n = 1000$.}
                \end{subfigure}
                \caption{Effects of the original training sample count $n$ on the feature space's confidence regions. With the increase of $n$, our proposed algorithm could more precisely identify high-confidence regions ($p>\epsilon$). The narrower regions ensured that fewer unrepresentative samples were synthesised.}\label{fig:toySampleEffect}
            \end{figure}

            As with Deep Learning training, we assume that a larger training dataset will improve synthesis performance. Therefore, this section investigates the effect of the original training sample count $n$ on the feature space confidence regions. \Cref{fig:toySampleEffect} illustrates this relationship for class 0 on three subsets of Train$_{\text{orig}}$ with $n \in \Set{150, 300, 1000}$. The confidence regions of the feature space became visibly sharper with increasing $n$, narrowing around the original samples (\Cref{fig:toyOriginalDataset}). Assuming that narrower prediction sets reduce the inclusion of unrepresentative samples, we expect the performance of a model trained on the synthetically extended training sets to improve with increasing $n$.

\FloatBarrier{}
            \Cref{tab:toySampleEffect} presents the performance of models trained on the original and synthesised data with varying $n$ ($\epsilon=0.2$, $\gamma=0.1$). As expected with the insights from \Cref{fig:toySampleEffect}, the number of synthesised samples decreased as $n$ grew because the confidence regions became narrower. Turning our attention to the Train$_{\text{ext}}$ dataset's F$_1$-scores, each synthesis extension significantly improved results between 12--27 percentage points. In particular, the potential for improvement through synthesis was the highest for the smallest datasets $n \in \Set{150, 300}$ with as few as 15 minority-class samples in Train$_{\text{orig}}$ (\Cref{tab:toyDatasetSampleCounts}). Examining Train$_{\text{syn}}$ revealed the same trend, indicating the improvements were likely not due to more real data in the training set. The Wilcoxon test $p$-values $p_W$ in \Cref{tab:toySampleEffectWilcoxon} confirmed that the performance improvements related to the original training sample count $n$ were statistically significant.
            Note that this article synthesises all high-confidence grid points as synthetic samples. In future, further synthesis sampling techniques could additionally investigate the precise effects of varying synthesised sample counts on the performance \Cref{sec:discussion}).

            \begin{table}[!h]
                \centering
                \setlength{\tabcolsep}{3.95pt}
                \captionsetup{width=\textwidth}
                \caption{Mean and standard deviation prediction results with varying numbers of original training samples $n$. Larger values of $n$ resulted in fewer synthetic samples. The increased model performance indicates that the number of unrepresentative synthesised samples was especially reduced. The Wilcoxon test confirmed that all Train$_{\text{ext}}$ and Train$_{\text{syn}}$ results statistically improved on the original data ($p_W < 0.1$).}\label{tab:toySampleEffect}
                \begin{tabular}{rcrrrrrrr}  % chktex 44
                    \toprule
                    \multicolumn{2}{c}{\textbf{Samples}} & \multicolumn{1}{c}{\textbf{Train$_\mathbf{\text{orig}}$}} & \multicolumn{3}{c}{\textbf{Train$_\mathbf{\text{ext}}$}} & \multicolumn{3}{c}{\textbf{Train$_\mathbf{\text{syn}}$}} \\
                    \cmidrule(lr){1-2} \cmidrule(lr){3-3} \cmidrule(lr){4-6} \cmidrule(lr){7-9}
                    \multicolumn{1}{c}{\textbf{$n$}} & \multicolumn{1}{c}{\textbf{Syn.}} & \multicolumn{1}{c}{\textbf{F$_{\mathbf{1}}$-score}} & \multicolumn{1}{c}{\textbf{F$_{\mathbf{1}}$-score}} & \multicolumn{1}{c}{\textbf{Precision}} & \multicolumn{1}{c}{\textbf{Recall}} & \multicolumn{1}{c}{\textbf{F$_{\mathbf{1}}$-score}} & \multicolumn{1}{c}{\textbf{Precision}} & \multicolumn{1}{c}{\textbf{Recall}} \\
                    \midrule\midrule
                    \textbf{150} & 2,041 & 0.58 (.12) & 0.80 (.02) & 0.75 (.02) & 0.90 (.00) & 0.77 (.02) & 0.73 (.02) & 0.90 (.01) \\  % chktex 21
                    \textbf{300} & 1,350 & 0.63 (.22) & 0.90 (.00) & 0.89 (.01) & 0.91 (.00) & 0.90 (.00) & 0.89 (.01) & 0.91 (.00) \\  % chktex 21
                    \textbf{1,000} & 1,110 & 0.79 (.19) & 0.91 (.00) & 0.92 (.00) & 0.90 (.01) & 0.90 (.01) & 0.90 (.01) & 0.91 (.00) \\ % chktex 21
                    \bottomrule
                \end{tabular}
            \end{table}

            \begin{table}[!h]
                \centering
                \captionsetup{width=\textwidth}
                \caption{The Wilcoxon results $p_W$ comparing the F$_1$-scores of multiple test iterations on the Train$_\mathbf{\text{ext}}$ and Train$_\mathbf{\text{syn}}$ sets. In all cases, the $n$-driven improvements reported in \Cref{tab:toySampleEffect} were found to be statistically significant ($p_W < 0.1$), marked with *.}\label{tab:toySampleEffectWilcoxon}
                \begin{tabular}{rrrrrrr}
                    \toprule
                    \multicolumn{1}{l}{} & \multicolumn{3}{c}{\textbf{Train$_\mathbf{\text{ext}}$}} & \multicolumn{3}{c}{\textbf{Train$_\mathbf{\text{syn}}$}} \\
                    \cmidrule(lr){2-4} \cmidrule(lr){5-7}
                    \multicolumn{1}{c}{\textbf{$n$}} & \multicolumn{1}{c}{\textbf{150}} & \multicolumn{1}{c}{\textbf{300}} & \multicolumn{1}{c}{\textbf{1,000}} & \multicolumn{1}{c}{\textbf{150}} & \multicolumn{1}{c}{\textbf{300}} & \multicolumn{1}{c}{\textbf{1,000}} \\
                    \midrule\midrule
                    \textbf{150} & 1.00\:\: & 0.00* & 0.02* & 1.00\:\: & 0.00* & 0.00* \\
                    \textbf{300} & 0.00* & 1.00\:\: & 0.00* & 0.00* & 1.00\:\: & 0.09* \\
                    \textbf{1,000} & 0.02* & 0.00* & 1.00\:\: & 0.00* & 0.09* & 1.00\:\: \\
                    \bottomrule
                \end{tabular}
            \end{table}

        \subsubsection[Grid step]{Grid step $\gamma$}\label{sec:toyGridstepEffect}
\begin{figure}[!h]
                \centering
                \begin{subfigure}[b]{0.49\textwidth}
                    \centering
                    \includegraphics[width=\textwidth]{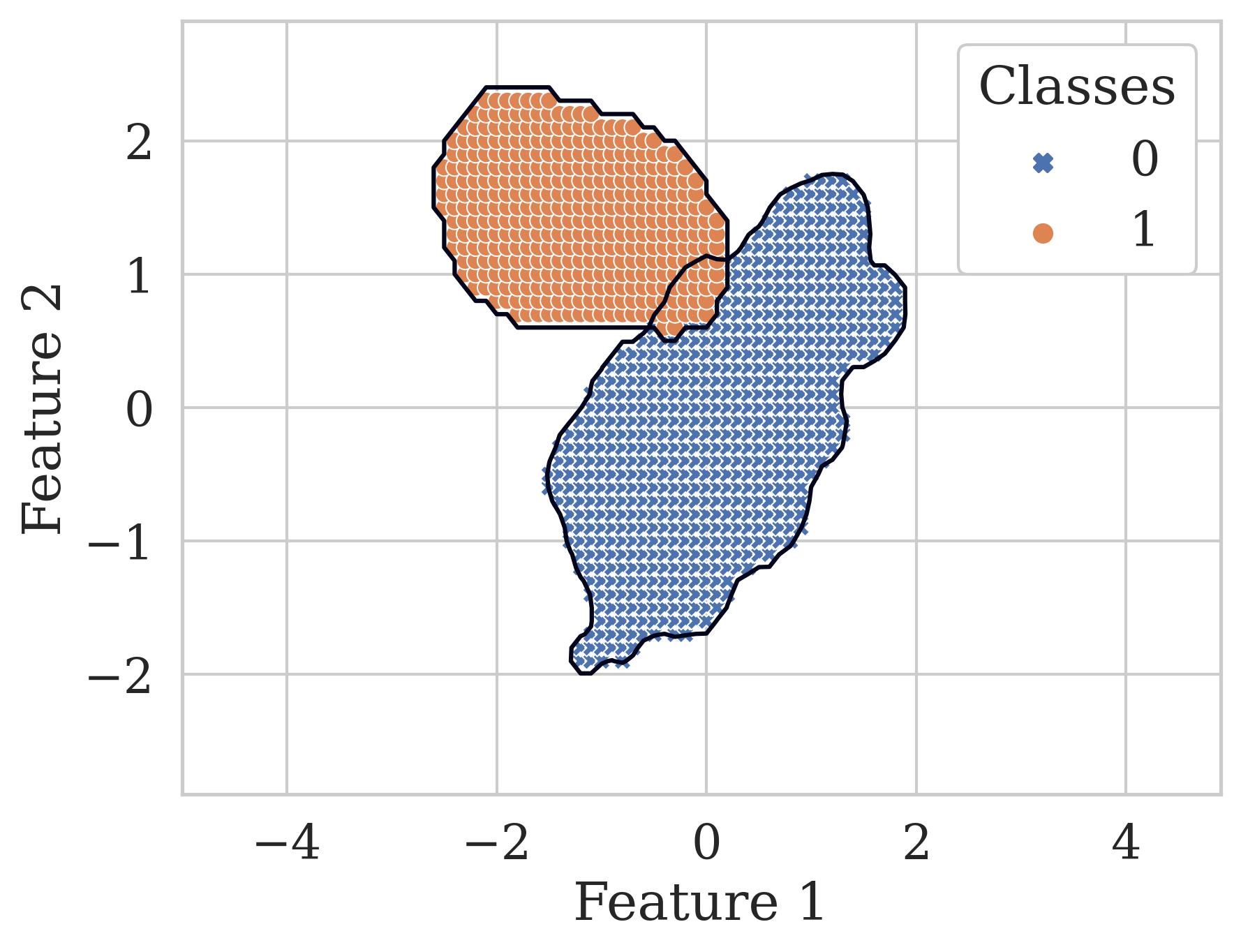}
                    \caption{$\gamma = 0.1$.}
                \end{subfigure}
                \hfill
                \begin{subfigure}[b]{0.49\textwidth}
                    \centering
                    \includegraphics[width=\textwidth]{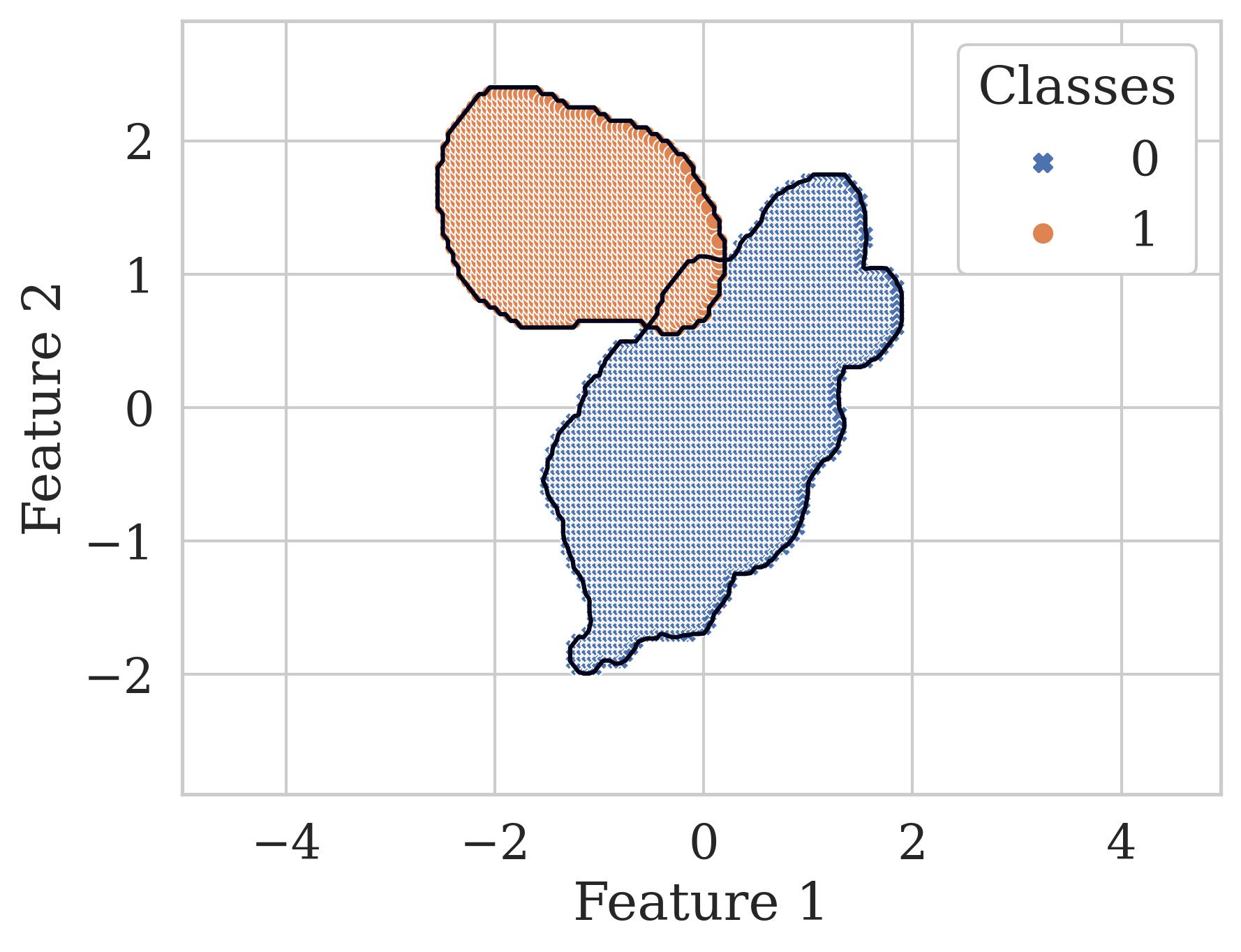}
                    \caption{$\gamma = 0.05$.}
                \end{subfigure}
                \caption{Effects of the grid step $\gamma$ on the feature space's resolution. The points in the graphs represent the grid points that fell into the outlined high-confidence regions ($\epsilon = 0.1$). A smaller $\gamma$ means a higher density of synthetic samples may be generated from the same high-confidence regions.}\label{fig:toyGridstepEffect}
            \end{figure}

            The grid step $\gamma$ defines the resolution of our feature space, bounded by the minima and maxima of the original dataset's features. Each grid point may be considered a potential synthetic sample with label-conditional $p$-values tested against the significance level $\epsilon$. Therefore, the smaller $\gamma$ is, the more synthetic data points will be sampled from our high-confidence feature space regions, as shown in~\Cref{fig:toyGridstepEffect}. A higher density of synthetic samples is desirable because Deep Learning models tend to generalise better with larger training sets. However, the trade-off is a higher required computing power, as each grid point must be evaluated per class. In areas where the classes' high-confidence regions overlap, both classes are assigned the same synthetic samples. Such behaviour would most likely occur for datasets with overlapping classes where the Bayes error, i.e., the probability of a perfect model making a prediction error~\cite{Salazar2023}, is non-zero.

\FloatBarrier{}

            \Cref{tab:toyGridStepEffect} shows these effects on our toy dataset. All extended training sets significantly improved on the original performance, increasing the F$_1$-score by up to 29 percentage points. Additionally to the improved generalisation, the reduction of the F$_1$-score's standard deviation by a factor of 10 indicates that the models were trained more robustly. Varying $\gamma \in \Set{0.1, 0.05, 0.01}$ while holding the other parameters steady ($\epsilon=0.1$, $n=150$), we found that the number of synthetic samples generated from the exact same high-confidence regions significantly increased with a decrease in the grid step. Consequently, the same Deep Learning model's generalisation performance increased by up to 9 percentage points: from F$_1$ = 78\% when trained on the dataset extended with $\gamma=0.1$, to F$_1$ = 87\% when trained on the dataset extended with $\gamma=0.01$. In particular, recall scores were significantly improved due to the stronger representation of the minority class with an increased sample count. Investigating the Wilcoxon $p$-values $p_W$ confirmed that the $\gamma$-induced improvements were statistically significant for all combinations.

            \begin{table}[!h]
                \centering
                \setlength{\tabcolsep}{4.8pt}
                \captionsetup{width=\textwidth}
                \caption{Mean and standard deviation prediction results with varying grid step $\gamma$. The smaller $\gamma$ was, the more synthetic samples were generated. Consequently, the models' generalisation was improved. The Wilcoxon test confirmed that the $\gamma$-related improvements were significant in all cases ($p_W<0.1$), marked with *.}\label{tab:toyGridStepEffect}
                \begin{tabular}{rrrrrr|rrrr}  % chktex 44
                    \toprule
                     & \multicolumn{1}{c}{\textbf{Samples}} & \multicolumn{1}{c}{\textbf{Train$_{\mathbf{\text{orig}}}$}} & \multicolumn{3}{c}{\textbf{Train$_{\mathbf{\text{ext}}}$}} & \multicolumn{4}{|c}{\textbf{$\mathbf{p_{W}}$   (F$_{\mathbf{1}}$-score)}} \\
                    \cmidrule(lr){2-2} \cmidrule(lr){3-3} \cmidrule(lr){4-6} \cmidrule(lr){7-10}
                    \multicolumn{1}{c}{\textbf{$\gamma$}} & \multicolumn{1}{c}{\textbf{Ext.}} & \multicolumn{1}{c}{\textbf{F$_{\mathbf{1}}$-score}} & \multicolumn{1}{c}{\textbf{F$_{\mathbf{1}}$-score}} & \multicolumn{1}{c}{\textbf{Precision}} & \multicolumn{1}{c}{\textbf{Recall}} & \multicolumn{1}{|c}{\textbf{$\mathbf{\gamma}$}} & \multicolumn{1}{c}{\textbf{0.1}} & \multicolumn{1}{c}{\textbf{0.05}} & \multicolumn{1}{c}{\textbf{0.01}} \\
                    \midrule\midrule
                    \textbf{0.1} & 2,580 & \multirow{3}{*}{0.58 (.12)} & 0.78 (.03) & 0.74 (.02) & 0.89 (.01) & \textbf{0.1} & 1.00\:\: & 0.00* & 0.00* \\  % chktex 21
                    \textbf{0.05} & 10,330 &  & 0.84 (.01) & 0.80 (.01) & 0.89 (.00) & \textbf{0.05} & 0.00* & 1.00\:\: & 0.00* \\  % chktex 21
                    \textbf{0.01} & 258,397 &  & 0.87 (.01) & 0.87 (.02) & 0.87 (.01) & \textbf{0.01} & 0.00* & 0.00* & 1.00\:\: \\  % chktex 21
                    \bottomrule
                \end{tabular}
            \end{table}

        \subsubsection{Bringing everything together}
            After investigating the three parameters of our conformal synthesis algorithm individually, this section illustrates the interactions between the significance $\epsilon$, the number of original training samples $n$, and the grid step $\gamma$. All three influence the core of our algorithm by widening or narrowing the confidence regions in their own way. Consequently, the parameter values and the underlying original dataset define the extended training set Train$_{\text{ext}}$.

\begin{figure}[!h]
                \centering
                \begin{subfigure}[b]{0.325\textwidth}
                    \centering
                    \includegraphics[width=\textwidth]{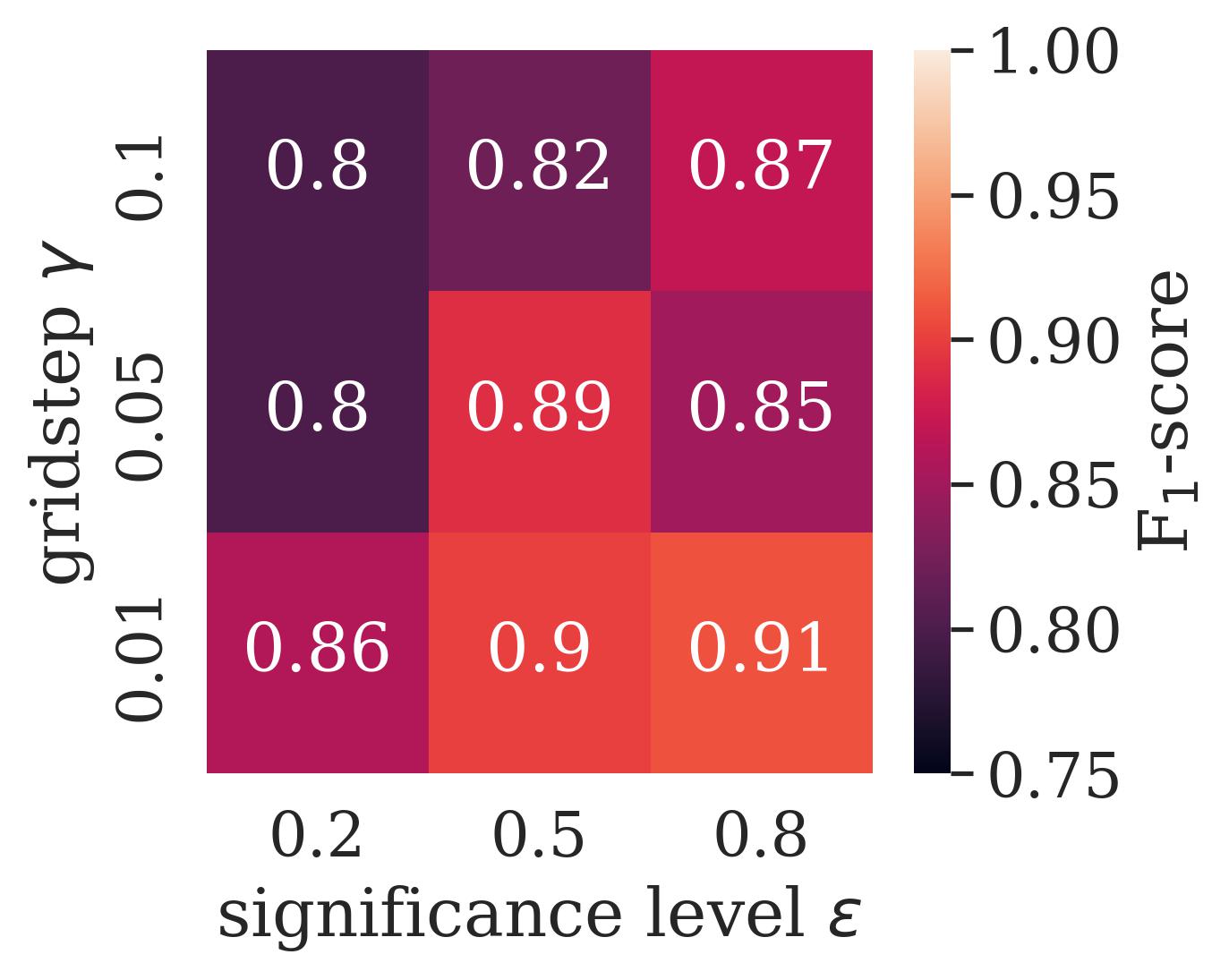}
                    \caption{$n = 150$.}
                \end{subfigure}
                \hfill
                \begin{subfigure}[b]{0.325\textwidth}
                    \centering
                    \includegraphics[width=\textwidth]{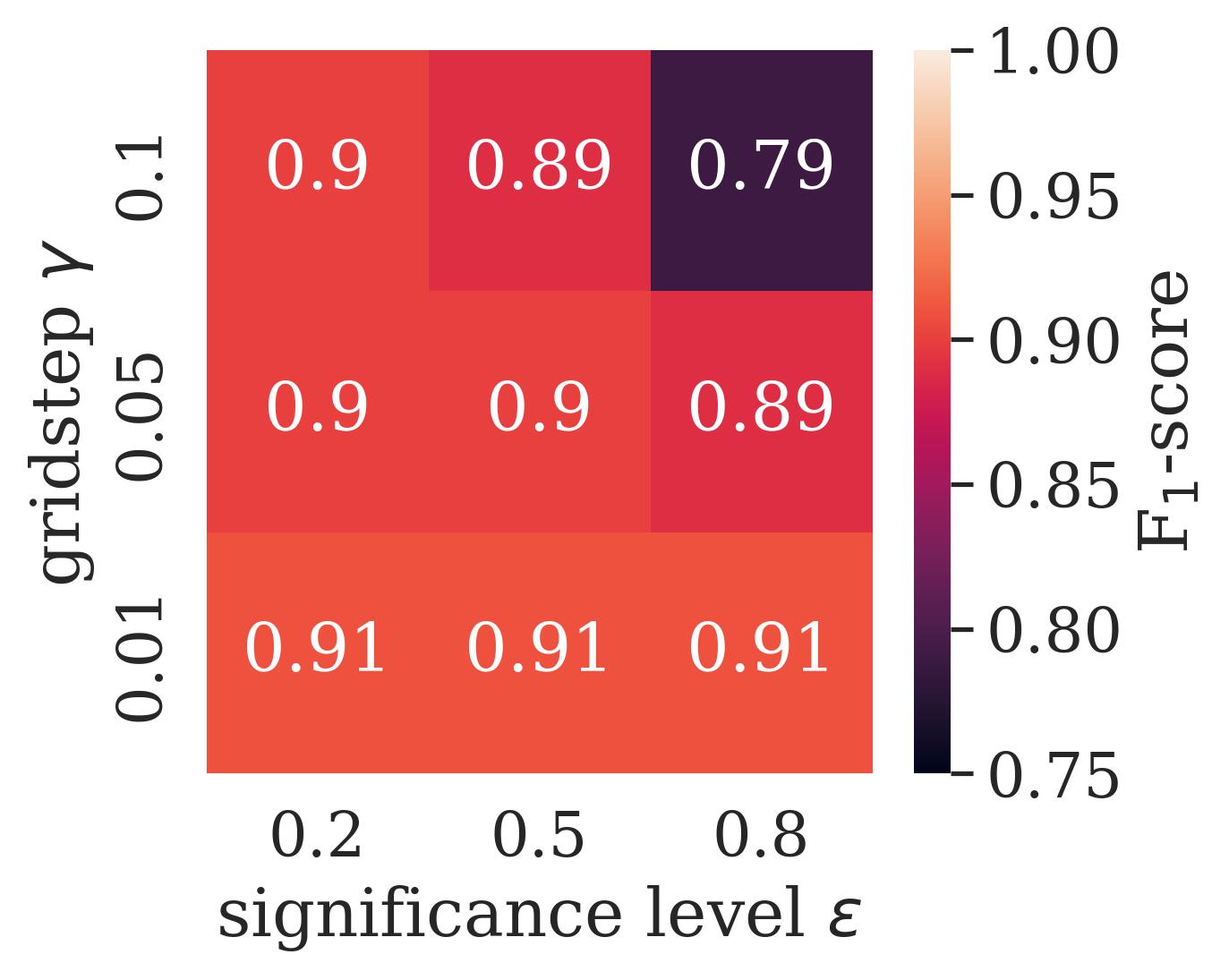}
                    \caption{$n = 300$.}
                \end{subfigure}
                \hfill
                \begin{subfigure}[b]{0.325\textwidth}
                    \centering
                    \includegraphics[width=\textwidth]{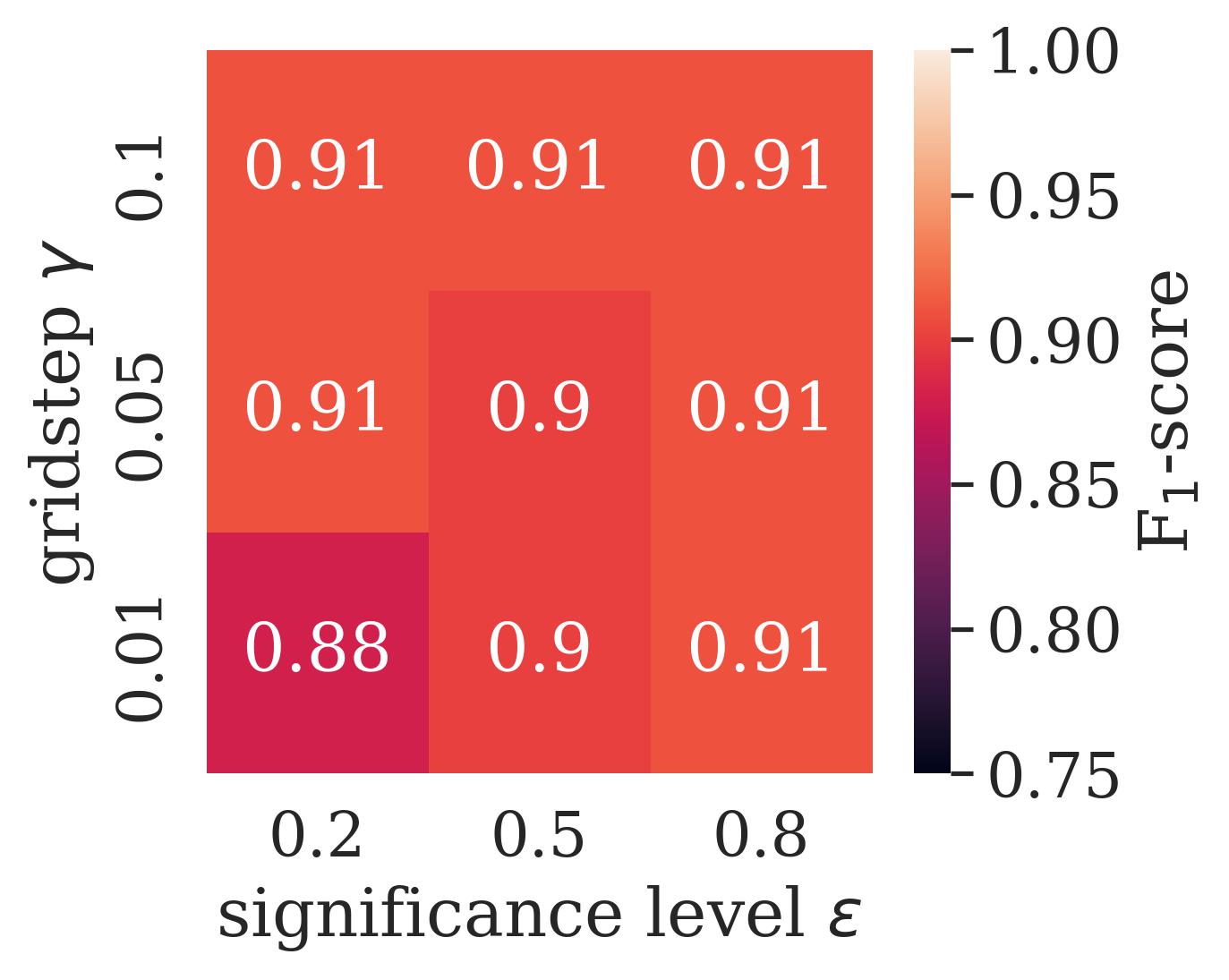}
                    \caption{$n = 1000$.}
                \end{subfigure}
                \caption{Overview of Deep Learning performance on the extended training sets Train$_{\text{ext}}$, synthesised with varying parameters. Each square represents the model's mean F$_1$-score on the same original test set. Results tended to improve with larger significance levels $\epsilon$, smaller grid steps $\gamma$, and larger original training sample counts $n$.}\label{fig:toyParameterHeatmap}
            \end{figure}

            \Cref{fig:toyParameterHeatmap} shows a heat map of the mean F$_1$-scores achieved by a Deep Learning model trained on the extended dataset Train$_{\text{ext}}$ five times. Interestingly, we observed a pattern with the changing parameters. The results tended to increase with the following:
            \begin{itemize}
                \item A larger significance level $\epsilon$, which tightened the high-confidence feature space regions;
                \item A smaller grid step $\gamma$, leading to a higher resolution of the feature space;
                \item And a larger number of original training points $n$, allowing for more precise modelling of the feature space confidence.
            \end{itemize}
            Note that these trends are subject to randomness during the models' training and are not guaranteed results.
            The conformal performance characteristics discussed in \Cref{sec:proposal} apply only to the synthesis process, not the downstream model prediction performances. However, \Cref{fig:toyParameterHeatmap} indicates that the relationship between model performance and $\epsilon$ are closely related: prediction performances tended to improve in parallel, subject to randomness in the model's generalisation. A notable exception is the F$_1$-score on data synthesised with $n=300$, $\gamma=0.1$, and $\epsilon=0.8$. The significantly reduced model performance is likely related to an insufficient number of synthetic samples being generated as a consequence of narrow confidence regions (caused by high $\epsilon$) and a low resolution of the feature space (caused by low $\gamma$).

            Evaluating the performance results compared to the Train$_{\text{orig}}$ baseline results in \Cref{tab:toyOverallResults} revealed that the majority of extended datasets strongly increased model performance by 9--33 percentage points F$_1$-score across all investigated synthesis parameter settings. The standard deviation was also significantly decreased, indicating the synthetic samples enabled the Deep Learning models to generalise the dataset more robustly. This insight was further supported by the Wilcoxon test, which revealed that the improvements compared to the baseline were statistically significant. Only three models narrowly did not pass the test ($p_W < 0.1$), all occurring at $n=1000$. These results can be traced back to the insight that the potential for dataset improvement through synthesis is highest for small datasets (\Cref{sec:toyDatasetOriginalSampleCount}).

            \begin{table}[!h]
                \centering
                \setlength{\tabcolsep}{6.8pt}
                \captionsetup{width=\textwidth}
                \caption{Mean and standard deviation F$_1$-score results with varying conformal synthesis parameters. The best results per category are highlighted in bold. The Wilcoxon test confirmed that the improvements were largely statistically significant ($p_W < 0.1$), marked with *. The improvement potential of conformal synthesis is largest for small datasets, explaining the small number of exceptions with $n=1000$.}\label{tab:toyOverallResults}
                \begin{tabular}{rrrrrrrrr}
                    \toprule
                    & \multicolumn{1}{c}{} & \multicolumn{1}{c}{\textbf{}} & \multicolumn{2}{c}{\textbf{$\mathbf{\epsilon=0.2}$}} & \multicolumn{2}{c}{\textbf{$\mathbf{\epsilon=0.5}$}} & \multicolumn{2}{c}{\textbf{$\mathbf{\epsilon=0.8}$}} \\
                    \cmidrule(lr){4-5} \cmidrule(lr){6-7} \cmidrule(lr){8-9}
                    \multicolumn{1}{c}{\textbf{$n$}} & \multicolumn{1}{c}{\textbf{$\mathbf{\gamma}$}} & \multicolumn{1}{c}{\textbf{Train$_\mathbf{\text{orig}}$}} & \multicolumn{1}{c}{\textbf{Train$_\mathbf{\text{ext}}$}} & \multicolumn{1}{c}{\textbf{$\mathbf{p_W}$}} & \multicolumn{1}{c}{\textbf{Train$_\mathbf{\text{ext}}$}} & \multicolumn{1}{c}{\textbf{$\mathbf{p_W}$}} & \multicolumn{1}{c}{\textbf{Train$_\mathbf{\text{ext}}$}} & \multicolumn{1}{c}{\textbf{$\mathbf{p_W}$}} \\
                    \midrule\midrule
                    \multirow{3}{*}{\textbf{150}} & \textbf{0.1} & \multirow{3}{*}{0.58 (.12)} & 0.80 (.02) & 0.00* & 0.82 (.04) & 0.00* & 0.87 (.06) & 0.00* \\
                     & \textbf{0.05} &  & 0.80 (.03) & 0.00* & 0.89 (.01) & 0.00* & 0.85 (.06) & 0.00* \\
                     & \textbf{0.01} &  & \textbf{0.86} (.00) & 0.00* & \textbf{0.90} (.00) & 0.00* & \textbf{0.91} (.00) & 0.00* \\
                    \midrule
                    \multirow{3}{*}{\textbf{300}} & \textbf{0.1} & \multirow{3}{*}{0.63 (.22)} & 0.90 (.00) & 0.02* & 0.89 (.02) & 0.03* & 0.79 (.15) & 0.02* \\
                     & \textbf{0.05} &  & 0.90 (.00) & 0.02* & 0.90 (.01) & 0.02* & 0.89 (.02) & 0.02* \\
                     & \textbf{0.01} &  & \textbf{0.91} (.01) & 0.02* & \textbf{0.91} (.00) & 0.02* & \textbf{0.91} (.00) & 0.02* \\
                    \midrule
                    \multirow{3}{*}{\textbf{1,000}} & \textbf{0.1} & \multirow{3}{*}{0.79 (.18)} & \textbf{0.91} (.00) & 0.08* & \textbf{0.91} (.00) & 0.09* & 0.91 (.01) & 0.09* \\
                     & \textbf{0.05} &  & \textbf{0.91} (.00) & 0.11\:\: & 0.90 (.01) & 0.13\:\: & \textbf{0.91} (.00) & 0.07* \\  % chktex 21
                     & \textbf{0.01} &  & 0.88 (.03) & 0.09* & 0.90 (.00) & 0.11\:\: & \textbf{0.91} (.00) & 0.08* \\  % chktex 21
                    \bottomrule
                \end{tabular}
            \end{table}

            Because the synthesis algorithm does not guarantee the models' prediction performance, the way the original dataset is split may impact generalisation. Therefore, additional random splits of the original samples into the test and Train$_{\text{orig}}$ datasets were evaluated, which underpin all following subsets (e.g., the proper training and calibration set for synthesis). \Cref{tab:toyDataSplitSignificance} presents the Wilcoxon test $p$-values comparing pairs of model performances quantified by the F$_1$-scores. Three new data splits were compared against the results achieved on the original data split in turn. For the majority of additional data splits, we achieved the desired result of failing to reject the H$_0$ hypothesis, indicating the Deep Learning results were statistically similar. Therefore, we may assume that the data split had relatively little impact on the synthesis process and the subsequent model training.
            Now that we have systematically and comprehensively evaluated a simple toy dataset, we turn to real-world datasets to verify our proposed conformal synthesis algorithm. The insights gained in this section informed the following algorithm parameter choices.

            {\renewcommand{\arraystretch}{0.95}
            \begin{table}[!h]
                \centering
                \captionsetup{width=\textwidth}
                \caption{Wilcoxon test results evaluating the effect of the data split on the models' F$_1$-scores. Three random data splits of the original samples were evaluated against the previous results (\Cref{tab:toyOverallResults}) in turn, with the range of $p$-values reported. The desirable outcome to consistently fail the test $p_W < 0.1$ was reached in the majority of cases, marked with *. Therefore, we conclude that conformal synthesis and subsequent model performances were largely invariant to the data split.}\label{tab:toyDataSplitSignificance}
                \begin{tabular}{crrrrr}
                    \toprule
                    & \multicolumn{1}{c}{} & \multicolumn{1}{c}{\textbf{Train$_\mathbf{\text{orig}}$}} & \multicolumn{3}{c}{\textbf{Train$_\mathbf{\text{ext}}$}} \\
                    \cmidrule(lr){4-6}
                    \textbf{$n$} & \multicolumn{1}{c}{\textbf{$\mathbf{\gamma}$}} & \multicolumn{1}{l}{} & \multicolumn{1}{c}{\textbf{$\mathbf{\epsilon=0.2}$}} & \multicolumn{1}{c}{\textbf{$\mathbf{\epsilon=0.5}$}} & \multicolumn{1}{c}{\textbf{$\mathbf{\epsilon=0.8}$}} \\
                    \midrule\midrule
                    \multirow{3}{*}{\textbf{150}} & \textbf{0.1} & \multirow{3}{*}{0.61--0.88*} & 0.08--0.19\:\: & 0.16--0.30* & 0.16--0.82* \\  % chktex 21
                     & \textbf{0.05} &  & 0.19--0.45* & 0.14--0.20* & 0.25--0.34* \\
                     & \textbf{0.01} &  & 0.12--0.36* & 0.35--0.61* & 0.54--1.00* \\
                    \midrule
                    \multirow{3}{*}{\textbf{300}} & \textbf{0.1} & \multirow{3}{*}{0.93--0.97*} & 0.14* & 0.36--0.55* & 0.14--0.38* \\
                     & \textbf{0.05} &  & 1.00* & 0.09--1.00\:\: & 0.17--0.19* \\  % chktex 21
                     & \textbf{0.01} &  & 0.09--0.27\:\: & 0.24--0.54* & 0.24--0.35* \\  % chktex 21
                    \midrule
                    \multirow{3}{*}{\textbf{1,000}} & \textbf{0.1} & \multirow{3}{*}{0.89--0.96*} & 0.10--0.27\:\: & 0.14--1.00* & 0.11--0.35* \\  % chktex 21
                     & \textbf{0.05} &  & 0.29--1.00* & 0.14--1.00* & 0.20* \\
                     & \textbf{0.01} &  & 0.24--0.57* & 0.13--0.24* & 0.09--0.24\:\: \\  % chktex 21
                    \bottomrule
                \end{tabular}
            \end{table}}

    \subsection{Evaluating real-world datasets}\label{sec:realDatasets}
        To assess the practical benefits of our conformal synthesis algorithm, we tested its application to four realistic benchmark datasets. The datasets were carefully selected to showcase our algorithm's range on some of the most prevalent data challenges limiting Machine Learning implementations: Low sample count, class imbalance and overlap, and data privacy.

        Following the details laid out in \Cref{sec:experimentalSetup}, we measured the success of our algorithm by comparing the prediction results of a feedforward neural network trained on the original data Train$_{\text{orig}}$ and our extended data Train$_{\text{ext}}$ on the same test set. The optimal parameters for data synthesis were identified with the insights gathered in \Cref{sec:toyDataset}. Note that we used UMAP~\cite{McInnes2018} to reduce all datasets' dimensions to two features to improve the computational complexity of synthesis, discussed in depth in \Cref{sec:discussion}.

        \subsubsection{Small dataset}\label{sec:smallDataResults}
\begin{figure}[!ht]
                \centering
                \begin{subfigure}[b]{0.8\textwidth}
                    \centering
                    \includegraphics[width=\textwidth]{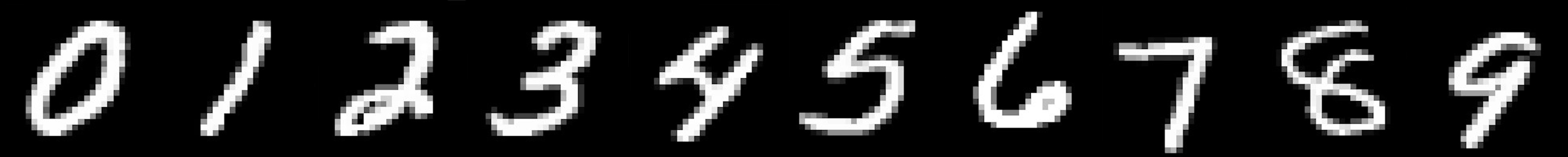}
                \caption{Handwritten digit samples.}\label{fig:mnistImages}
                \end{subfigure}
                \qquad
                \begin{subfigure}[b]{0.495\textwidth}
                    \centering
                    \includegraphics[width=\textwidth]{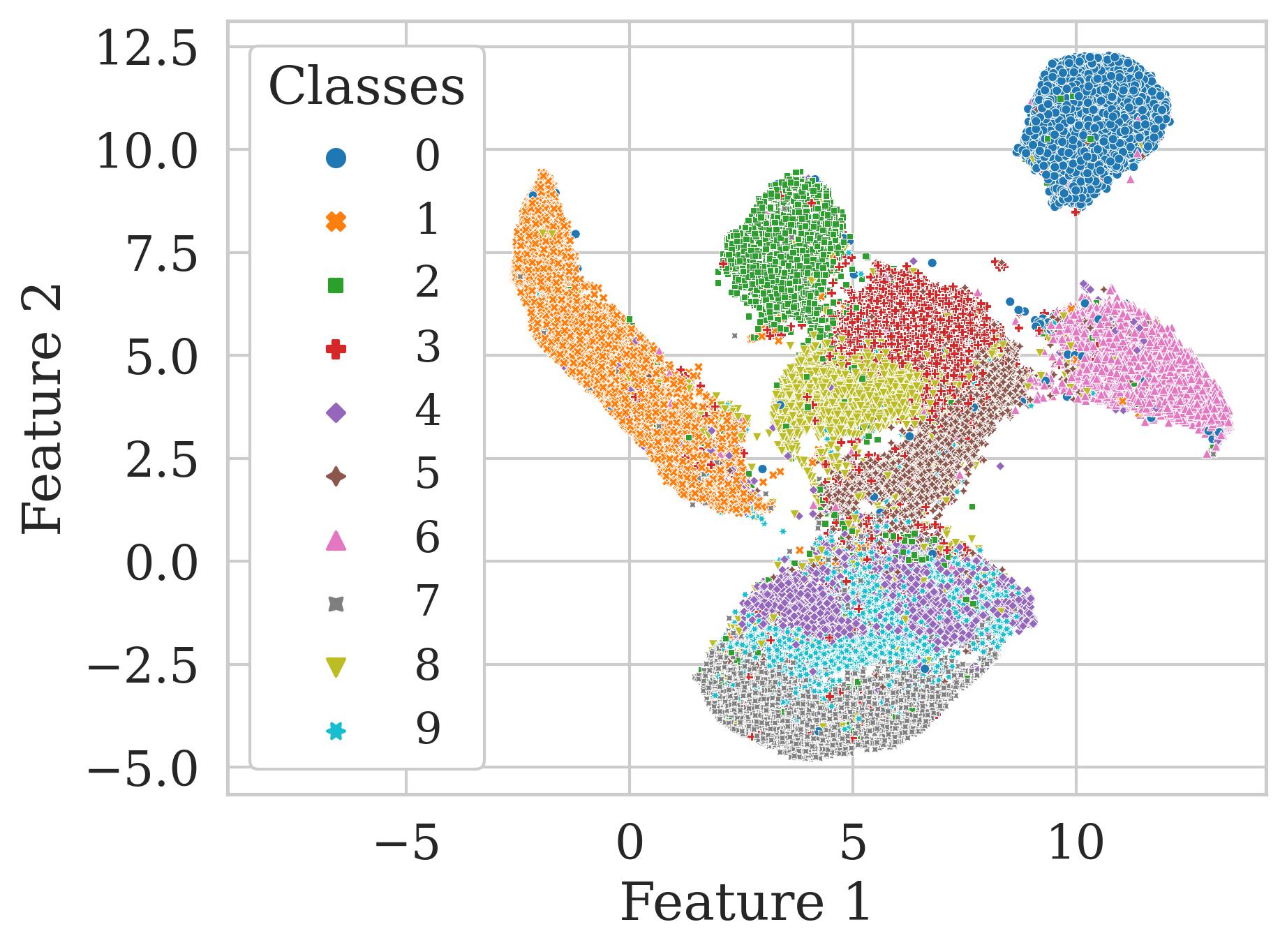}
                    \caption{Pre-processed samples.}\label{fig:mnistSamples}
                \end{subfigure}
                \hfill
                \begin{subfigure}[b]{0.495\textwidth}
                    \centering
                    \includegraphics[width=\textwidth]{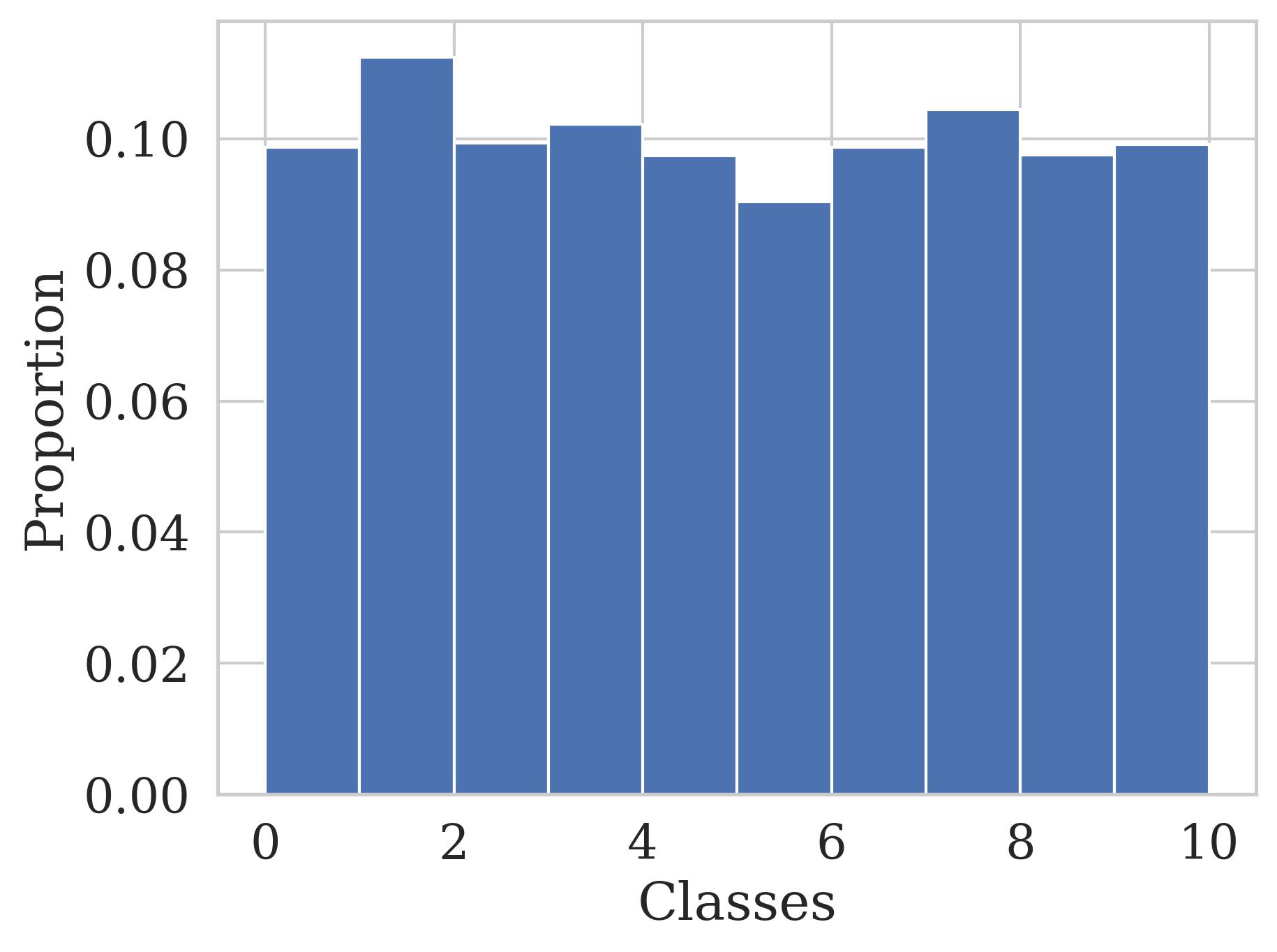}
                    \caption{Class distribution.}\label{fig:mnistClassDistribution}
                \end{subfigure}
                \caption{The MNIST dataset of handwritten digits. Each image was reduced to two dimensions with UMAP before model training. Most classes were adjacent with some overlap, making class separation challenging. The relative class distributions were maintained in all data subsets.}\label{fig:mnistDataset}
            \end{figure}

            Low sample counts are a ubiquitous challenge for Deep Learning, often caused by the difficulties and costs of data collection. We used the popular MNIST dataset~\cite{Lecun1998} to simulate a low training sample count. With 70,000 original samples, we could conduct an in-depth investigation of our algorithm's benefits across a range of dataset sizes. 10,000 samples were held back to evaluate the model performances on the test set. \Cref{fig:mnistDataset} visualises the dataset, including a representative selection of handwritten digit samples (0--9). Each 28$\times$28 pixel image was scaled to $(0, 1)$ and reduced to two dimensions before synthesis. \Cref{fig:mnistSamples} shows the samples' distribution in the feature space after pre-processing. While some classes were clearly linearly separated, most were adjacent with limited overlap. Overall, the classes were roughly equally distributed and this distribution was maintained for each data subset (\Cref{fig:mnistClassDistribution}). 

\begin{figure}[!ht]
                \centering
                \includegraphics[width=0.495\textwidth]{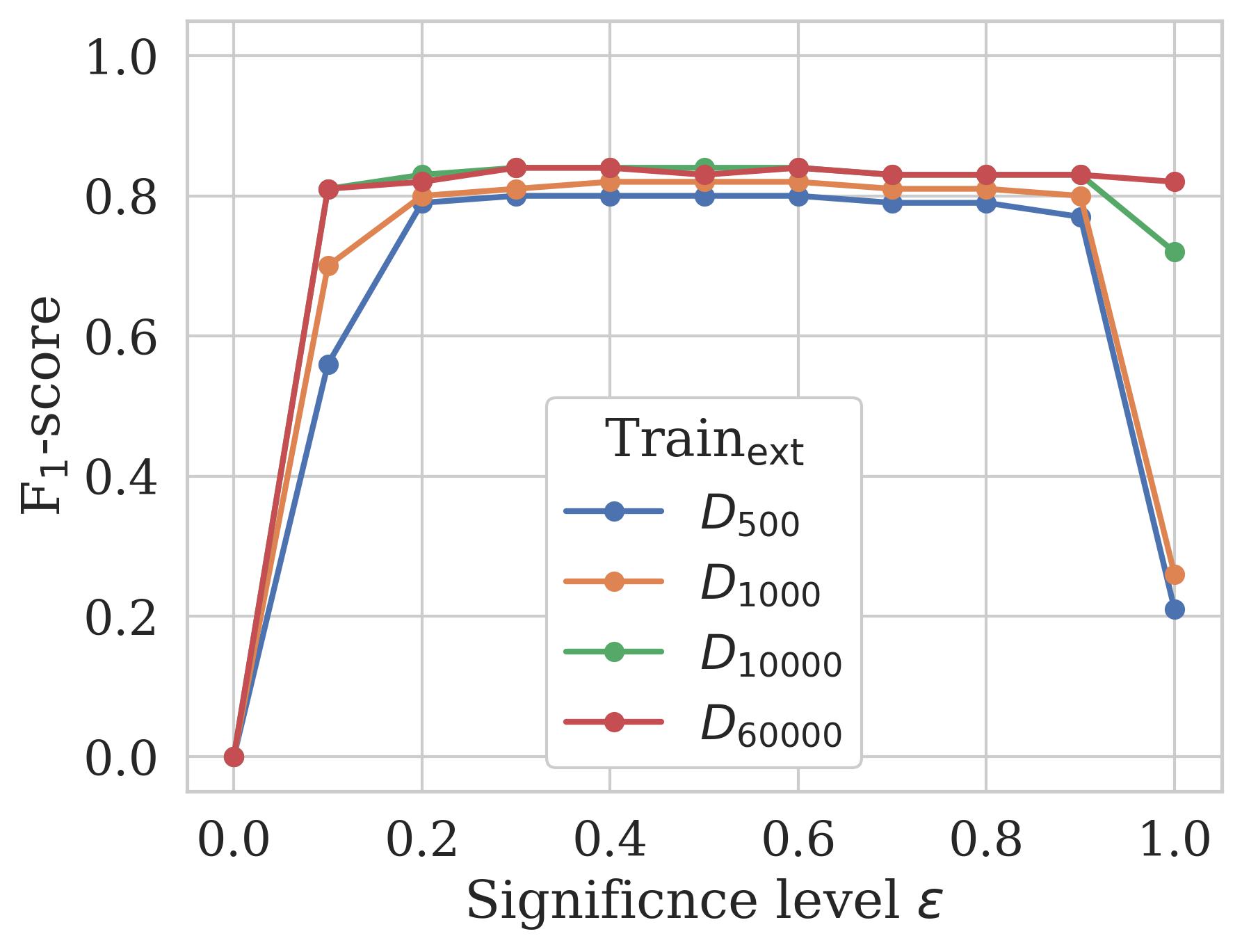}
                \caption{Mean model performances on the extended datasets synthesised from Train$_{\text{orig}}$ subsets with 500, 1000, 10000, and 600000 samples. The larger the original training dataset was, the faster the performance increased, allowing us to identify lower $\epsilon$ values with the elbow method ($\epsilon=0.2$ for the two smaller subsets and $\epsilon=0.1$ for the two larger datasets).}\label{fig:mnistEpsilonSelection}
            \end{figure}

            The core purpose of our proposed algorithm is to support Deep Learning generalisation by extending datasets with synthetic training samples. To showcase our approach's benefits across varying set sizes, we artificially under-sampled Train$_{\text{orig}}$ creating four subsets $D_n$ with $n \in \Set{500, 1000, 10000, 60000}$ training samples. \Cref{fig:mnistEpsilonSelection} reveals that all four subsets followed the expected model performance pattern for different $\epsilon$ when trained on the corresponding extended Train$_{\text{ext}}$ sets ($\gamma=0.01$): A sharp increase, followed by a plateau and then converging with the baseline performance on Train$_{\text{orig}}$. The larger the original training set was, the more quickly the performance increased. We employ the elbow method to select the optimal $\epsilon$, identifying $\epsilon=0.2$ for $D_{500}$ and $D_{1000}$, and $\epsilon=0.1$ for $D_{10000}$ and $D_{60000}$.

            Generally, the more original training points were available, the smaller the number of synthesised samples was, caused by more precise (and therefore more narrow) confidence regions. To illustrate the relationship between $\epsilon$, the original training set size, and the number of synthesised samples, \Cref{tab:mnistSamples} reports the exact sample counts. In particular, we investigated the optimal $\epsilon$ for each training subset as well as $\epsilon=0.9$, since models performed roughly as well on these Train$_{\text{ext}}$ sets as for their optimal $\epsilon$ (\Cref{fig:mnistEpsilonSelection}). Notably, the number of synthesised samples decreased as the number of original samples $n$ and the significance level $\epsilon$ increased. Intuitively, increasing both parameters caused the confidence regions to narrow and surround the original training samples more precisely, reducing the feature space area from which new samples are synthesised. $\epsilon$ was the single most influential factor, reducing synthesised sample counts by a factor of up to around 20, going from $\epsilon=0.1$ to $\epsilon=0.9$.

            \begin{table}[!h]
                \captionsetup{width=\textwidth}
                \caption{Original and synthesised sample counts. The larger the original dataset was, the narrower and more precise the confidence regions were, and the fewer new sample points were synthesised. Note that the number of synthesised samples also depended on the significance level. Therefore, the comparison is most clear when comparing variations of $n$ for the same $\epsilon$.}\label{tab:mnistSamples}
                \begin{tabular}{ccrrr|rrr}  % chktex 44
                    \toprule
                    \multicolumn{1}{c}{\textbf{Subset}} & \textbf{Test} & \multicolumn{3}{c}{\textbf{Train$\mathbf{_\text{orig}}$}} & \multicolumn{1}{|c}{} & \multicolumn{1}{c}{\textbf{Train$\mathbf{_\text{syn}}$}} & \multicolumn{1}{c}{\textbf{Train$\mathbf{_\text{ext}}$}} \\
                    \cmidrule(lr){2-2} \cmidrule(rr){3-5} \cmidrule(lr){7-7} \cmidrule(lr){8-8}
                    \multicolumn{1}{c}{} & \multicolumn{1}{c}{\textbf{All}} & \multicolumn{1}{c}{\textbf{Prop.}} & \multicolumn{1}{c}{\textbf{Calib.}} & \multicolumn{1}{c}{\textbf{All ($n$)}} & \multicolumn{1}{|c}{\textbf{$\mathbf{\epsilon}$}} & \multicolumn{1}{c}{\textbf{All}} & \multicolumn{1}{c}{\textbf{All}} \\
                    \midrule\midrule
                    \multirow{2}{*}{\textbf{$\mathbf{D_{500}}$}} & \multirow{2}{*}{10,000} & \multirow{2}{*}{300} & \multirow{2}{*}{200} & \multirow{2}{*}{500} & \textbf{0.2} & \multicolumn{1}{r}{1,213,241} & 1,213,741 \\
                     &  &  &  &  & \textbf{0.9} & 126,190 & 126,690 \\
                    \multirow{2}{*}{\textbf{$\mathbf{D_{1000}}$}} & \multirow{2}{*}{10,000} & \multirow{2}{*}{600} & \multirow{2}{*}{400} & \multirow{2}{*}{1,000} & \textbf{0.2} & \multicolumn{1}{r}{1,028,384} & 1,029,384 \\
                     &  &  &  &  & \textbf{0.9} & 75,067 & 76,067 \\
                    % \midrule
                    \multirow{2}{*}{\textbf{$\mathbf{D_{10000}}$}} & \multirow{2}{*}{10,000} & \multirow{2}{*}{6,000} & \multirow{2}{*}{4,000} & \multirow{2}{*}{10,000} & \textbf{0.1} & \multicolumn{1}{r}{1,287,020} & 1,297,020 \\
                     &  &  &  &  & \textbf{0.9} & 49,908 & 59,908 \\
                    % \midrule
                    \multirow{2}{*}{\textbf{$\mathbf{D_{60000}}$}} & \multirow{2}{*}{10,000} & \multirow{2}{*}{36,000} & \multirow{2}{*}{24,000} & \multirow{2}{*}{60,000} & \textbf{0.1} & \multicolumn{1}{r}{1,128,440} & 1,188,440 \\
                     &  &  &  &  & \textbf{0.9} & 44,242 & 104,242 \\
                    \bottomrule
                \end{tabular}
            \end{table}

            Investigating this aspect further in \Cref{tab:mnistResults}, we note that in many cases, training sets extended with samples synthesised with optimal low $\epsilon$ and $\epsilon=0.9$ confidence thresholds performed very similarly on F$_1$-score, precision, and recall. While we prioritised low $\epsilon$ to reduce the synthesis error in this article, a different heuristic may be to select the highest possible $\epsilon$ with the elbow method to reduce the number of required synthesis samples, constructing the most efficient extended training set to maximise model performance (discussed further in \Cref{sec:discussion}). Comparing the performance against the baseline, models trained on Train$_{\text{orig}}$ performed very poorly on low training sample counts as expected (under 30\% F$_1$-score for D$_{500}$ and D$_{1000}$). The score increased up to 82\% with the maximum number of original training samples (D$_{60000}$). In contrast, our synthetic dataset extension surpassed that performance from 10,000 original training samples. With only 500 original samples, our algorithm synthesised samples that improved model performance from 21\% to 79\% F$_1$-score. While the rate of improvement decreased with increasing training samples, our extended Train$_\text{ext}$ consistently outperformed Train$_{\text{orig}}$, reaching a maximum of 83\% on the full dataset. Because the classes were roughly equally balanced in the original dataset, this increase was caused by precision and recall improving in equal measures across the classes.

           \begin{table}[!ht]
                \centering
                \setlength{\tabcolsep}{2.1pt}
                \captionsetup{width=\textwidth}
                \caption{Deep Learning results on the MNIST dataset, investigating the impact of reducing the size of samples in Train$_{\text{orig}}$ before synthesis and model training. The mean and standard deviation of five trials are reported. As the number of original samples increased, so did the models' performance, although at different rates. The precision and recall scores stayed level with each other, implying that synthesis successfully maintained the class' original balance. The Wilcoxon test confirmed that the improvements were statistically significant ($p_W < 0.1$).}\label{tab:mnistResults}
                \begin{tabular}{ccrrrrrrrrr}
                    \toprule
                     &  & \multicolumn{3}{c}{\textbf{F$_{\mathbf{1}}$-score}} & \multicolumn{3}{c}{\textbf{Precision}} & \multicolumn{3}{c}{\textbf{Recall}} \\
                     \cmidrule(lr){3-5} \cmidrule(lr){6-8} \cmidrule(lr){9-11}
                     & \textbf{$\mathbf{\epsilon}$} & \multicolumn{1}{c}{\textbf{Train$_\text{orig}$}} & \multicolumn{1}{c}{\textbf{Train$_\text{ext}$}} & \multicolumn{1}{c}{\textbf{$\mathbf{p_W}$}} & \multicolumn{1}{c}{\textbf{Train$_\text{orig}$}} & \multicolumn{1}{c}{\textbf{Train$_\text{ext}$}} & \multicolumn{1}{c}{\textbf{$\mathbf{p_W}$}} & \multicolumn{1}{c}{\textbf{Train$_\text{orig}$}} & \multicolumn{1}{c}{\textbf{Train$_\text{ext}$}} & \multicolumn{1}{c}{\textbf{$\mathbf{p_W}$}} \\
                    \midrule\midrule
                    \multirow{2}{*}{\textbf{$\mathbf{D_{500}}$}} & \textbf{0.2} & \multirow{2}{*}{0.21 (.07)} & \textbf{0.79} (.07) & 0.00* & \multirow{2}{*}{0.21 (.06)} & \textbf{0.80} (.00) & 0.00* & \multirow{2}{*}{0.29 (.09)} & \textbf{0.79} (.01) & 0.00* \\
                     & \textbf{0.9} &  & 0.77 (.01) & 0.00* &  & 0.78 (.01) & 0.00* &  & 0.77 (.01) & 0.00* \\
                    \midrule
                    \multirow{2}{*}{\textbf{$\mathbf{D_{1000}}$}} & \textbf{0.2} & \multirow{2}{*}{0.26 (.04)} & \textbf{0.80} (.01) & 0.00* & \multirow{2}{*}{0.27 (.05)} & \textbf{0.82} (.01) & 0.00* & \multirow{2}{*}{0.36 (.05)} & \textbf{0.80} (.01) & 0.00* \\
                     & \textbf{0.9} &  & \textbf{0.80} (.01) & 0.00* &  & 0.80 (.01) & 0.00* &  & \textbf{0.80} (.01) & 0.00* \\
                    \midrule
                    \multirow{2}{*}{\textbf{$\mathbf{D_{10000}}$}} & \textbf{0.1} & \multirow{2}{*}{0.72 (.04)} & 0.81 (.01) & 0.00* & \multirow{2}{*}{0.76 (.02)} & 0.82 (.01) & 0.00* & \multirow{2}{*}{0.74 (.03)} & 0.81 (.01) & 0.00* \\
                     & \textbf{0.9} &  & \textbf{0.83} (.00) & 0.00* &  & \textbf{0.84} (.00) & 0.00* &  & \textbf{0.83} (.00) & 0.00* \\
                    \midrule
                    \multirow{2}{*}{\textbf{$\mathbf{D_{60000}}$}} & \textbf{0.1} & \multirow{2}{*}{0.82 (.01)} & 0.81 (.01) & 0.00* & \multirow{2}{*}{0.84 (.02)} & 0.82 (.01) & 0.00* & \multirow{2}{*}{0.82 (.02)} & 0.81 (.00) & 0.00* \\
                     & \textbf{0.9} &  & \textbf{0.83} (.01) & 0.09* &  & \textbf{0.84} (.00) & 0.11\:\: &  & \textbf{0.83} (.01) & 0.08* \\  % chktex 21
                    \bottomrule
                \end{tabular}
            \end{table}

            These results show that our synthesis algorithm can successfully identify high-confidence class regions of the feature space with very few original training samples. This makes our proposal highly advantageous for Deep Learning, as its synthesised samples allow for significantly improved model generalisation without requiring additional data collection.

        \subsubsection{Imbalanced data}\label{sec:imbalancedDataResults}
\begin{figure}[!h]
                \centering
                \includegraphics[width=0.5\linewidth]{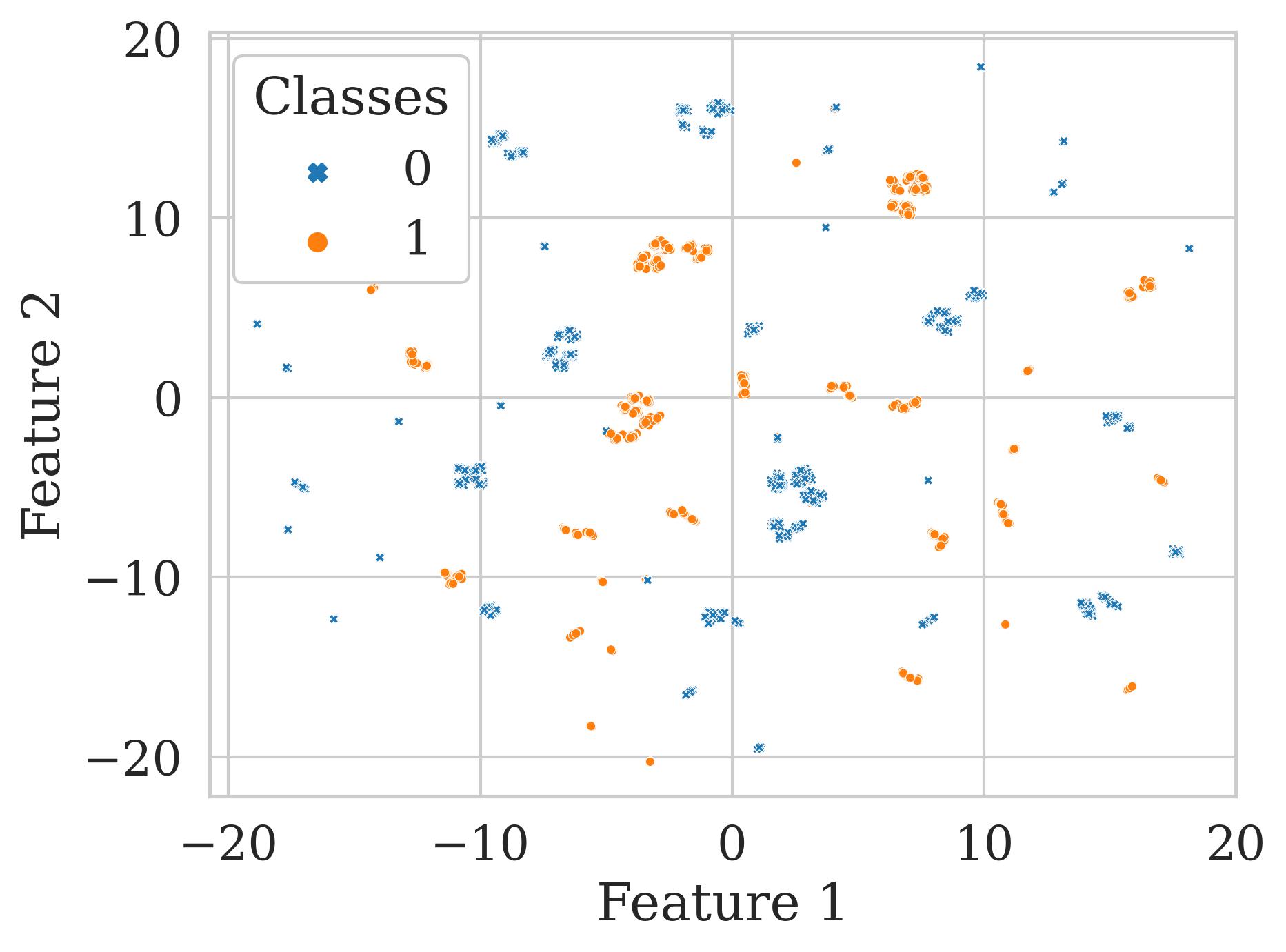}
                \caption{The MSHRM dataset. Originally roughly equally distributed, we artificially sub-sampled the data to simulate class imbalances of varying severity for synthesis.}\label{fig:mshrmOriginalDataset}
            \end{figure}

            Class imbalances are a frequent and often inevitable occurrence in real-world datasets due to the underlying population's unequal distribution (e.g., healthy vs infected). Even if the total number of available samples is large, the minority class's under-representation may lead to skewed results. To illustrate the effects of imbalance on model performance and the combating benefits of our data synthesis algorithm, we use the MSHRM benchmark dataset~\cite{Knopf1981}. MSHRM contains records of around 8,100 samples separated into edible (class 0) and poisonous (class 1) mushrooms. The pre-processing steps include replacing categorical variables with dummies and reducing the samples to two dimensions with UMAP (\Cref{fig:mshrmOriginalDataset}).

\FloatBarrier{}

            Once again, we simulated different levels of class imbalance by sub-sampling the original dataset to create four subsets $D$ with 1:1, 1:2, 1:4, and 1:9 class ratios. \Cref{tab:mshrmSampleCounts} contains the tested subsets and their per-class sample counts. Following the procedure in \Cref{sec:smallDataResults}, $\epsilon=0.1$ was selected via the elbow method.
            Worth noting is that the level of imbalance had a drastic impact on the number and class distribution of our synthetic samples, far out-reaching the class ratio of Train$_{\text{orig}}$ (e.g., $D_{1:9}$ with 1:9 original vs 1:55 synthetic sample class ratios).

            \begin{table}[!h]
                \centering
                \captionsetup{width=\textwidth}
                \caption{MSHRM sample counts with varying imbalance, differentiated by class. The original data's imbalance significantly impacted the balance of the synthetic samples ($\epsilon=0.1, \gamma=0.01$). The larger the imbalance was, the more samples were synthesised for the minority class 1.}\label{tab:mshrmSampleCounts}
                \begin{tabular}{ccrrrrrr}
                    \toprule
                     \multicolumn{1}{l}{} & \textbf{} & \multicolumn{1}{c}{\textbf{Test}} & \multicolumn{3}{c}{\textbf{Train$_{\mathbf{orig}}$}} & \multicolumn{1}{c}{\textbf{Train$_{\mathbf{syn}}$}} & \multicolumn{1}{c}{\textbf{Train$_{\mathbf{ext}}$}} \\
                    \cmidrule(lr){3-3} \cmidrule(lr){4-6} \cmidrule(lr){7-7} \cmidrule(lr){8-8}
                    \textbf{Subset} & \textbf{Class} & \multicolumn{1}{c}{\textbf{All}} & \multicolumn{1}{c}{\textbf{Prop.}} & \multicolumn{1}{c}{\textbf{Calib.}} & \multicolumn{1}{c}{\textbf{All}} & \multicolumn{1}{c}{\textbf{All}} & \multicolumn{1}{c}{\textbf{All}} \\
                    \midrule\midrule
                    \multirow{3}{*}{\textbf{$\mathbf{D_{1:1}}$}} & \textbf{0} & 1,394 & 1,698 & 1,132 & 2,830 & 26,585 & 29,415 \\
                     & \textbf{1} & 1,287 & 1,568 & 1,045 & 2,613 & 16,926 & 19,539 \\
                     & \textbf{All} & 2,681 & 3,266 & 2,177 & 5,443 & 43,511 & 48,954 \\
                    \midrule
                    \multirow{3}{*}{\textbf{$\mathbf{D_{1:2}}$}} & \textbf{0} & 1,394 & 1,696 & 1,130 & 2,826 & 29,251 & 32,077 \\
                     & \textbf{1} & 1,287 & 836 & 557 & 1,393 & 65,423 & 66,816 \\
                     & \textbf{All} & 2,681 & 2,532 & 1,687 & 4,219 & 94,674 & 98,893 \\
                    \midrule
                    \multirow{3}{*}{\textbf{$\mathbf{D_{1:4}}$}} & \textbf{0} & 1,394 & 1,699 & 1,132 & 2,831 & 27,162 & 29,993 \\
                     & \textbf{1} & 1,287 & 425 & 283 & 708 & 197,530 & 198,238 \\
                     & \textbf{All} & 2,681 & 2,124 & 1,415 & 3,539 & 224,692 & 228,231 \\
                    \midrule
                    \multirow{3}{*}{\textbf{$\mathbf{D_{1:9}}$}} & \textbf{0} & 1,394 & 1,697 & 1,131 & 2,828 & 30,383 & 33,211 \\
                     & \textbf{1} & 1,287 & 189 & 126 & 315 & 1,658,057 & 1,658,372 \\
                     & \textbf{All} & 2,681 & 1,886 & 1,257 & 3,143 & 1,688,440 & 1,691,583 \\
                    \bottomrule
                \end{tabular}
            \end{table}

\begin{figure}[!ht]
                \centering
                \begin{subfigure}[b]{0.48\textwidth}
                    \centering
                    \includegraphics[width=\textwidth]{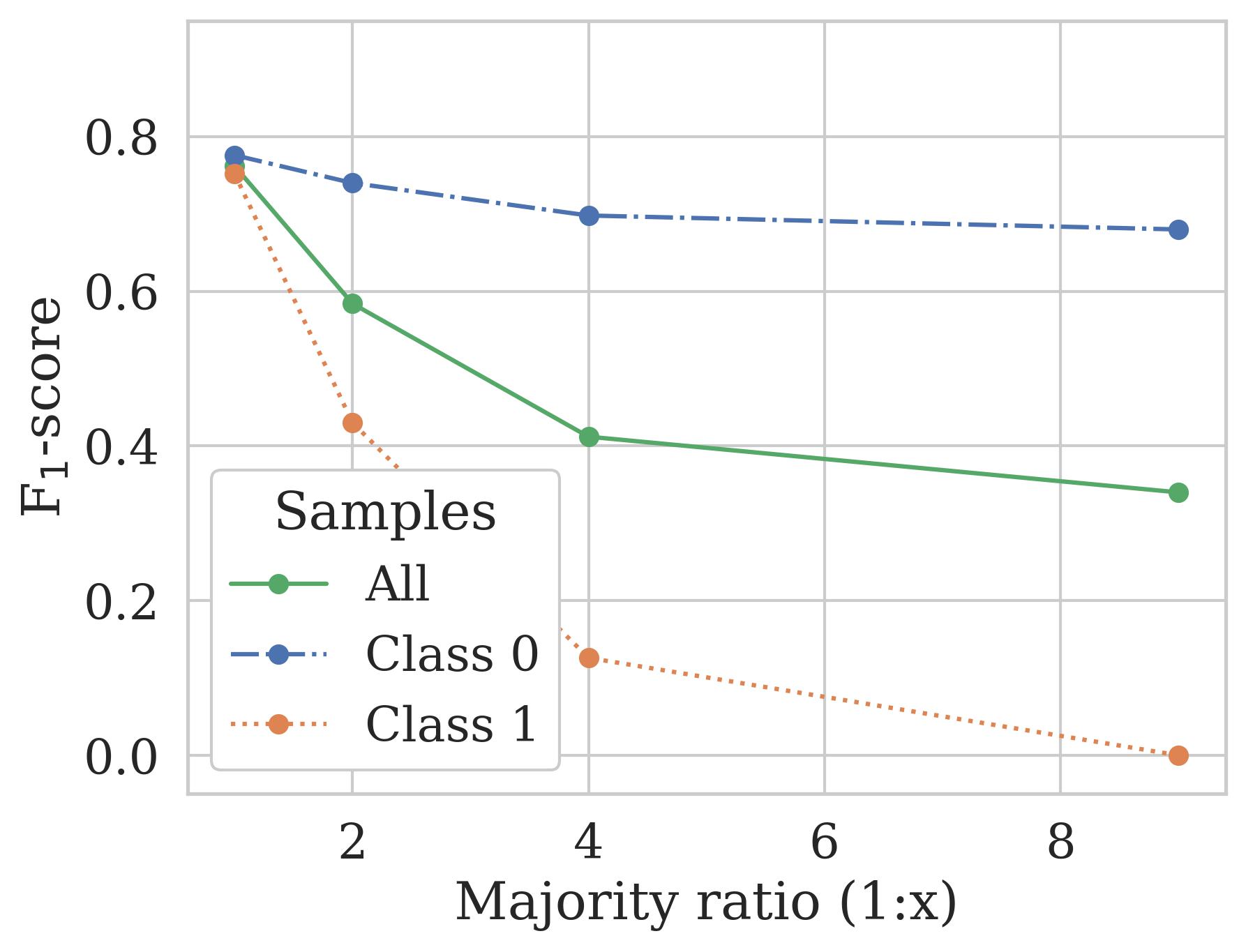}
                    \caption{Train$_{\text{orig}}$.}\label{fig:mshrmOriginal}
                \end{subfigure}
                \hfill
                \begin{subfigure}[b]{0.495\textwidth}
                    \centering
                    \includegraphics[width=\textwidth]{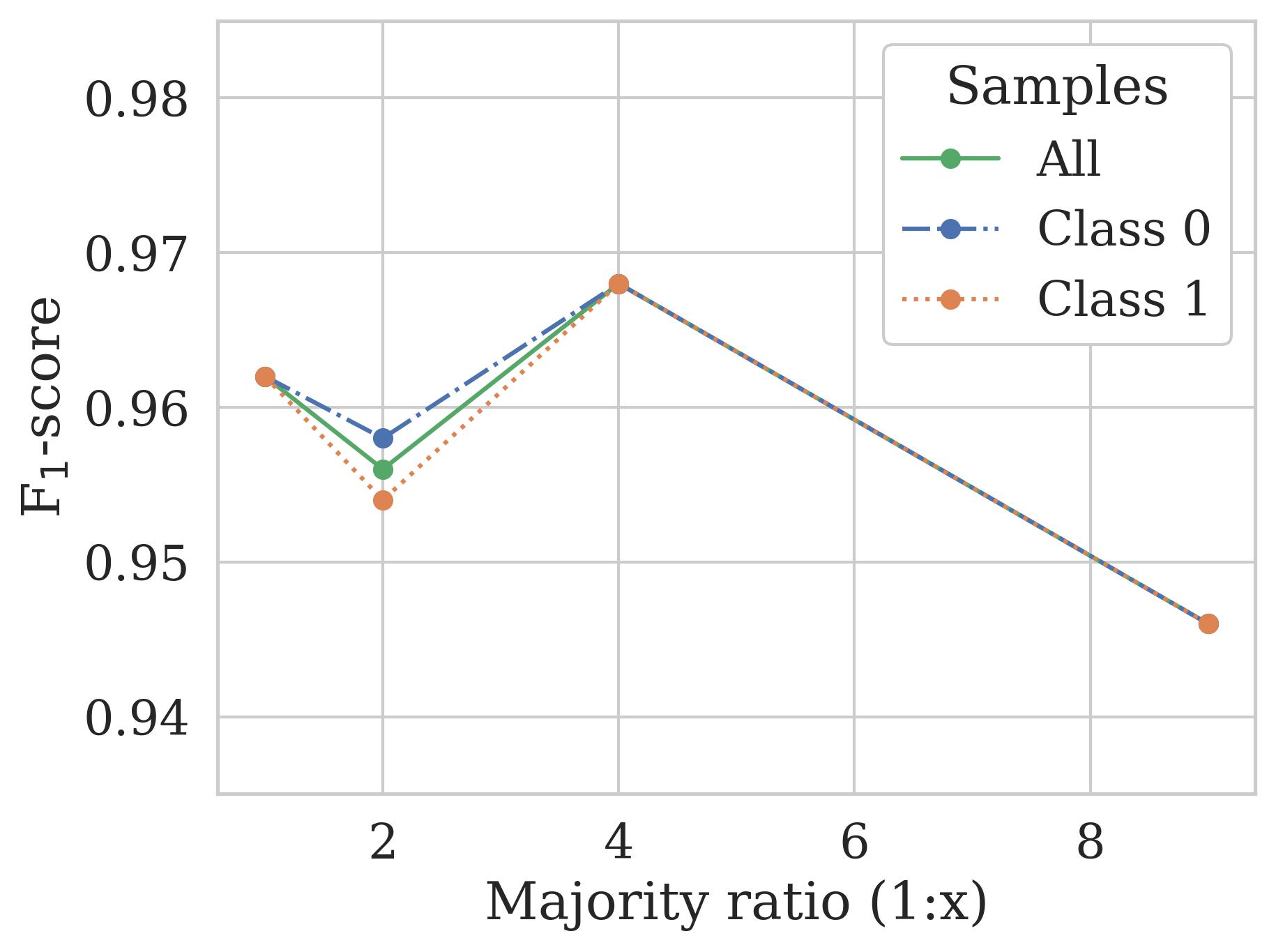}
                    \caption{Train$_{\text{ext}}$.}\label{fig:mshrmExtended01}
                \end{subfigure}
                \caption{Mean Deep Learning performance on the MSHRM dataset with varying class imbalance. Performance on the original dataset decreased significantly with increasing imbalance, widening class discrepancies. Conversely, performance improved on the extended training set because the number of synthetic samples was increased exponentially (\Cref{tab:mshrmSampleCounts}).}\label{fig:mshrmPerformance}
            \end{figure}

            The Deep Learning results are shown in \Cref{tab:mshrmPerformance}. As expected, model performance on Train$_{\text{orig}}$ decreased severely with the increasing imbalance (76\%--34\%). The primary contributor was the F$_1$-score on the minority class 1, which dropped from 75\% to 0\%. In contrast, the model's overall F$_1$-score when trained on the extended training sets remained steady at around 96\%, increasing by 20--61 percentage points with increasing class imbalance. Interestingly, although the synthetic samples reintroduced the class imbalance (albeit with class 1 now as the majority class), the imbalance was not represented in the final results.
            Visualising this tendency in \Cref{fig:mshrmPerformance}, we found that over-supporting the minority class with synthetic samples did not have the same negative performance impacts as the original imbalance had.

            \begin{table}[!ht]
                \centering
                \setlength{\tabcolsep}{2.pt}
                \captionsetup{width=\textwidth}
                \caption{MSHRM Deep Learning results with varying class imbalance. We report the mean and standard deviation across five iterations. Performance on the original dataset decreased significantly with increasing imbalance. Conversely, performance on the synthetically extended training set was significantly improved compared to the baseline ($p_W < 0.1$) and remained stable across classes and as the imbalance shifted.}\label{tab:mshrmPerformance}
                \begin{tabular}{ccrrrrrrrrr}
                    \toprule
                    & \multicolumn{1}{c}{} & \multicolumn{3}{c}{\textbf{F$_{\mathbf{1}}$-score}} & \multicolumn{3}{c}{\textbf{Precision}} & \multicolumn{3}{c}{\textbf{Recall}} \\
                    \cmidrule(lr){3-5} \cmidrule(lr){6-8} \cmidrule(lr){9-11}
                    \textbf{Data} & \multicolumn{1}{c}{\textbf{Class}} & \multicolumn{1}{c}{\textbf{Train$_\text{orig}$}} & \multicolumn{1}{c}{\textbf{Train$_\text{ext}$}} & \multicolumn{1}{c}{\textbf{$\mathbf{p_W}$}} & \multicolumn{1}{c}{\textbf{Train$_\text{orig}$}} & \multicolumn{1}{c}{\textbf{Train$_\text{ext}$}} & \multicolumn{1}{c}{\textbf{$\mathbf{p_W}$}} & \multicolumn{1}{c}{\textbf{Train$_\text{orig}$}} & \multicolumn{1}{c}{\textbf{Train$_\text{ext}$}} & \multicolumn{1}{c}{\textbf{$\mathbf{p_W}$}} \\
                    \midrule\midrule
                    \multirow{3}{*}{\textbf{$\mathbf{D_{1:1}}$}} & \textbf{0} & 0.78 (.04) & \textbf{0.96} (.01) & 0.00* & 0.77 (.07) & \textbf{0.97} (.01) & 0.00* & 0.78 (.06) & \textbf{0.95} (.02) & 0.00* \\
                     & \textbf{1} & 0.75 (.07) & \textbf{0.96} (.01) & 0.00* & 0.76 (.04) & \textbf{0.95} (.02) & 0.00* & 0.75 (.11) & \textbf{0.97} (.01) & 0.00* \\
                     & \textbf{All} & 0.76 (.05) & \textbf{0.96} (.01) & 0.00* & 0.77 (.05) & \textbf{0.96} (.01) & 0.00* & 0.76 (.05) & \textbf{0.96} (.01) & 0.00* \\
                    \midrule
                    \multirow{3}{*}{\textbf{$\mathbf{D_{1:2}}$}} & \textbf{0} & 0.74 (.01) & \textbf{0.96} (.03) & 0.00* & 0.59 (.01) & \textbf{0.98} (.02) & 0.00* & \textbf{0.98} (.02) & 0.93 (.04) & 0.00* \\
                     & \textbf{1} & 0.43 (.05) & \textbf{0.95} (.03) & 0.00* & 0.93 (.05) & \textbf{0.93} (.04) & 0.08* & 0.28 (.05) & \textbf{0.98} (.02) & 0.00* \\
                     & \textbf{All} & 0.58 (.03) & \textbf{0.96} (.02) & 0.00* & 0.76 (.03) & \textbf{0.96} (.02) & 0.00* & 0.63 (.01) & \textbf{0.96} (.02) & 0.00* \\
                    \midrule
                    \multirow{3}{*}{\textbf{$\mathbf{D_{1:4}}$}} & \textbf{0} & 0.70 (.02) & \textbf{0.97} (.00) & 0.00* & 0.54 (.03) & \textbf{0.97} (.00) & 0.00* & \textbf{1.00} (.00) & 0.97 (.01) & 0.00* \\
                     & \textbf{1} & 0.13 (.17) & \textbf{0.97} (.00) & 0.00* & 0.39 (.54) & \textbf{0.97} (.01) & 0.04* & 0.08 (.11) & \textbf{0.96} (.00) & 0.00* \\
                     & \textbf{All} & 0.41 (.10) & \textbf{0.97} (.00) & 0.00* & 0.47 (.28) & \textbf{0.97} (.00) & 0.00* & 0.54 (.05) & \textbf{0.97} (.00) & 0.00* \\
                    \midrule
                    \multirow{3}{*}{\textbf{$\mathbf{D_{1:9}}$}} & \textbf{0} & 0.68 (.00) & \textbf{0.95} (.01) & 0.00* & 0.52 (.00) & \textbf{0.96} (.00) & 0.00* & \textbf{1.00} (.00) & 0.94 (.02) & 0.00* \\
                     & \textbf{1} & 0.00 (.00) & \textbf{0.95} (.01) & 0.00* & 0.00 (.00) & \textbf{0.94} (.02) & 0.00* & 0.00 (.00) & \textbf{0.96} (.00) & 0.00* \\
                     & \textbf{All} & 0.34 (.00) & \textbf{0.95} (.01) & 0.00* & 0.26 (.00) & \textbf{0.95} (.01) & 0.00* & 0.50 (.00) & \textbf{0.95} (.01) & 0.00* \\
                    \bottomrule
                \end{tabular}
            \end{table}

\FloatBarrier{}
        \subsubsection{Overlapping classes}\label{sec:overlappingClassesResults}
\begin{figure}[!h]
                \centering
                \begin{subfigure}[b]{0.325\textwidth}
                    \centering
                    \includegraphics[width=\textwidth]{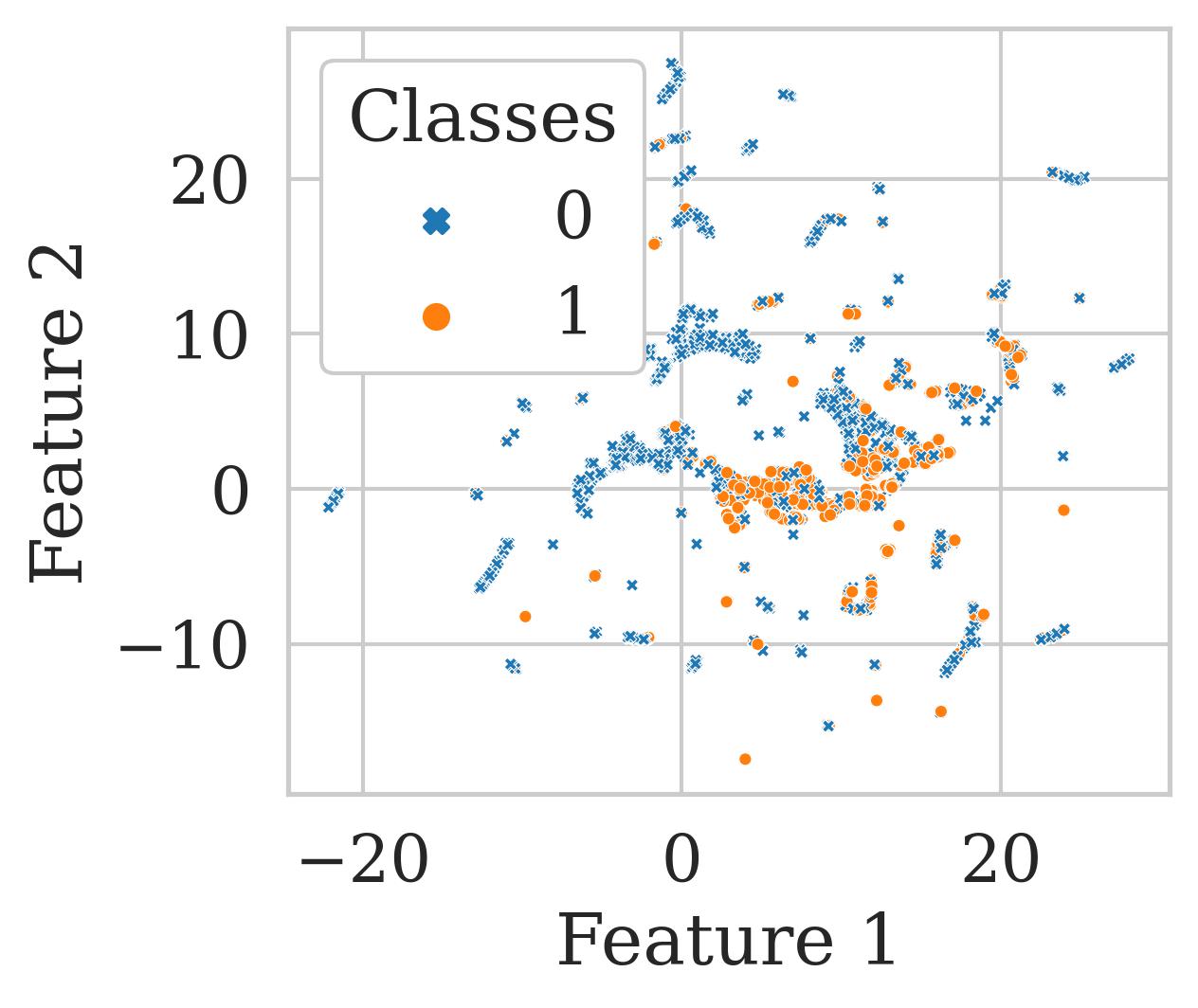}
                    \caption{All classes.}
                \end{subfigure}
                \hfill
                \begin{subfigure}[b]{0.325\textwidth}
                    \centering
                    \includegraphics[width=\textwidth]{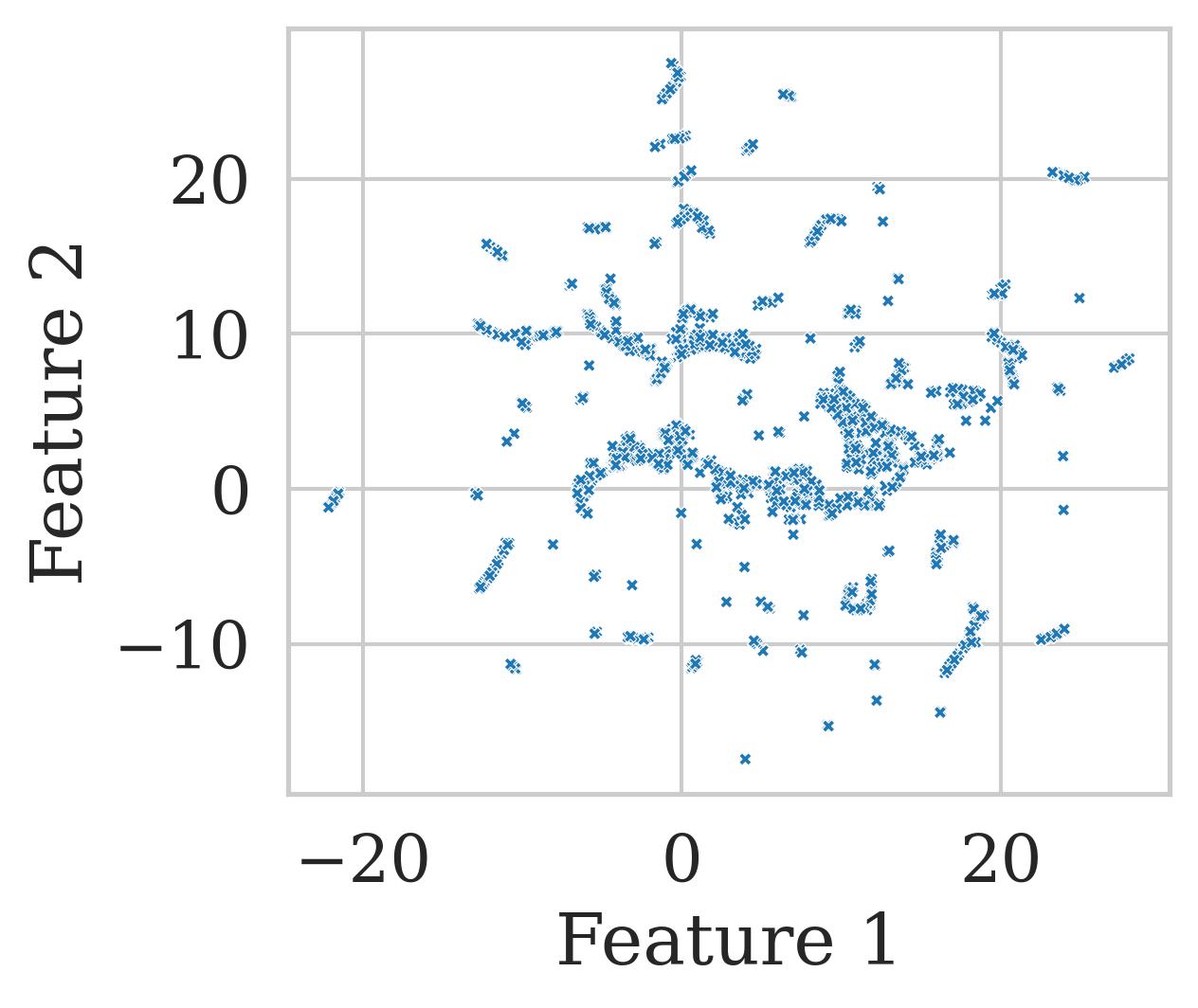}
                    \caption{Class 0.}
                \end{subfigure}
                \hfill
                \begin{subfigure}[b]{0.325\textwidth}
                    \centering
                    \includegraphics[width=\textwidth]{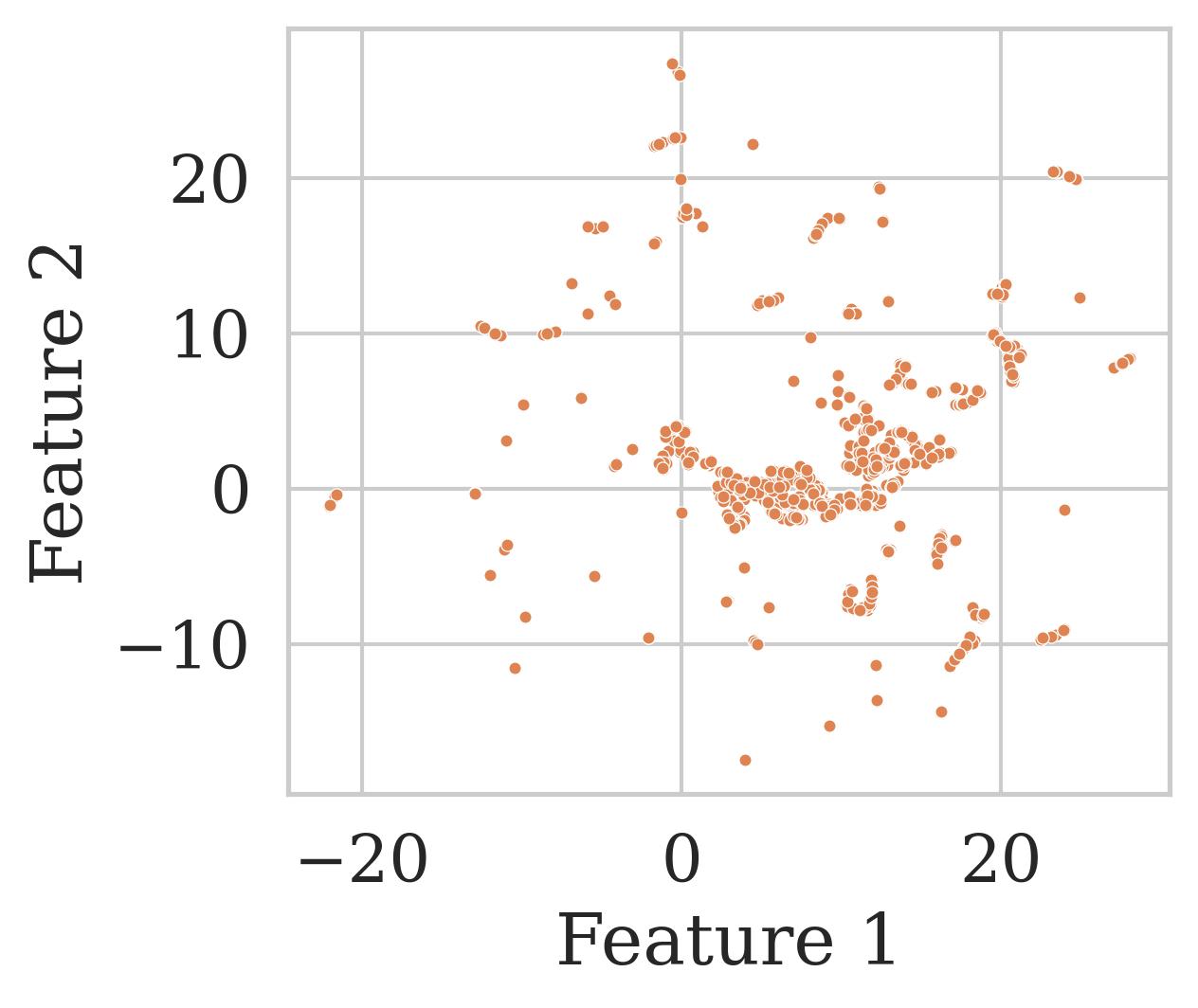}
                    \caption{Class 1.}
                \end{subfigure}
                \caption{The WINE dataset showed significant class overlap in the feature space.  Consequently, the high-confidence regions identified by our synthesis algorithm were also overlapped. The non-separability of the classes made accurate predictions challenging.}\label{fig:wineOriginalDataset}
            \end{figure}

            Deep Learning modelling of prediction tasks relies on the separability of the classes. However, real-world datasets often do not have clear separation due to sample noise and uninformative features.
            We demonstrate the advantages of our proposed algorithm in these cases on the WINE benchmark dataset~\cite{Cortez2009}, where we classify whether a sample represents white (class 0) or red (class 1) wine based on chemical measurements. As shown in \Cref{fig:wineOriginalDataset}, this ``difficult'' dataset had no class separability, with both classes overlapping quite significantly in the feature space. Consequently, the high-confidence regions for data synthesis overlapped as well.
            \Cref{tab:wineSampleCounts} records the evaluated original, synthetic, and extended sample counts ($\epsilon=0.2, \gamma=0.1$).

            \begin{table}[!h]
                \centering
                \captionsetup{width=\textwidth}
                \caption{WINE original and synthetic sample counts. While synthesis generally maintained the class's non-separability, significantly increasing the number of available training samples may nonetheless support Deep Learning generalisation.}\label{tab:wineSampleCounts}
                \begin{tabular}{crrrrcc}
                    \toprule
                    \multicolumn{1}{c}{\textbf{Classes}}& \multicolumn{1}{c}{\textbf{Test}} & \multicolumn{3}{c}{\textbf{Train$\mathbf{_\text{orig}}$}} & \textbf{Train$\mathbf{_\text{syn}}$} & \multicolumn{1}{c}{\textbf{Train$\mathbf{_\text{ext}}$}} \\
                    \cmidrule(lr){2-2} \cmidrule(lr){3-5} \cmidrule(lr){6-6} \cmidrule(lr){7-7}
                    & \multicolumn{1}{c}{\textbf{All}} & \multicolumn{1}{l}{\textbf{Prop.}} & \multicolumn{1}{c}{\textbf{Calib.}} & \multicolumn{1}{c}{\textbf{All}} & \textbf{All} & \multicolumn{1}{c}{\textbf{All}} \\
                    \midrule\midrule
                    \textbf{Class 0} & 536 & 653 & 435 & 1,088 & 29,169 & 30,257 \\
                    \textbf{Class 1} & 536 & 653 & 435 & 1,088 & 31,149 & 32,237 \\
                    \textbf{All} & 1,072 & 1,306 & 870 & 2,176 & 60,318 & 62,494 \\
                    \bottomrule
                \end{tabular}
            \end{table}

\begin{figure}[!ht]
                \centering
                \includegraphics[width=0.5\linewidth]{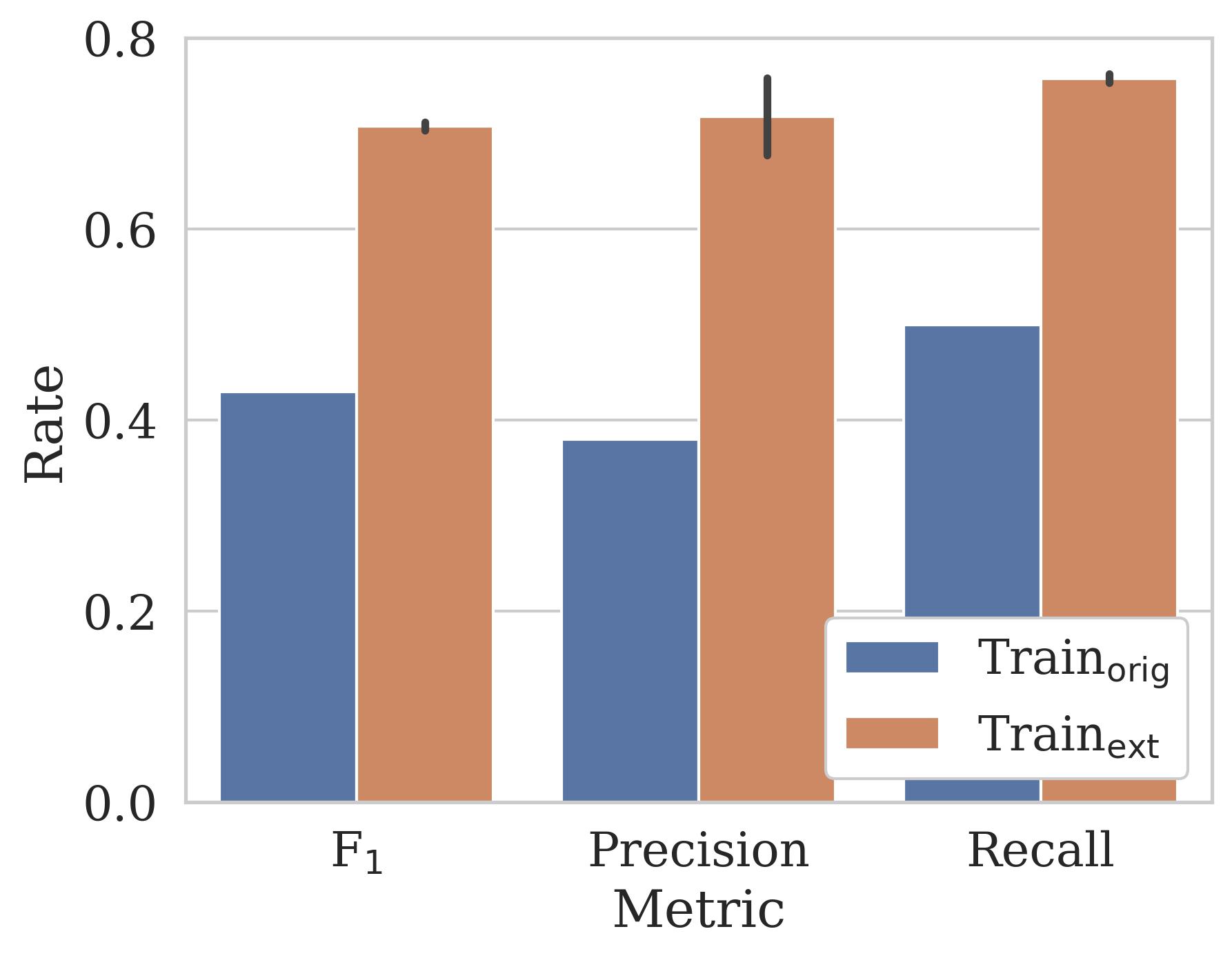}
                \caption{Mean Deep Learning performance on the WINE dataset. The error bars show the standard deviation. Even though synthesis did not improve the classes' separability, the large quantity of additional samples significantly improved the models' generalisation. All improvements were deemed statistically significant by the Wilcoxon test ($p_W < 0.1$).}\label{fig:winePerformance}
            \end{figure}

\FloatBarrier{}
            \Cref{fig:winePerformance} visualises the mean and standard deviation performance of five models trained on the original Train$_{\text{orig}}$ and the extended Train$_{\text{ext}}$ datasets. Compared to the 43\% F$_1$-score achieved on the original data, including the synthesised data for training increased the performance by 26 percentage points to a total of 69\%. Precision and recall results confirmed that the model's ability to identify relevant samples was indeed strengthened. In summary, even though our proposed algorithm did not increase the linear separability of the classes, it nonetheless significantly improved the model's performance. The model's learning was supported through the synthesis of a large number of new training samples.

        \subsubsection{Synthetic replacement}\label{sec:synteticReplacementResults}
\begin{figure}[!h]
                \centering
                \begin{subfigure}[b]{0.4\textwidth}
                    \centering
                    \includegraphics[width=\textwidth]{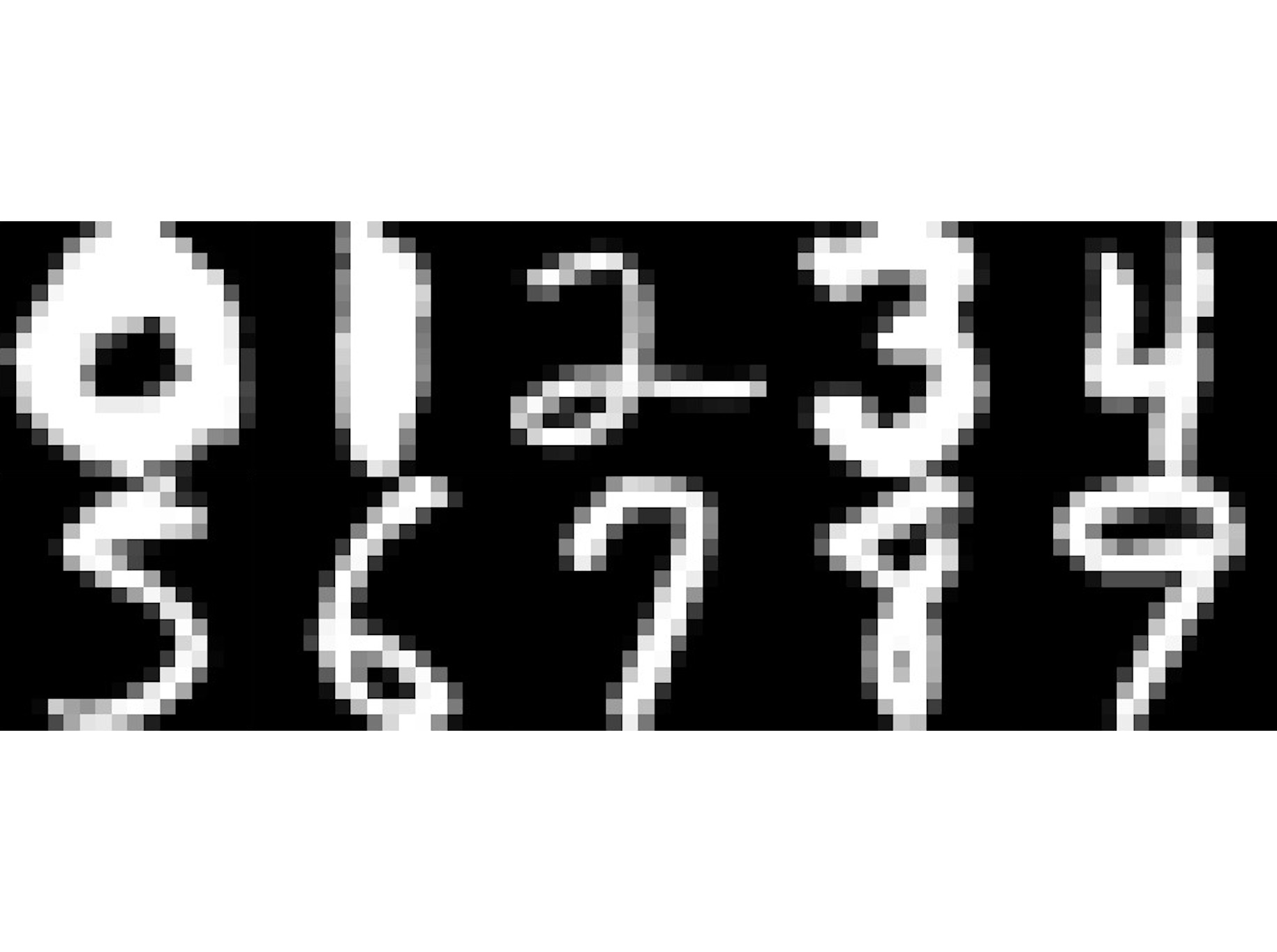}
                    \caption{Handwritten digit samples.}\label{fig:uspsImages}
                \end{subfigure}
                \hfill
                \begin{subfigure}[b]{0.4\textwidth}
                    \centering
                    \includegraphics[width=\textwidth]{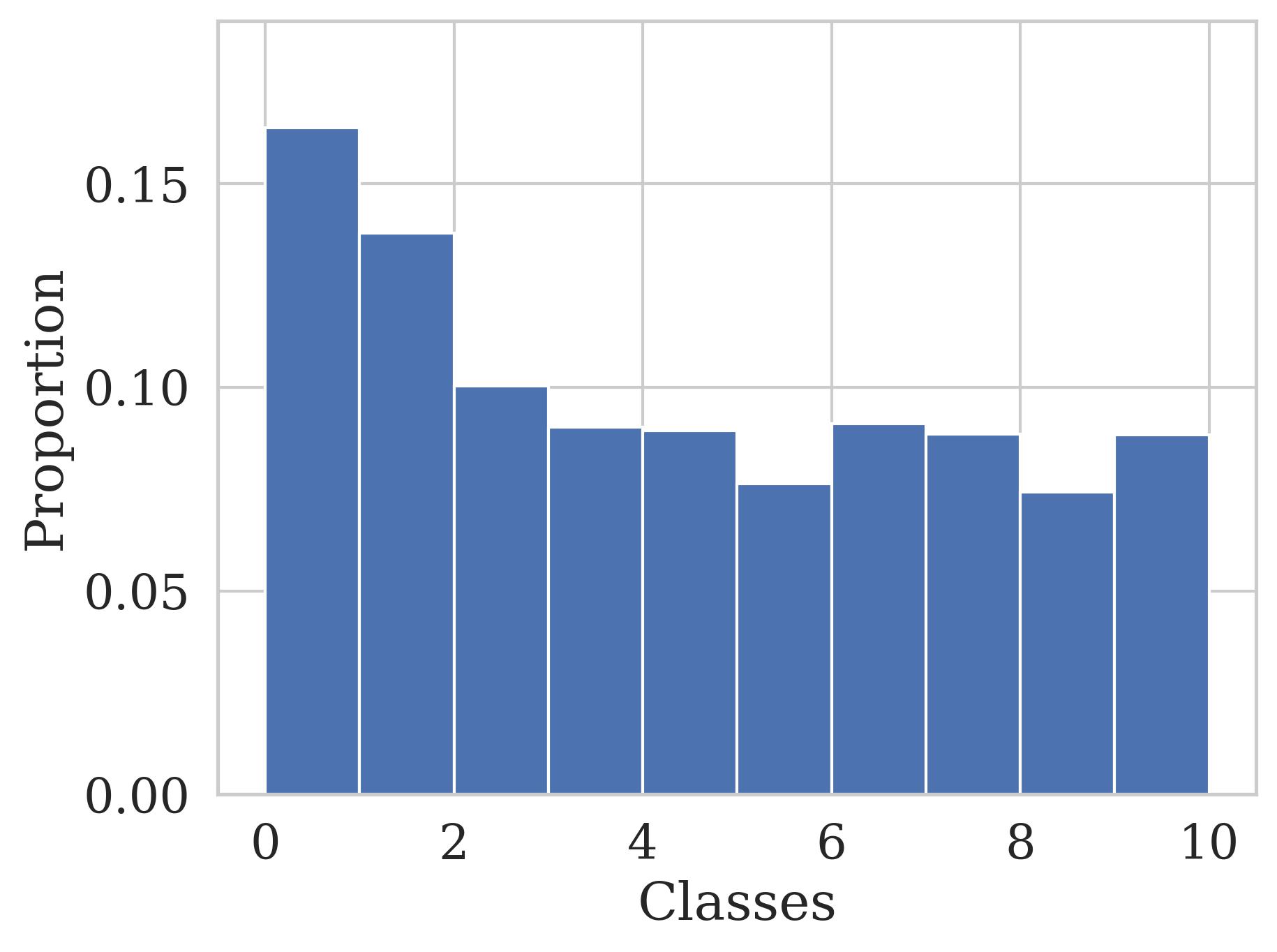}
                    \caption{Class distribution.}\label{fig:uspsClassDistribution}
                \end{subfigure}
                \caption{Overview of the USPS dataset. Each sample is a grey-scale image of a handwritten digit in a 16$\times$16 pixel format.}\label{fig:uspsDataset}
            \end{figure}

            As a final show of our algorithm's ability to accurately synthesise new samples, we tested a complete synthetic replacement of the original dataset for Deep Learning training. The experiments were carried out on the USPS dataset of digits scanned from envelopes by the U.S Postal Service~\cite{Hull1994}, shown in \Cref{fig:uspsImages}. The classes were roughly balanced, except for a slight majority in classes 0 and 1 (\Cref{fig:uspsClassDistribution}). The relative proportions were maintained in all sampled subsets. Of the roughly 9,300 available samples, 2,000 were held back for testing. The remaining 7,291 training samples were temporarily split into 60\% proper training and 40\% calibration subsets for conformal synthesis.

\begin{figure}[!hp]
                \begin{subfigure}[b]{0.49\textwidth}
                    \centering
                    \includegraphics[width=\textwidth]{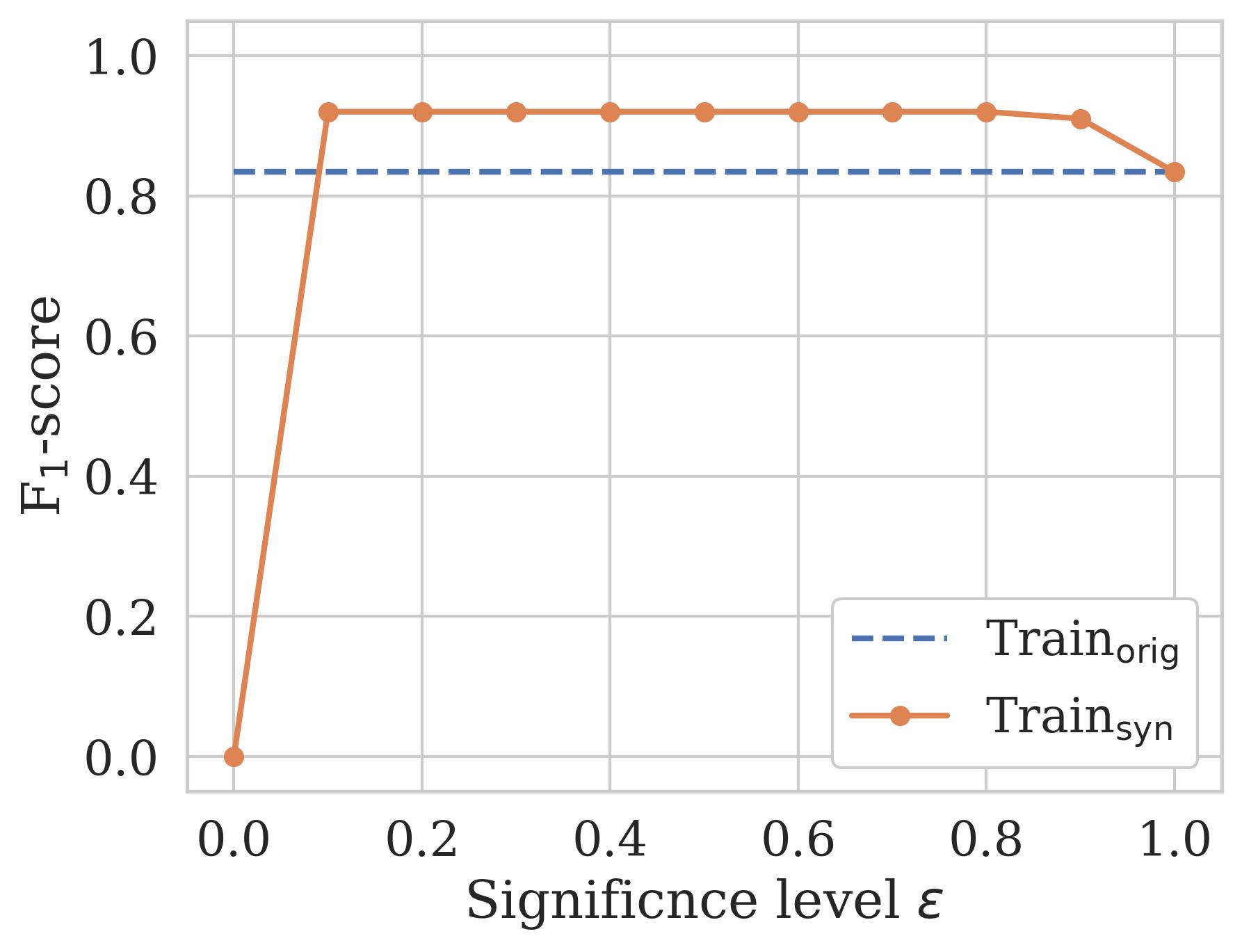}
                    \caption{Model performance with varying $\epsilon$.}\label{fig:uspsEpsilonSelection}
                \end{subfigure}
                \hfill
                \begin{subfigure}[b]{0.49\textwidth}
                    \centering
                    \includegraphics[width=\textwidth]{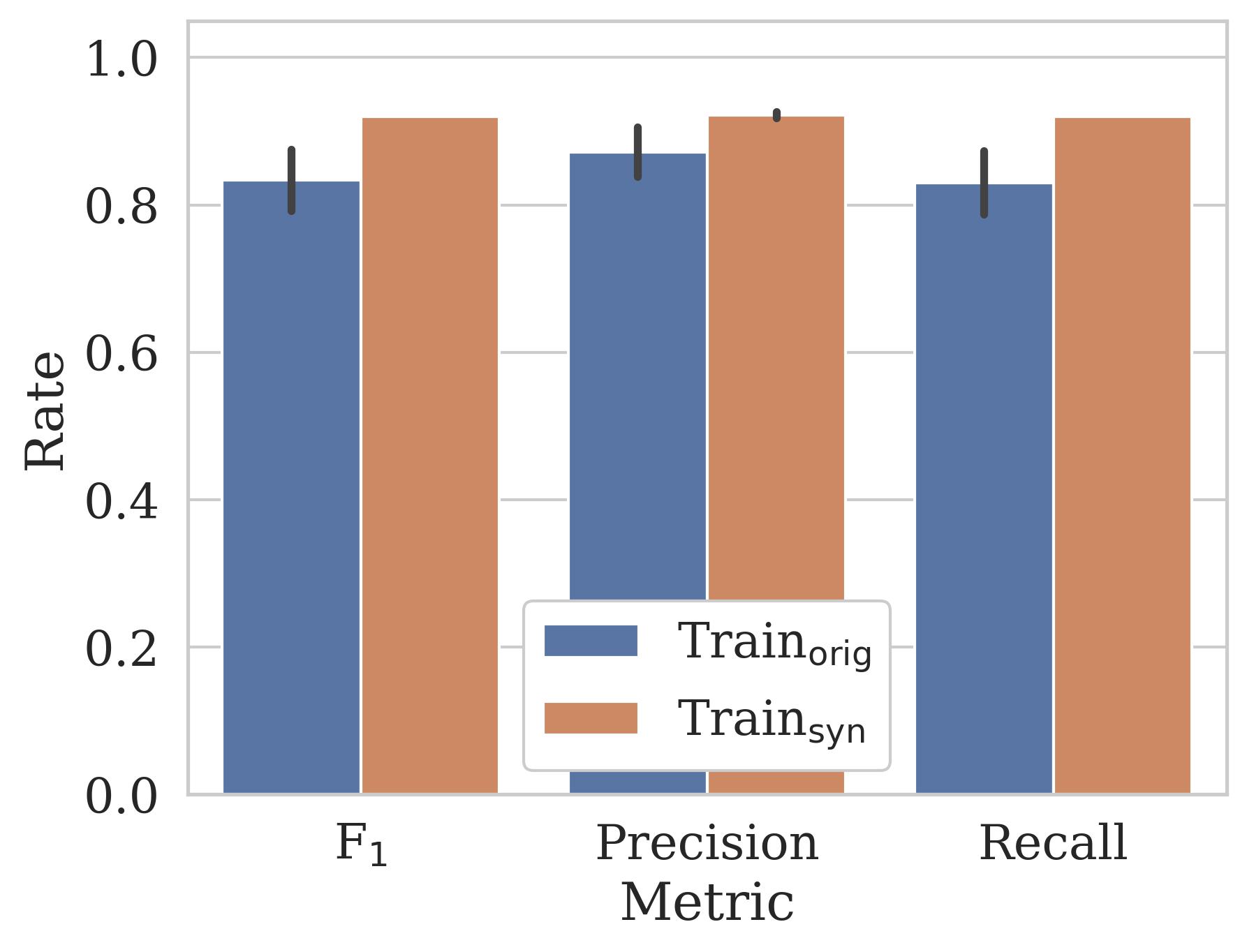}
                    \caption{Model performance with $\epsilon=0.1$.}\label{fig:uspsResults}
                \end{subfigure}
                \caption{USPS performance results on the baseline Train$_{\text{orig}}$ and a fully synthetic dataset ($\gamma=0.01$). The mean of five Deep Learning iterations was reported, including the standard deviation as error bars in (b). All synthetic improvements were found to be statistically significant with the Wilcoxon test ($p_W < 0.1$).}\label{fig:uspsSyntheticReplacementResults}
            \end{figure}

            Conformal synthesis was performed using all available training samples and the grid step $\gamma=0.01$. \Cref{fig:uspsSyntheticReplacementResults} summarises the results as $\epsilon$ varies. Compared to the baseline on Train$_{\text{orig}}$, the mean F$_1$-score performance improved by around 10 percentage points across all $\epsilon$. Since the results were consistent on different Train$_{\text{syn}}$, we may choose a low significance level to increase confidence. Therefore, \Cref{fig:uspsResults} visualises the models' performances for $\epsilon=0.1$. In addition to the mean of all metrics being improved, the standard deviation was also decreased, indicating a more robust model generalisation. Precision and recall results indicated that our synthesis algorithm preserved the original classes' balance, allowing models trained on the synthetic datasets to maintain or even strengthen their ability to identify relevant samples. The Wilcoxon test confirmed that all reported performance improvements achieved by Train$_{\text{syn}}$ were statistically significant, with $p_W$ values falling in the range 0.00--0.02.

\begin{figure}[!hp]
                \centering
                \begin{subfigure}[b]{0.49\textwidth}
                    \centering
                    \includegraphics[width=\textwidth]{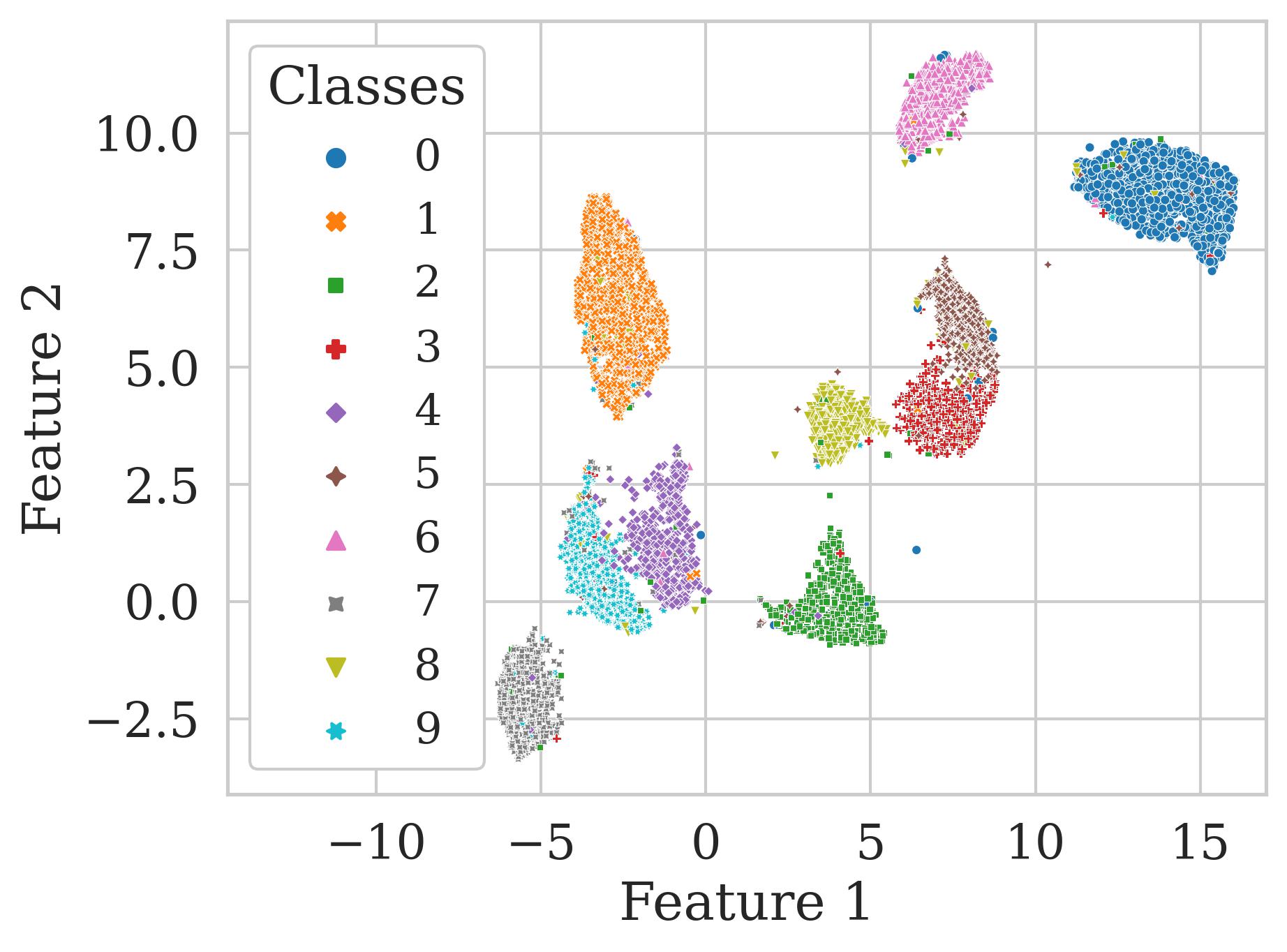}
                    \caption{Train$_{\text{orig}}$ points per class.}
                \end{subfigure}
                \hfill
                \begin{subfigure}[b]{0.49\textwidth}
                    \centering
                    \includegraphics[width=\textwidth]{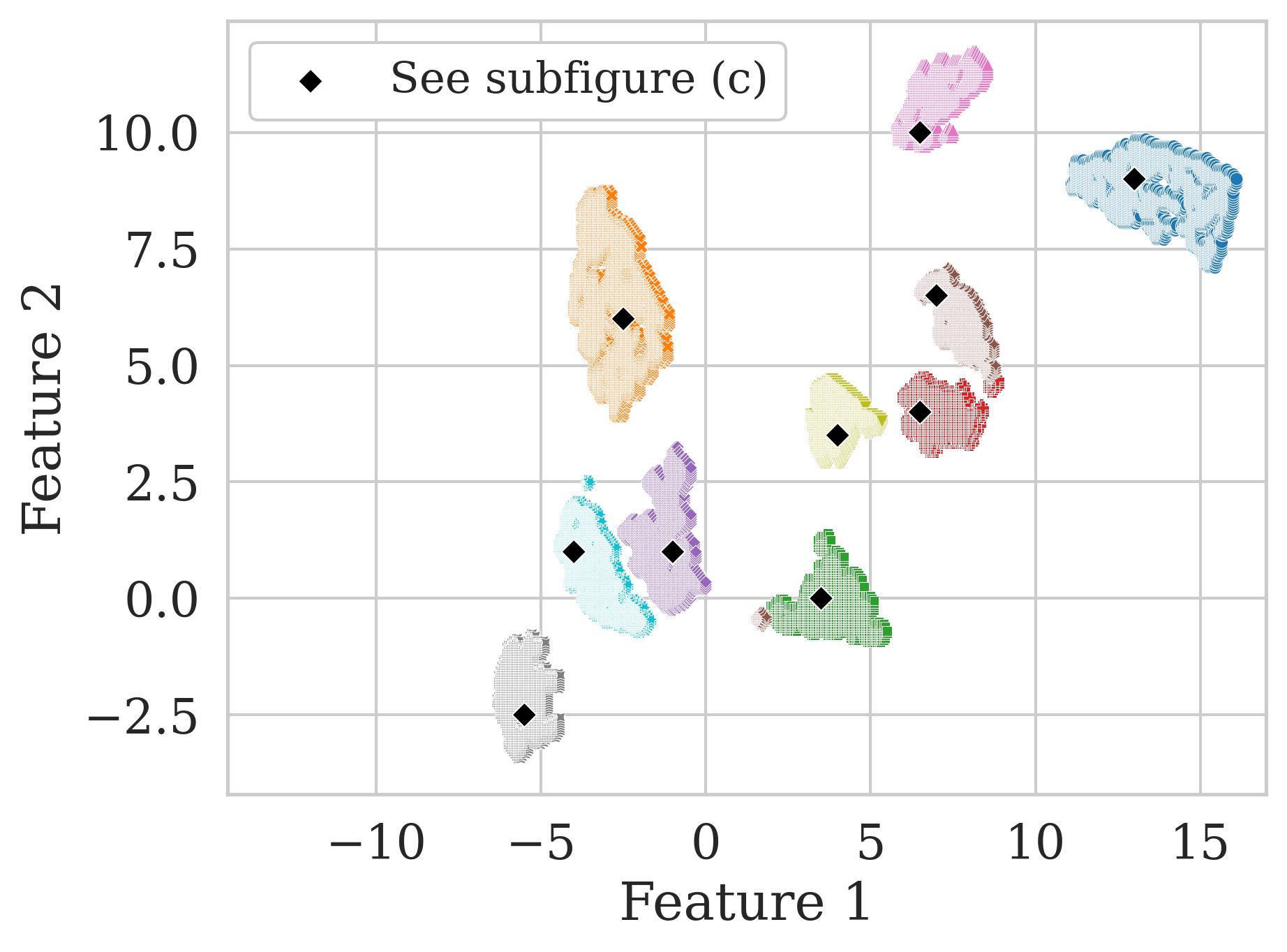}
                    \caption{Train$_{\text{syn}}$ points per class.}\label{fig:uspsSynthesisedFeatureSpace}
                \end{subfigure}
                \qquad
                \begin{subfigure}[b]{0.8\textwidth}
                    \centering
                    \includegraphics[width=\textwidth]{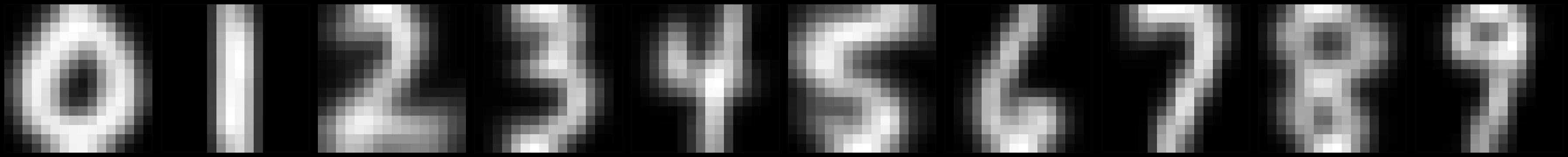}
                    \caption{Synthesised samples corresponding to the diamond markers in (b).}\label{fig:uspsInvertedSynthesisImages}
                    \end{subfigure}
                \caption{Original and generated samples of the USPS dataset after UMAP processing. Conformal synthesis successfully maintained the distribution of samples and classes in the feature space. Additionally, inverting the UMAP transformation on random synthesised points revealed recognisable digits similar to the original samples (\Cref{fig:uspsImages}).}\label{fig:uspsSyntheticReplacement}
            \end{figure}

            Finally, \Cref{fig:uspsSyntheticReplacement} investigates the synthesised samples in more detail. Comparing the original feature space and the synthesised feature space after UMAP dimensionality reduction, we see clear parallels in the sample and class distributions. Additionally, inverting a random synthesised test sample per class revealed recognisable digit images (\Cref{fig:uspsInvertedSynthesisImages}) similar to the original samples (\Cref{fig:uspsImages}), supporting the feature space confidence approach of our proposed algorithm.

        \subsubsection{Comparison to density-based synthesis}\label{sec:kdeSynthesisComparison}
\begin{figure}[!p]
                \centering
                \begin{subfigure}[b]{0.9\textwidth}
                    \centering
                    \includegraphics[width=\textwidth]{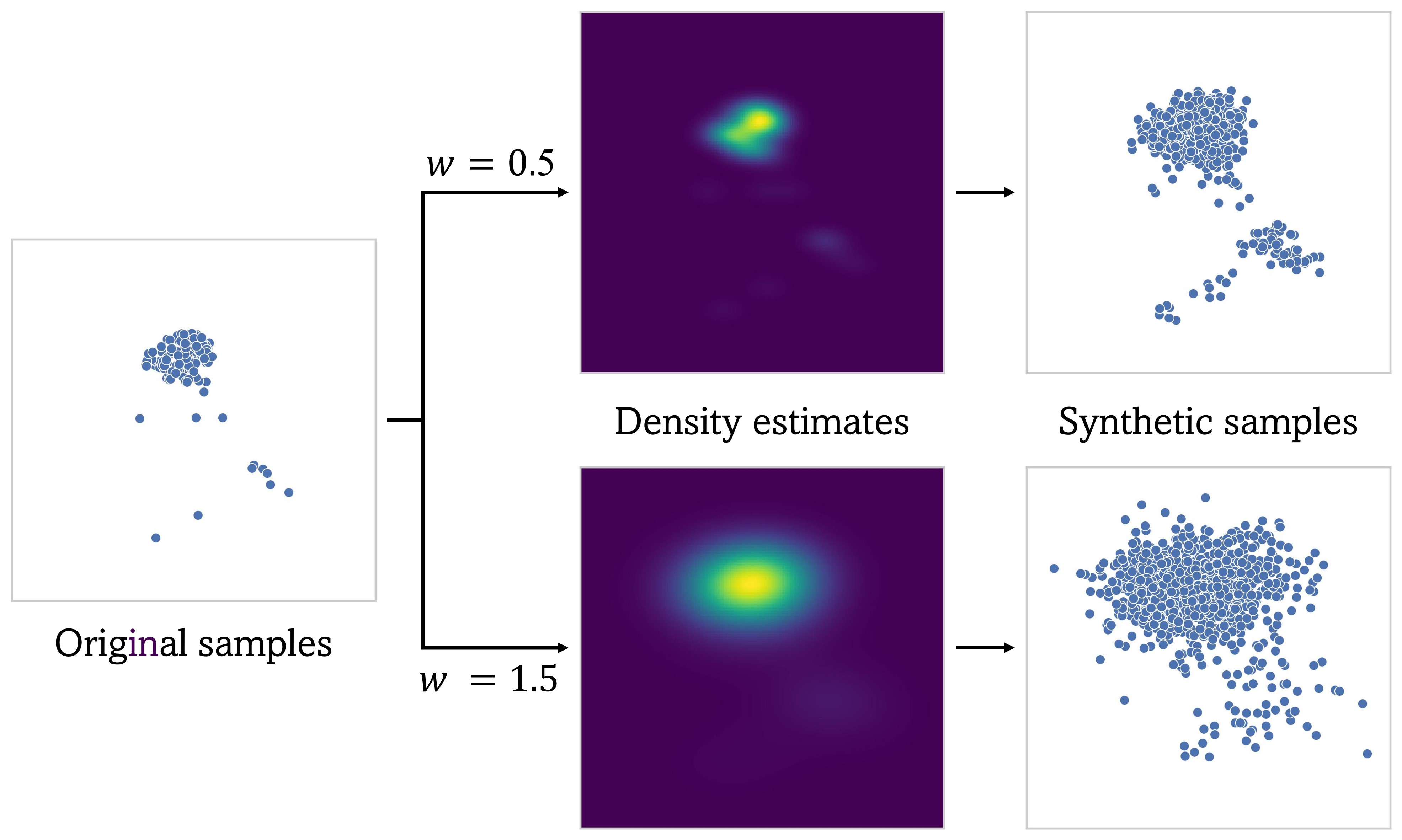}
                    \caption{Density estimation synthesis with KDE using bandwidth $w$.}\label{fig:kdeSynthesisProcedure}
                \end{subfigure}
                \qquad
                \begin{subfigure}[b]{0.9\textwidth}
                    \centering 
                    \includegraphics[width=\textwidth]{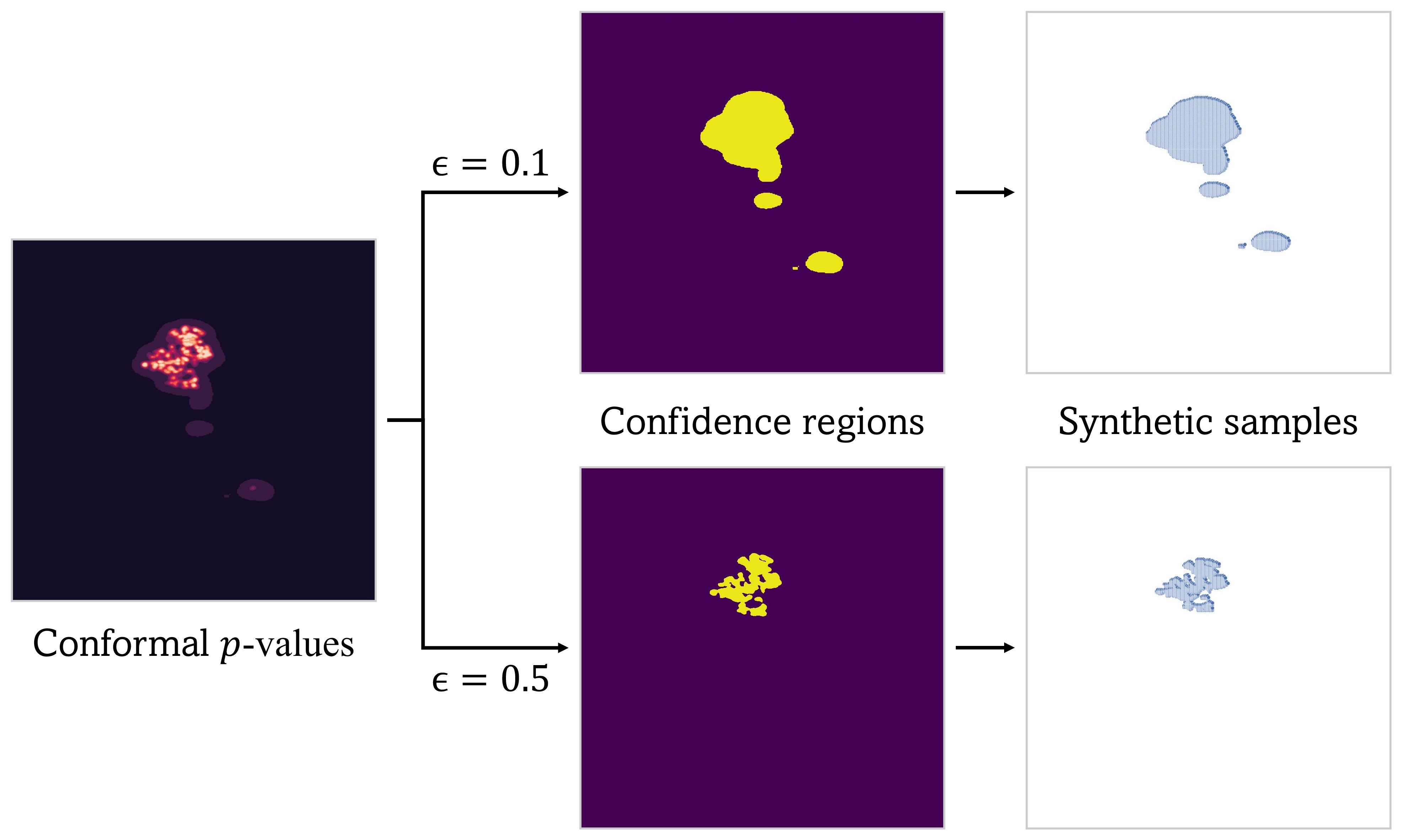}
                    \caption{Confident synthesis with the proposed algorithm using the significance level $\epsilon$.}\label{fig:conformalSynthesisProcedure}
                \end{subfigure}
                \caption{Comparison of the KDE and proposed synthesis algorithms on MNIST samples (class = 2). The KDE technique requires random sampling from its learned density distribution, which is heavily influenced by the freely selected bandwidth parameter $w$. In contrast, conformal synthesis relies on a confidence threshold $\epsilon$, which bounds and defines the synthesis regions.}\label{fig:conformalSynthesisVsKdeSynthesis}
            \end{figure}

           In this section, we compare our conformal synthesis algorithm against a state-of-the-art density-based technique to highlight our proposal's unique properties when generating new points from small datasets. In particular, we employ KDE-based density estimation with a Gaussian kernel and the Euclidean distance metric (\Cref{sec:relatedWork}). \Cref{fig:kdeSynthesisProcedure} illustrates the synthesis procedure. Regulated by the bandwidth parameter $w$, new data points may be synthesised from the data's density estimation. Intuitively, $w$ regulates the estimator's bias-variance trade-off. Larger values lead to overly smooth estimates. In contrast, smaller values cause estimates that are too strongly influenced by the data's variance. To select a value for $w$, we must rely on empirical performance or heuristics that may produce sub-optimal results. We employed the Silverman rule of thumb, a popular heuristic~\cite{Belhaj2024}. In contrast, our conformal synthesis algorithm (\Cref{sec:proposal}) relies on the confidence of feature space regions, thresholded by the significance level $\epsilon$ (\Cref{fig:conformalSynthesisProcedure}). Unlike the bandwidth parameter $w$, $\epsilon$ provides a statistically meaningful boundary based on hypothesis testing, discussed in \Cref{sec:proposedAlgorithm}.

    % \FloatBarrier{}

            The synthesis performance was evaluated on the most difficult variant of each dataset with the conformal synthesis parameters used in \Cref{sec:realDatasets}:
            \begin{itemize}
                \item MNIST ($D_{500}$): $\epsilon=0.2$, $\gamma=0.01$.
                \item MSHRM ($D_{1:9}$): $\epsilon=0.1$, $\gamma=0.01$.
                \item WINE:\@ $\epsilon=0.2$, $\gamma=0.01$.
                \item USPS:\@ $\epsilon=0.1$, $\gamma=0.01$.
            \end{itemize}
            Since the number of synthesised samples follows from the conformal synthesis parameters, we synthesised the same number of samples from the KDE model to ensure a fair comparison. The sample statistics per dataset are summarised in \Cref{tab:kdeSynthesisSamples}, with class-conditional details in \Cref{sec:realDatasets}. All models were trained on the extended Train$_{\text{ext}}$ sets and evaluated on the same held-back test sets except for USPS models, which were trained on the synthesised samples Train$_{\text{syn}}$ instead.

            \begin{table}[!ht]
                \setlength{\tabcolsep}{2.8pt}
                \centering
                \captionsetup{width=\textwidth}
                \caption{Sample counts per dataset for the KDE comparison. Models were trained on Train$_{\text{ext}}$ as identified by the proposed conformal and KDE synthesis algorithms, except for the USPS dataset, which was evaluated on Train$_{\text{syn}}$.}\label{tab:kdeSynthesisSamples}
                \begin{tabular}{lrrrr}
                    \toprule
                    \multicolumn{1}{c}{} & \multicolumn{1}{c}{\textbf{Test}} & \multicolumn{1}{c}{\textbf{Train$_{\text{orig}}$}} & \multicolumn{1}{c}{\textbf{Train$_{\text{syn}}$}} & \multicolumn{1}{c}{\textbf{Train$_{\text{ext}}$}} \\
                    \midrule\midrule
                    \textbf{MNIST} & 10,000 & 500 & 1,213,241 & 1,213,741 \\
                    \textbf{MSHRM} & 2,681 & 3,143 & 1,688,440 & 1,691,583 \\
                    \textbf{WINE} & 1,072 & 2,176 & 60,318 & 62,494 \\
                    \textbf{USPS} & 2,007 & 7,291 & 406,251 & \multicolumn{1}{c}{---} \\
                    \bottomrule
                \end{tabular}
            \end{table}

\FloatBarrier{}

            \Cref{tab:kdeSynthesisResults} presents the mean results across five training iterations. While KDE tended to improve on the baseline models trained on the original datasets (\Cref{sec:realDatasets}), our confidently synthesised training sets outperformed KDE on all datasets. The likely cause was that the KDE bias-variance trade-off was sub-optimal, resulting in under- or overfitted density estimates. However, improving the regulating parameter $w$ is challenging because it does not have a transparent theoretical background compared to the parallel conformal $\epsilon$ (\Cref{sec:proposedAlgorithm}). The techniques' performance differences were most prominent on the MSHRM and WINE datasets (+53 and +28 percentage points F$_1$-score, respectively), indicating the potential of conformal synthesis for challenging datasets with imbalanced and overlapping classes in particular. The Wilcoxon test confirmed that all performance improvements were statistically significant.

            \begin{table}[!h]
                \setlength{\tabcolsep}{2.8pt}
                \centering
                \captionsetup{width=\textwidth}
                \caption{Comparison of conformal synthesis and density-based synthesis samples. Models were trained on Train$_{\text{ext}}$ generated by the proposed algorithm (CPS) and by KDE.\@ The mean and standard deviation across five iterations were reported with the best performance per dataset in bold. All CPS improvements were confirmed as statistically significant by the Wilcoxon test ($p_W < 0.1$), marked with *.}\label{tab:kdeSynthesisResults}
                \begin{tabular}{lrrrrrrrrr}
                    \toprule
                    & \multicolumn{3}{c}{\textbf{F$_{\mathbf{1}}$-score}} & \multicolumn{3}{c}{\textbf{Precision}} & \multicolumn{3}{c}{\textbf{Recall}} \\
                    \cmidrule(lr){2-4} \cmidrule(lr){5-7} \cmidrule(lr){8-10}
                    \textbf{} & \multicolumn{1}{c}{\textbf{CPS}} & \multicolumn{1}{c}{\textbf{KDE}} & \multicolumn{1}{c}{\textbf{$\mathbf{p_W}$}} & \multicolumn{1}{c}{\textbf{CPS}} & \multicolumn{1}{c}{\textbf{KDE}} & \multicolumn{1}{c}{\textbf{$\mathbf{p_W}$}} & \multicolumn{1}{c}{\textbf{CPS}} & \multicolumn{1}{c}{\textbf{KDE}} & \multicolumn{1}{c}{\textbf{$\mathbf{p_W}$}} \\
                    \midrule\midrule
                    \textbf{MNIST} & \textbf{0.79} (.07) & 0.62 (.08) & 0.00* & \textbf{0.80} (.00) & 0.66 (.10) & 0.02* & \textbf{0.79} (.01) & 0.67 (.05) & 0.00* \\
                    \textbf{MSHRM} & \textbf{0.95} (.01) & 0.42 (.08) & 0.00* & \textbf{0.95} (.01) & 0.65 (.21) & 0.02* & \textbf{0.95} (.01) & 0.55 (.04) & 0.00* \\
                    \textbf{WINE} & \textbf{0.71} (.00) & 0.43 (.00) & 0.00* & \textbf{0.72} (.04) & 0.38 (.00) & 0.00* & \textbf{0.76} (.00) & 0.50 (.00) & 0.00* \\
                    \textbf{USPS} & \textbf{0.92} (.00) & 0.85 (.04) & 0.01* & \textbf{0.92} (.01) & 0.88 (.03) & 0.03* & \textbf{0.92} (.00) & 0.84 (.04) & 0.01* \\
                    \bottomrule
                \end{tabular}
            \end{table}

\section{Practical advice and future work}\label{sec:discussion}
    Small sample counts, data imbalances, and overlapping classes are ubiquitous challenges that regularly reduce prediction performances on real-world datasets. Through extensive experimentation (\Cref{sec:results}), we have comprehensively showcased the potential of conformal synthesis to confidently generate new data points for these difficult datasets, ultimately improving prediction performance. It is important to note that the conformal confidence guarantees are somewhat limited by the extension to data synthesis (\Cref{sec:proposal}), and furthermore do not automatically guarantee downstream prediction performance. However, the extensive results analysis in this article indicates that optimising $\epsilon$ is closely paralleled by increasing model performance. An interesting avenue to explore in future would be to derive further guarantees about the synthesised data, and to combine conformal synthesis with traditional conformal predictors to investigate the interactions of confidence-aware data generation and prediction.

    Apart from the original data (e.g., the number of available samples $n$), the type and number of generated samples are significantly influenced by a range of synthesis parameters. The primary contributors are the grid step $\gamma$ and the significance level $\epsilon$. $\gamma$ can be interpreted as the resolution of the feature space, where the confidence of each grid point is evaluated for synthesis. Smaller values result in more grid points and, consequently, more potential synthesis samples (\Cref{sec:toyGridstepEffect}). Note that in this article, we synthesised all high-confidence grid points as synthetic samples. More sophisticated sampling techniques could be investigated in future to generate a pre-selected number of synthetic samples. In addition to $\gamma$, $\epsilon \in (0, 1)$ represents the confidence threshold above which grid points are included in the high-confidence regions for synthesis (\Cref{sec:toyDatasetSignificanceLevel}). Intuitively, smaller $\epsilon$ lead to larger confidence areas, increasing the synthesised sample count. In this article, we prioritised low $\epsilon$ to maximise the inclusion of correctly-labelled synthesis samples (\Cref{sec:proposedAlgorithm}). An alternative approach that merits further investigation would be to maximise $\epsilon$, generating an information-efficient synthetic representation of the original dataset by minimising the number of required samples. Potentially, this approach could be useful for confident data anonymisation tasks in the future.

    Additionally, the underlying non-conformity measure (NCM) may significantly influence the feature space confidence regions. Inherited from the Conformal Prediction framework, the NCM drives the proposed synthesis algorithm's capacity to effectively distinguish between low and high confidence regions without affecting conformal validity (\Cref{sec:cpBasics}). In this article, we employed a KNN-based NCM to evaluate feature space confidence as a robust baseline (\Cref{sec:dataAndSynthesisParams}). Future work could investigate more sophisticated definitions to improve synthesis and subsequent prediction performances.

    Theoretically, the proposed synthesis algorithm is compatible with any dimension and complexity of feature spaces. However, the confidence region computations may become prohibitively expensive for high-dimensional datasets (\Cref{eq:synthesisComplexity}, \Cref{sec:proposedAlgorithm}). Therefore, techniques to reduce the investigated feature space are vital for practical use. For example, this challenge may be addressed with dimensionality reduction techniques (e.g., UMAP), scaling features to a smaller range, and choosing an appropriate resolution of the feature space with the grid step $\gamma$. Avenues to improve the synthesis algorithm's complexity in the future include pre-selecting feature space regions of interest for which confidence scores are calculated rather than processing the entire space.

\FloatBarrier{}
\section{Conclusion}\label{sec:conclusion}
    In conclusion, we have presented a unique conformal data synthesis algorithm that utilises label-conditional feature space confidence for the data generation process. In addition to a systematic investigation of our proposal's parameters and characteristics, we presented extensive empirical experiments on five benchmark datasets. The comprehensive results demonstrated our algorithm's advantages for a variety of ubiquitous real-world challenges:
    \begin{itemize}
        \item Synthesising new data to significantly boost the sample size of small datasets, as well as correcting class imbalances,
        \item Supporting a model's learned feature space representations of non-separable classes,
        \item And replacing an entire dataset with synthetic samples, maintaining Deep Learning prediction performance.
    \end{itemize}
    While our proposed algorithm is capable of synthesising any data, generating associated ground truths is currently limited to classification labels and may be an interesting avenue for further extension.

\backmatter{}
\section*{Declarations}
    \begin{itemize}
        \item \textbf{Competing interests:} The authors declare that they have no conflict of interest.
        \item \textbf{Funding:} No funding was received for conducting this study.
        \item \textbf{Author contributions:} All authors contributed to the study's conception and design. J.M.\ carried out the experimentation, performed the analysis, and drafted the manuscript under K.N.'s supervision. All authors critically reviewed and approved the final manuscript.    % chktex 38
        \item \textbf{Availability of data and materials:} The benchmark datasets evaluated in this article are available from the UCI Machine Learning repository (MNIST dataset~\cite{lecun2021mnistDataset}, Mushroom dataset~\cite{schlimmer1987mushroomDataset}, Wine dataset~\cite{Cortez2009WineDataset}) and from Kaggle (USPS dataset~\cite{hull1994uspsDataset}).
        \item \textbf{Code availability:} The implementation is available at \url{https://github.com/juliameister/conformalised-data-synthesis}.
        \item \textbf{Ethics approval:} Not applicable.
        \item \textbf{Consent to participate:} Not applicable.
        \item \textbf{Consent for publication:} Not applicable.
    \end{itemize}

\bibliography{sn-bibliography}


\begin{thebibliography}{76}
\ifx \bisbn   \undefined \def \bisbn  #1{ISBN #1}\fi
\ifx \binits  \undefined \def \binits#1{#1}\fi
\ifx \bauthor  \undefined \def \bauthor#1{#1}\fi
\ifx \batitle  \undefined \def \batitle#1{#1}\fi
\ifx \bjtitle  \undefined \def \bjtitle#1{#1}\fi
\ifx \bvolume  \undefined \def \bvolume#1{\textbf{#1}}\fi
\ifx \byear  \undefined \def \byear#1{#1}\fi
\ifx \bissue  \undefined \def \bissue#1{#1}\fi
\ifx \bfpage  \undefined \def \bfpage#1{#1}\fi
\ifx \blpage  \undefined \def \blpage #1{#1}\fi
\ifx \burl  \undefined \def \burl#1{\textsf{#1}}\fi
\ifx \doiurl  \undefined \def \doiurl#1{\url{https://doi.org/#1}}\fi
\ifx \betal  \undefined \def \betal{\textit{et al.}}\fi
\ifx \binstitute  \undefined \def \binstitute#1{#1}\fi
\ifx \binstitutionaled  \undefined \def \binstitutionaled#1{#1}\fi
\ifx \bctitle  \undefined \def \bctitle#1{#1}\fi
\ifx \beditor  \undefined \def \beditor#1{#1}\fi
\ifx \bpublisher  \undefined \def \bpublisher#1{#1}\fi
\ifx \bbtitle  \undefined \def \bbtitle#1{#1}\fi
\ifx \bedition  \undefined \def \bedition#1{#1}\fi
\ifx \bseriesno  \undefined \def \bseriesno#1{#1}\fi
\ifx \blocation  \undefined \def \blocation#1{#1}\fi
\ifx \bsertitle  \undefined \def \bsertitle#1{#1}\fi
\ifx \bsnm \undefined \def \bsnm#1{#1}\fi
\ifx \bsuffix \undefined \def \bsuffix#1{#1}\fi
\ifx \bparticle \undefined \def \bparticle#1{#1}\fi
\ifx \barticle \undefined \def \barticle#1{#1}\fi
\bibcommenthead
\ifx \bconfdate \undefined \def \bconfdate #1{#1}\fi
\ifx \botherref \undefined \def \botherref #1{#1}\fi
\ifx \url \undefined \def \url#1{\textsf{#1}}\fi
\ifx \bchapter \undefined \def \bchapter#1{#1}\fi
\ifx \bbook \undefined \def \bbook#1{#1}\fi
\ifx \bcomment \undefined \def \bcomment#1{#1}\fi
\ifx \oauthor \undefined \def \oauthor#1{#1}\fi
\ifx \citeauthoryear \undefined \def \citeauthoryear#1{#1}\fi
\ifx \endbibitem  \undefined \def \endbibitem {}\fi
\ifx \bconflocation  \undefined \def \bconflocation#1{#1}\fi
\ifx \arxivurl  \undefined \def \arxivurl#1{\textsf{#1}}\fi
\csname PreBibitemsHook\endcsname

%%% 1
\bibitem[\protect\citeauthoryear{Brigato and Iocchi}{2021}]{Brigato2021}
\begin{bchapter}
\bauthor{\bsnm{Brigato}, \binits{L.}},
\bauthor{\bsnm{Iocchi}, \binits{L.}}:
\bctitle{{A close look at Deep Learning with small data}}.
In: \bbtitle{2020 25th International Conference on Pattern Recognition (ICPR)},
pp. \bfpage{2490}--\blpage{2497}.
\bpublisher{IEEE},
\blocation{New York}
(\byear{2021}).
\doiurl{10.1109/ICPR48806.2021.9412492}
\end{bchapter}
\endbibitem

%%% 2
\bibitem[\protect\citeauthoryear{Moreno-Barea et~al.}{2020}]{Moreno-Barea2020}
\begin{barticle}
\bauthor{\bsnm{Moreno-Barea}, \binits{F.J.}},
\bauthor{\bsnm{Jerez}, \binits{J.M.}},
\bauthor{\bsnm{Franco}, \binits{L.}}:
\batitle{{Improving classification accuracy using data augmentation on small data sets}}.
\bjtitle{Expert Systems with Applications}
\bvolume{161}(\bissue{0957-4174}),
\bfpage{113696}
(\byear{2020})
\doiurl{10.1016/j.eswa.2020.113696}
\end{barticle}
\endbibitem

%%% 3
\bibitem[\protect\citeauthoryear{Sarker}{2021}]{Sarker2021}
\begin{barticle}
\bauthor{\bsnm{Sarker}, \binits{I.H.}}:
\batitle{{Deep Learning: A comprehensive overview on techniques, taxonomy, applications and research directions}}.
\bjtitle{SN Computer Science}
\bvolume{2}(\bissue{6}),
\bfpage{420}
(\byear{2021})
\doiurl{10.1007/s42979-021-00815-1}
\end{barticle}
\endbibitem

%%% 4
\bibitem[\protect\citeauthoryear{Zhuang et~al.}{2019}]{Zhuang2019}
\begin{bchapter}
\bauthor{\bsnm{Zhuang}, \binits{P.}},
\bauthor{\bsnm{Schwing}, \binits{A.G.}},
\bauthor{\bsnm{Koyejo}, \binits{O.}}:
\bctitle{{FMRI data augmentation via synthesis}}.
In: \bbtitle{2019 IEEE 16th International Symposium on Biomedical Imaging (ISBI 2019)},
vol. \bseriesno{2019-April},
pp. \bfpage{1783}--\blpage{1787}.
\bpublisher{IEEE},
\blocation{New York}
(\byear{2019}).
\doiurl{10.1109/ISBI.2019.8759585}
\end{bchapter}
\endbibitem

%%% 5
\bibitem[\protect\citeauthoryear{Liu et~al.}{2022}]{Liu2022dataSynthesis}
\begin{barticle}
\bauthor{\bsnm{Liu}, \binits{S.}},
\bauthor{\bsnm{Jiang}, \binits{H.}},
\bauthor{\bsnm{Wu}, \binits{Z.}},
\bauthor{\bsnm{Li}, \binits{X.}}:
\batitle{{Data synthesis using deep feature enhanced Generative Adversarial Networks for rolling bearing imbalanced fault diagnosis}}.
\bjtitle{Mechanical Systems and Signal Processing}
\bvolume{163},
\bfpage{108139}
(\byear{2022})
\doiurl{10.1016/j.ymssp.2021.108139}
\end{barticle}
\endbibitem

%%% 6
\bibitem[\protect\citeauthoryear{Aggarwal et~al.}{2021}]{Aggarwal2021}
\begin{botherref}
\oauthor{\bsnm{Aggarwal}, \binits{A.}},
\oauthor{\bsnm{Mittal}, \binits{M.}},
\oauthor{\bsnm{Battineni}, \binits{G.}}:
{Generative Adversarial Network: An overview of theory and applications}.
International Journal of Information Management Data Insights
\textbf{1}(1)
(2021)
\doiurl{10.1016/j.jjimei.2020.100004}
\end{botherref}
\endbibitem

%%% 7
\bibitem[\protect\citeauthoryear{Grnarova et~al.}{2019}]{Grnarova2019}
\begin{bchapter}
\bauthor{\bsnm{Grnarova}, \binits{P.}},
\bauthor{\bsnm{Levy}, \binits{K.Y.}},
\bauthor{\bsnm{Lucchi}, \binits{A.}},
\bauthor{\bsnm{Perraudin}, \binits{N.}},
\bauthor{\bsnm{Goodfellow}, \binits{I.}},
\bauthor{\bsnm{Hofmann}, \binits{T.}},
\bauthor{\bsnm{Krause}, \binits{A.}}:
\bctitle{{A domain agnostic measure for monitoring and evaluating GANs}}.
In: \bbtitle{Proceedings of the 32nd Advances in Neural Information Processing Systems Conference (NeurIPS)}.
\bpublisher{Curran Associates, Inc.},
\blocation{New York}
(\byear{2019}).
\burl{https://proceedings.neurips.cc/paper_files/paper/2019/hash/692baebec3bb4b53d7ebc3b9fabac31b-Abstract.html}
Accessed 14/10/2024
\end{bchapter}
\endbibitem

%%% 8
\bibitem[\protect\citeauthoryear{Shafer and Vovk}{2008}]{shafer2008tutorial}
\begin{barticle}
\bauthor{\bsnm{Shafer}, \binits{G.}},
\bauthor{\bsnm{Vovk}, \binits{V.}}:
\batitle{{A tutorial on Conformal Prediction}}.
\bjtitle{Journal of Machine Learning Research}
\bvolume{9}(\bissue{3}),
\bfpage{371}--\blpage{421}
(\byear{2008})
\end{barticle}
\endbibitem

%%% 9
\bibitem[\protect\citeauthoryear{Cherubin et~al.}{2015}]{Cherubin2015}
\begin{bchapter}
\bauthor{\bsnm{Cherubin}, \binits{G.}},
\bauthor{\bsnm{Nouretdinov}, \binits{I.}},
\bauthor{\bsnm{Gammerman}, \binits{A.}},
\bauthor{\bsnm{Jordaney}, \binits{R.}},
\bauthor{\bsnm{Wang}, \binits{Z.}},
\bauthor{\bsnm{Papini}, \binits{D.}},
\bauthor{\bsnm{Cavallaro}, \binits{L.}}:
\bctitle{{Conformal clustering and its application to botnet traffic}}.
In: \bbtitle{2015 Statistical Learning and Data Sciences (SLDS): Lecture Notes in Computer Science},
vol. \bseriesno{9047},
pp. \bfpage{313}--\blpage{322}.
\bpublisher{Springer},
\blocation{Cham}
(\byear{2015}).
\doiurl{10.1007/978-3-319-17091-6_26}
\end{bchapter}
\endbibitem

%%% 10
\bibitem[\protect\citeauthoryear{Althnian et~al.}{2021}]{Althnian2021}
\begin{barticle}
\bauthor{\bsnm{Althnian}, \binits{A.}},
\bauthor{\bsnm{AlSaeed}, \binits{D.}},
\bauthor{\bsnm{Al-Baity}, \binits{H.}},
\bauthor{\bsnm{Samha}, \binits{A.}},
\bauthor{\bsnm{Dris}, \binits{A.B.}},
\bauthor{\bsnm{Alzakari}, \binits{N.}},
\bauthor{\bsnm{{Abou Elwafa}}, \binits{A.}},
\bauthor{\bsnm{Kurdi}, \binits{H.}}:
\batitle{{Impact of dataset size on classification performance: An empirical evaluation in the medical domain}}.
\bjtitle{Applied Sciences}
\bvolume{11},
\bfpage{1}--\blpage{18}
(\byear{2021})
\doiurl{10.3390/app11020796}
\end{barticle}
\endbibitem

%%% 11
\bibitem[\protect\citeauthoryear{Alauthman et~al.}{2023}]{Alauthman2023}
\begin{botherref}
\oauthor{\bsnm{Alauthman}, \binits{M.}},
\oauthor{\bsnm{Al-qerem}, \binits{A.}},
\oauthor{\bsnm{Sowan}, \binits{B.}},
\oauthor{\bsnm{Alsarhan}, \binits{A.}},
\oauthor{\bsnm{Eshtay}, \binits{M.}},
\oauthor{\bsnm{Aldweesh}, \binits{A.}},
\oauthor{\bsnm{Aslam}, \binits{N.}}:
{Enhancing small medical dataset classification performance using GAN}.
Informatics
\textbf{10}(1)
(2023)
\doiurl{10.3390/informatics10010028}
\end{botherref}
\endbibitem

%%% 12
\bibitem[\protect\citeauthoryear{Muramatsu et~al.}{2020}]{Muramatsu2020}
\begin{barticle}
\bauthor{\bsnm{Muramatsu}, \binits{C.}},
\bauthor{\bsnm{Nishio}, \binits{M.}},
\bauthor{\bsnm{Goto}, \binits{T.}},
\bauthor{\bsnm{Oiwa}, \binits{M.}},
\bauthor{\bsnm{Morita}, \binits{T.}},
\bauthor{\bsnm{Yakami}, \binits{M.}},
\bauthor{\bsnm{Kubo}, \binits{T.}},
\bauthor{\bsnm{Togashi}, \binits{K.}},
\bauthor{\bsnm{Fujita}, \binits{H.}}:
\batitle{{Improving breast mass classification by shared data with domain transformation using a Generative Adversarial Network}}.
\bjtitle{Computers in Biology and Medicine}
\bvolume{119},
\bfpage{103698}
(\byear{2020})
\doiurl{10.1016/j.compbiomed.2020.103698}
\end{barticle}
\endbibitem

%%% 13
\bibitem[\protect\citeauthoryear{Salazar et~al.}{2021}]{Salazar2021}
\begin{barticle}
\bauthor{\bsnm{Salazar}, \binits{A.}},
\bauthor{\bsnm{Vergara}, \binits{L.}},
\bauthor{\bsnm{Safont}, \binits{G.}}:
\batitle{{Generative Adversarial Networks and Markov Random Fields for oversampling very small training sets}}.
\bjtitle{Expert Systems with Applications}
\bvolume{163},
\bfpage{113819}
(\byear{2021})
\doiurl{10.1016/j.eswa.2020.113819}
\end{barticle}
\endbibitem

%%% 14
\bibitem[\protect\citeauthoryear{Koshino et~al.}{2021}]{Koshino2021}
\begin{barticle}
\bauthor{\bsnm{Koshino}, \binits{K.}},
\bauthor{\bsnm{Werner}, \binits{R.A.}},
\bauthor{\bsnm{Pomper}, \binits{M.G.}},
\bauthor{\bsnm{Bundschuh}, \binits{R.A.}},
\bauthor{\bsnm{Toriumi}, \binits{F.}},
\bauthor{\bsnm{Higuchi}, \binits{T.}},
\bauthor{\bsnm{Rowe}, \binits{S.P.}}:
\batitle{{Narrative review of Generative Adversarial Networks in medical and molecular imaging}}.
\bjtitle{Annals of Translational Medicine}
\bvolume{9}(\bissue{9}),
\bfpage{821}--\blpage{821}
(\byear{2021})
\doiurl{10.21037/atm-20-6325}
\end{barticle}
\endbibitem

%%% 15
\bibitem[\protect\citeauthoryear{Li et~al.}{2020}]{Li2020}
\begin{bchapter}
\bauthor{\bsnm{Li}, \binits{Q.}},
\bauthor{\bsnm{Zheng}, \binits{Z.}},
\bauthor{\bsnm{Wu}, \binits{F.}},
\bauthor{\bsnm{Chen}, \binits{G.}}:
\bctitle{{Generative Adversarial Networks-based privacy-preserving 3D reconstruction}}.
In: \bbtitle{2020 IEEE/ACM 28th International Symposium on Quality of Service (IWQoS)},
pp. \bfpage{1}--\blpage{10}.
\bpublisher{IEEE},
\blocation{New York}
(\byear{2020}).
\doiurl{10.1109/IWQoS49365.2020.9213037}
\end{bchapter}
\endbibitem

%%% 16
\bibitem[\protect\citeauthoryear{Yoon et~al.}{2020}]{Yoon2020}
\begin{barticle}
\bauthor{\bsnm{Yoon}, \binits{J.}},
\bauthor{\bsnm{Drumright}, \binits{L.N.}},
\bauthor{\bsnm{Schaar}, \binits{M.}}:
\batitle{{Anonymization through data synthesis using Generative Adversarial Networks (ADS-GAN)}}.
\bjtitle{IEEE Journal of Biomedical and Health Informatics}
\bvolume{24}(\bissue{8}),
\bfpage{2378}--\blpage{2388}
(\byear{2020})
\doiurl{10.1109/JBHI.2020.2980262}
\end{barticle}
\endbibitem

%%% 17
\bibitem[\protect\citeauthoryear{Thambawita et~al.}{2021}]{Thambawita2021}
\begin{barticle}
\bauthor{\bsnm{Thambawita}, \binits{V.}},
\bauthor{\bsnm{Isaksen}, \binits{J.L.}},
\bauthor{\bsnm{Hicks}, \binits{S.A.}},
\bauthor{\bsnm{Ghouse}, \binits{J.}},
\bauthor{\bsnm{Ahlberg}, \binits{G.}},
\bauthor{\bsnm{Linneberg}, \binits{A.}},
\bauthor{\bsnm{Grarup}, \binits{N.}},
\bauthor{\bsnm{Ellervik}, \binits{C.}},
\bauthor{\bsnm{Olesen}, \binits{M.S.}},
\bauthor{\bsnm{Hansen}, \binits{T.}},
\bauthor{\bsnm{Graff}, \binits{C.}},
\bauthor{\bsnm{Holstein-Rathlou}, \binits{N.-H.}},
\bauthor{\bsnm{Str{\"{u}}mke}, \binits{I.}},
\bauthor{\bsnm{Hammer}, \binits{H.L.}},
\bauthor{\bsnm{Maleckar}, \binits{M.M.}},
\bauthor{\bsnm{Halvorsen}, \binits{P.}},
\bauthor{\bsnm{Riegler}, \binits{M.A.}},
\bauthor{\bsnm{Kanters}, \binits{J.K.}}:
\batitle{{DeepFake electrocardiograms using Generative Adversarial Networks are the beginning of the end for privacy issues in medicine}}.
\bjtitle{Scientific Reports}
\bvolume{11}(\bissue{1}),
\bfpage{21896}
(\byear{2021})
\doiurl{10.1038/s41598-021-01295-2}
\end{barticle}
\endbibitem

%%% 18
\bibitem[\protect\citeauthoryear{Bashir et~al.}{2020}]{Bashir2020}
\begin{bchapter}
\bauthor{\bsnm{Bashir}, \binits{D.}},
\bauthor{\bsnm{Monta{\~{n}}ez}, \binits{G.D.}},
\bauthor{\bsnm{Sehra}, \binits{S.}},
\bauthor{\bsnm{Segura}, \binits{P.S.}},
\bauthor{\bsnm{Lauw}, \binits{J.}}:
\bctitle{{An information-theoretic perspective on overfitting and underfitting}}.
In: \bbtitle{AI 2020: AI 2020: Advances in Artificial Intelligence}
vol. \bseriesno{12576 LNAI},
pp. \bfpage{347}--\blpage{358}.
\bpublisher{Springer},
\blocation{Cham}
(\byear{2020}).
\doiurl{10.1007/978-3-030-64984-5_27}
\end{bchapter}
\endbibitem

%%% 19
\bibitem[\protect\citeauthoryear{Bejani and Ghatee}{2021}]{Bejani2021}
\begin{barticle}
\bauthor{\bsnm{Bejani}, \binits{M.M.}},
\bauthor{\bsnm{Ghatee}, \binits{M.}}:
\batitle{{A systematic review on overfitting control in shallow and deep Neural Networks}}.
\bjtitle{Artificial Intelligence Review}
\bvolume{54}(\bissue{8}),
\bfpage{6391}--\blpage{6438}
(\byear{2021})
\doiurl{10.1007/s10462-021-09975-1}
\end{barticle}
\endbibitem

%%% 20
\bibitem[\protect\citeauthoryear{Whang et~al.}{2023}]{Whang2023}
\begin{barticle}
\bauthor{\bsnm{Whang}, \binits{S.E.}},
\bauthor{\bsnm{Roh}, \binits{Y.}},
\bauthor{\bsnm{Song}, \binits{H.}},
\bauthor{\bsnm{Lee}, \binits{J.-G.}}:
\batitle{{Data collection and quality challenges in Deep Learning: A data-centric AI perspective}}.
\bjtitle{The VLDB Journal}
\bvolume{32}(\bissue{4}),
\bfpage{791}--\blpage{813}
(\byear{2023})
\doiurl{10.1007/s00778-022-00775-9}
\end{barticle}
\endbibitem

%%% 21
\bibitem[\protect\citeauthoryear{Goodfellow et~al.}{2020}]{Goodfellow2020}
\begin{barticle}
\bauthor{\bsnm{Goodfellow}, \binits{I.}},
\bauthor{\bsnm{Pouget-Abadie}, \binits{J.}},
\bauthor{\bsnm{Mirza}, \binits{M.}},
\bauthor{\bsnm{Xu}, \binits{B.}},
\bauthor{\bsnm{Warde-Farley}, \binits{D.}},
\bauthor{\bsnm{Ozair}, \binits{S.}},
\bauthor{\bsnm{Courville}, \binits{A.}},
\bauthor{\bsnm{Bengio}, \binits{Y.}}:
\batitle{{Generative Adversarial Networks}}.
\bjtitle{Communications of the ACM}
\bvolume{63}(\bissue{11}),
\bfpage{139}--\blpage{144}
(\byear{2020})
\doiurl{10.1145/3422622}
{\href{https://arxiv.org/abs/1406.2661}{{arXiv:1406.2661}}}
\end{barticle}
\endbibitem

%%% 22
\bibitem[\protect\citeauthoryear{Kammoun et~al.}{2022}]{Kammoun2023}
\begin{barticle}
\bauthor{\bsnm{Kammoun}, \binits{A.}},
\bauthor{\bsnm{Slama}, \binits{R.}},
\bauthor{\bsnm{Tabia}, \binits{H.}},
\bauthor{\bsnm{Ouni}, \binits{T.}},
\bauthor{\bsnm{Abid}, \binits{M.}}:
\batitle{{Generative Adversarial Networks for face generation: A survey}}.
\bjtitle{ACM Computing Surveys}
\bvolume{55}(\bissue{5}),
\bfpage{1}--\blpage{37}
(\byear{2022})
\doiurl{10.1145/3527850}
\end{barticle}
\endbibitem

%%% 23
\bibitem[\protect\citeauthoryear{Borji}{2022}]{Borji2022}
\begin{barticle}
\bauthor{\bsnm{Borji}, \binits{A.}}:
\batitle{{Pros and cons of GAN evaluation measures: New developments}}.
\bjtitle{Computer Vision and Image Understanding}
\bvolume{215},
\bfpage{103329}
(\byear{2022})
\doiurl{10.1016/j.cviu.2021.103329}
\end{barticle}
\endbibitem

%%% 24
\bibitem[\protect\citeauthoryear{Navidan et~al.}{2021}]{Navidan2021}
\begin{barticle}
\bauthor{\bsnm{Navidan}, \binits{H.}},
\bauthor{\bsnm{Moshiri}, \binits{P.F.}},
\bauthor{\bsnm{Nabati}, \binits{M.}},
\bauthor{\bsnm{Shahbazian}, \binits{R.}},
\bauthor{\bsnm{Ghorashi}, \binits{S.A.}},
\bauthor{\bsnm{Shah-Mansouri}, \binits{V.}},
\bauthor{\bsnm{Windridge}, \binits{D.}}:
\batitle{{Generative Adversarial Networks (GANs) in networking: A comprehensive survey and evaluation}}.
\bjtitle{Computer Networks}
\bvolume{194},
\bfpage{108149}
(\byear{2021})
\doiurl{10.1016/j.comnet.2021.108149}
\end{barticle}
\endbibitem

%%% 25
\bibitem[\protect\citeauthoryear{Brophy et~al.}{2023}]{Brophy2023}
\begin{barticle}
\bauthor{\bsnm{Brophy}, \binits{E.}},
\bauthor{\bsnm{Wang}, \binits{Z.}},
\bauthor{\bsnm{She}, \binits{Q.}},
\bauthor{\bsnm{Ward}, \binits{T.}}:
\batitle{{Generative Adversarial Networks in time series: A systematic literature review}}.
\bjtitle{ACM Computing Surveys}
\bvolume{55}(\bissue{10}),
\bfpage{1}--\blpage{31}
(\byear{2023})
\doiurl{10.1145/3559540}
\end{barticle}
\endbibitem

%%% 26
\bibitem[\protect\citeauthoryear{Saxena and Cao}{2022}]{Saxena2022}
\begin{barticle}
\bauthor{\bsnm{Saxena}, \binits{D.}},
\bauthor{\bsnm{Cao}, \binits{J.}}:
\batitle{{Generative Adversarial Networks (GANs): Challenges, solutions, and future directions}}.
\bjtitle{ACM Computing Surveys}
\bvolume{54}(\bissue{3}),
\bfpage{1}--\blpage{42}
(\byear{2022})
\doiurl{10.1145/3446374}
\end{barticle}
\endbibitem

%%% 27
\bibitem[\protect\citeauthoryear{Bhattarai et~al.}{2020}]{Bhattarai2020}
\begin{bchapter}
\bauthor{\bsnm{Bhattarai}, \binits{B.}},
\bauthor{\bsnm{Baek}, \binits{S.}},
\bauthor{\bsnm{Bodur}, \binits{R.}},
\bauthor{\bsnm{Kim}, \binits{T.-K.}}:
\bctitle{{Sampling strategies for GAN synthetic data}}.
In: \bbtitle{ICASSP 2020 - 2020 IEEE International Conference on Acoustics, Speech and Signal Processing (ICASSP)},
pp. \bfpage{2303}--\blpage{2307}.
\bpublisher{IEEE},
\blocation{New York}
(\byear{2020}).
\doiurl{10.1109/ICASSP40776.2020.9054677}
\end{bchapter}
\endbibitem

%%% 28
\bibitem[\protect\citeauthoryear{Nie and Shen}{2020}]{Nie2020}
\begin{barticle}
\bauthor{\bsnm{Nie}, \binits{D.}},
\bauthor{\bsnm{Shen}, \binits{D.}}:
\batitle{{Adversarial confidence learning for medical image segmentation and synthesis}}.
\bjtitle{International Journal of Computer Vision}
\bvolume{128}(\bissue{10-11}),
\bfpage{2494}--\blpage{2513}
(\byear{2020})
\doiurl{10.1007/s11263-020-01321-2}
\end{barticle}
\endbibitem

%%% 29
\bibitem[\protect\citeauthoryear{Du et~al.}{2022}]{Du2022}
\begin{bchapter}
\bauthor{\bsnm{Du}, \binits{Y.}},
\bauthor{\bsnm{Quan}, \binits{Q.}},
\bauthor{\bsnm{Han}, \binits{H.}},
\bauthor{\bsnm{Zhou}, \binits{S.K.}}:
\bctitle{{Semi-supervised pseudo-healthy image synthesis via confidence augmentation}}.
In: \bbtitle{2022 IEEE 19th International Symposium on Biomedical Imaging (ISBI)},
pp. \bfpage{1}--\blpage{4}.
\bpublisher{IEEE},
\blocation{New York}
(\byear{2022}).
\doiurl{10.1109/ISBI52829.2022.9761522}
\end{bchapter}
\endbibitem

%%% 30
\bibitem[\protect\citeauthoryear{Abdusalomov et~al.}{2023}]{Abdusalomov2023}
\begin{botherref}
\oauthor{\bsnm{Abdusalomov}, \binits{A.B.}},
\oauthor{\bsnm{Nasimov}, \binits{R.}},
\oauthor{\bsnm{Nasimova}, \binits{N.}},
\oauthor{\bsnm{Muminov}, \binits{B.}},
\oauthor{\bsnm{Whangbo}, \binits{T.K.}}:
Evaluating synthetic medical images using {Artificial Intelligence} with the {GAN} algorithm.
Sensors
\textbf{23}(7)
(2023)
\doiurl{10.3390/s23073440}
\end{botherref}
\endbibitem

%%% 31
\bibitem[\protect\citeauthoryear{Wang et~al.}{2023}]{Wang2023}
\begin{botherref}
\oauthor{\bsnm{Wang}, \binits{J.}},
\oauthor{\bsnm{Xie}, \binits{G.}},
\oauthor{\bsnm{Huang}, \binits{Y.}},
\oauthor{\bsnm{Lyu}, \binits{J.}},
\oauthor{\bsnm{Zheng}, \binits{F.}},
\oauthor{\bsnm{Zheng}, \binits{Y.}},
\oauthor{\bsnm{Jin}, \binits{Y.}}:
{FedMed-GAN: Federated domain translation on unsupervised cross-modality brain image synthesis}.
Neurocomputing
\textbf{546}
(2023)
\doiurl{10.1016/j.neucom.2023.126282}
\end{botherref}
\endbibitem

%%% 32
\bibitem[\protect\citeauthoryear{Shi et~al.}{2023}]{Shi2023}
\begin{barticle}
\bauthor{\bsnm{Shi}, \binits{J.}},
\bauthor{\bsnm{Liu}, \binits{W.}},
\bauthor{\bsnm{Zhou}, \binits{G.}},
\bauthor{\bsnm{Zhou}, \binits{Y.}}:
\batitle{{AutoInfo GAN: Toward a better image synthesis GAN framework for high-fidelity few-shot datasets via NAS and contrastive learning}}.
\bjtitle{Knowledge-Based Systems}
\bvolume{276},
\bfpage{110757}
(\byear{2023})
\doiurl{10.1016/j.knosys.2023.110757}
\end{barticle}
\endbibitem

%%% 33
\bibitem[\protect\citeauthoryear{Zhao and Bilen}{2022}]{zhao2022synthesizing}
\begin{bchapter}
\bauthor{\bsnm{Zhao}, \binits{B.}},
\bauthor{\bsnm{Bilen}, \binits{H.}}:
\bctitle{{Synthesizing informative training samples with GAN}}.
In: \bbtitle{NeurIPS 2022 Workshop on Synthetic Data for Empowering ML Research}
(\byear{2022}).
\burl{https://openreview.net/forum?id=frAv0jtUMfS}
Accessed 14/10/2024
\end{bchapter}
\endbibitem

%%% 34
\bibitem[\protect\citeauthoryear{Huang and Jafari}{2023}]{Huang2023}
\begin{botherref}
\oauthor{\bsnm{Huang}, \binits{G.}},
\oauthor{\bsnm{Jafari}, \binits{A.H.}}:
{Enhanced balancing GAN: Minority-class image generation}.
Neural Computing and Applications
\textbf{35}(7)
(2023)
\doiurl{10.1007/s00521-021-06163-8}
\end{botherref}
\endbibitem

%%% 35
\bibitem[\protect\citeauthoryear{Plesovskaya and Ivanov}{2021}]{Plesovskaya2021}
\begin{barticle}
\bauthor{\bsnm{Plesovskaya}, \binits{E.}},
\bauthor{\bsnm{Ivanov}, \binits{S.}}:
\batitle{{An empirical analysis of KDE-based generative models on small datasets}}.
\bjtitle{Procedia Computer Science}
\bvolume{193},
\bfpage{442}--\blpage{452}
(\byear{2021})
\doiurl{10.1016/j.procs.2021.10.046}
\end{barticle}
\endbibitem

%%% 36
\bibitem[\protect\citeauthoryear{Park and Pardalos}{2024}]{Park2024}
\begin{botherref}
\oauthor{\bsnm{Park}, \binits{S.}},
\oauthor{\bsnm{Pardalos}, \binits{P.M.}}:
{Deep data density estimation through Donsker-Varadhan representation}.
Annals of Mathematics and Artificial Intelligence,
1--11
(2024)
\doiurl{10.1007/s10472-024-09943-9}
\end{botherref}
\endbibitem

%%% 37
\bibitem[\protect\citeauthoryear{Bauer et~al.}{2024}]{Bauer2024}
\begin{botherref}
\oauthor{\bsnm{Bauer}, \binits{A.}},
\oauthor{\bsnm{Trapp}, \binits{S.}},
\oauthor{\bsnm{Stenger}, \binits{M.}},
\oauthor{\bsnm{Leppich}, \binits{R.}},
\oauthor{\bsnm{Kounev}, \binits{S.}},
\oauthor{\bsnm{Leznik}, \binits{M.}},
\oauthor{\bsnm{Chard}, \binits{K.}},
\oauthor{\bsnm{Foster}, \binits{I.}}:
{Comprehensive exploration of synthetic data generation: A survey}
(2024).
\url{http://arxiv.org/abs/2401.02524}
Accessed 14/10/2024
\end{botherref}
\endbibitem

%%% 38
\bibitem[\protect\citeauthoryear{Belhaj}{2024}]{Belhaj2024}
\begin{barticle}
\bauthor{\bsnm{Belhaj}, \binits{E.I.}}:
\batitle{{A modified rule-of-thumb method for kernel density estimation}}.
\bjtitle{Journal of Mathematical Problems, Equations and Statistics}
\bvolume{5}(\bissue{1}),
\bfpage{143}--\blpage{149}
(\byear{2024}).
Accessed 14/10/2024
\end{barticle}
\endbibitem

%%% 39
\bibitem[\protect\citeauthoryear{Pozi and Omar}{2020}]{Pozi2020}
\begin{barticle}
\bauthor{\bsnm{Pozi}, \binits{M.S.M.}},
\bauthor{\bsnm{Omar}, \binits{M.H.}}:
\batitle{{A kernel density estimation method to generate synthetic shifted datasets in privacy-preserving task}}.
\bjtitle{Journal of Internet Services and Information Security}
\bvolume{10}(\bissue{4}),
\bfpage{70}--\blpage{89}
(\byear{2020})
\doiurl{10.22667/JISIS.2020.11.30.070}
\end{barticle}
\endbibitem

%%% 40
\bibitem[\protect\citeauthoryear{Falxa et~al.}{2023}]{Falxa2023}
\begin{botherref}
\oauthor{\bsnm{Falxa}, \binits{M.}},
\oauthor{\bsnm{Babak}, \binits{S.}},
\oauthor{\bsnm{{Le Jeune}}, \binits{M.}}:
{Adaptive kernel density estimation proposal in gravitational wave data analysis}.
Physical Review D
\textbf{107}(2)
(2023)
\doiurl{10.1103/PhysRevD.107.022008}
\end{botherref}
\endbibitem

%%% 41
\bibitem[\protect\citeauthoryear{Liu et~al.}{2021}]{Liu2021}
\begin{barticle}
\bauthor{\bsnm{Liu}, \binits{L.}},
\bauthor{\bsnm{Zhan}, \binits{X.}},
\bauthor{\bsnm{Wu}, \binits{R.}},
\bauthor{\bsnm{Guan}, \binits{X.}},
\bauthor{\bsnm{Wang}, \binits{Z.}},
\bauthor{\bsnm{Zhang}, \binits{W.}},
\bauthor{\bsnm{Pilanci}, \binits{M.}},
\bauthor{\bsnm{Wang}, \binits{Y.}},
\bauthor{\bsnm{Luo}, \binits{Z.}},
\bauthor{\bsnm{Li}, \binits{G.}}:
\batitle{{Boost AI power: Data augmentation strategies with unlabeled data and Conformal Prediction, A case in alternative herbal medicine discrimination with electronic nose}}.
\bjtitle{IEEE Sensors Journal}
\bvolume{21}(\bissue{20}),
\bfpage{22995}--\blpage{23005}
(\byear{2021})
\doiurl{10.1109/JSEN.2021.3102488}
\end{barticle}
\endbibitem

%%% 42
\bibitem[\protect\citeauthoryear{Liu et~al.}{2022}]{Liu2022}
\begin{barticle}
\bauthor{\bsnm{Liu}, \binits{L.}},
\bauthor{\bsnm{Zhan}, \binits{X.}},
\bauthor{\bsnm{Yang}, \binits{X.}},
\bauthor{\bsnm{Guan}, \binits{X.}},
\bauthor{\bsnm{Wu}, \binits{R.}},
\bauthor{\bsnm{Wang}, \binits{Z.}},
\bauthor{\bsnm{Luo}, \binits{Z.}},
\bauthor{\bsnm{Wang}, \binits{Y.}},
\bauthor{\bsnm{Li}, \binits{G.}}:
\batitle{{CPSC: Conformal Prediction with shrunken centroids for efficient prediction reliability quantification and data augmentation, A case in alternative herbal medicine classification with electronic nose}}.
\bjtitle{IEEE Transactions on Instrumentation and Measurement}
\bvolume{71},
\bfpage{1}--\blpage{11}
(\byear{2022})
\doiurl{10.1109/TIM.2021.3134321}
\end{barticle}
\endbibitem

%%% 43
\bibitem[\protect\citeauthoryear{Zhang et~al.}{2021}]{Zhang2021}
\begin{barticle}
\bauthor{\bsnm{Zhang}, \binits{J.}},
\bauthor{\bsnm{Norinder}, \binits{U.}},
\bauthor{\bsnm{Svensson}, \binits{F.}}:
\batitle{{Deep Learning-based Conformal Prediction of toxicity}}.
\bjtitle{Journal of Chemical Information and Modeling}
\bvolume{61}(\bissue{6}),
\bfpage{2648}--\blpage{2657}
(\byear{2021})
\doiurl{10.1021/acs.jcim.1c00208}
\end{barticle}
\endbibitem

%%% 44
\bibitem[\protect\citeauthoryear{Johansson et~al.}{2014}]{Johansson2014}
\begin{bchapter}
\bauthor{\bsnm{Johansson}, \binits{U.}},
\bauthor{\bsnm{Bostr{\"{o}}m}, \binits{H.}},
\bauthor{\bsnm{L{\"{o}}fstr{\"{o}}m}, \binits{T.}},
\bauthor{\bsnm{Linusson}, \binits{H.}}:
\bctitle{{Regression Conformal Prediction with Random Forests}}.
In: \bbtitle{Machine Learning},
vol. \bseriesno{97},
pp. \bfpage{155}--\blpage{176}.
\bpublisher{Springer},
\blocation{Cham}
(\byear{2014}).
\doiurl{10.1007/s10994-014-5453-0}
\end{bchapter}
\endbibitem

%%% 45
\bibitem[\protect\citeauthoryear{Angelopoulos et~al.}{2020}]{Angelopoulos2021}
\begin{bchapter}
\bauthor{\bsnm{Angelopoulos}, \binits{A.}},
\bauthor{\bsnm{Bates}, \binits{S.}},
\bauthor{\bsnm{Malik}, \binits{J.}},
\bauthor{\bsnm{Jordan}, \binits{M.I.}}:
\bctitle{{Uncertainty sets for image classifiers using Conformal Prediction}}.
In: \bbtitle{ICLR 2021 - 9th International Conference on Learning Representations}.
\bpublisher{OpenReview},
\blocation{New York}
(\byear{2020}).
\burl{https://openreview.net/forum?id=eNdiU_DbM9}
Accessed 14/10/2024
\end{bchapter}
\endbibitem

%%% 46
\bibitem[\protect\citeauthoryear{Messoudi et~al.}{2020}]{Messoudi2020}
\begin{bchapter}
\bauthor{\bsnm{Messoudi}, \binits{S.}},
\bauthor{\bsnm{Rousseau}, \binits{S.}},
\bauthor{\bsnm{Destercke}, \binits{S.}}:
\bctitle{{Deep Conformal Prediction for robust models}}.
In: \bbtitle{Communications in Computer and Information Science},
vol. \bseriesno{1237 CCIS},
pp. \bfpage{528}--\blpage{540}.
\bpublisher{Springer},
\blocation{Cham}
(\byear{2020}).
\doiurl{10.1007/978-3-030-50146-4_39}
\end{bchapter}
\endbibitem

%%% 47
\bibitem[\protect\citeauthoryear{Zhan et~al.}{2020}]{Zhan2020}
\begin{barticle}
\bauthor{\bsnm{Zhan}, \binits{X.}},
\bauthor{\bsnm{Wang}, \binits{Z.}},
\bauthor{\bsnm{Yang}, \binits{M.}},
\bauthor{\bsnm{Luo}, \binits{Z.}},
\bauthor{\bsnm{Wang}, \binits{Y.}},
\bauthor{\bsnm{Li}, \binits{G.}}:
\batitle{{An electronic nose-based assistive diagnostic -prototype for lung cancer detection with Conformal Prediction}}.
\bjtitle{Measurement}
\bvolume{158},
\bfpage{107588}
(\byear{2020})
\doiurl{10.1016/j.measurement.2020.107588}
\end{barticle}
\endbibitem

%%% 48
\bibitem[\protect\citeauthoryear{Papadopoulos}{2008}]{Papadopoulos08}
\begin{bchapter}
\bauthor{\bsnm{Papadopoulos}, \binits{H.}}:
\bctitle{{Inductive Conformal Prediction: Theory and application to Neural Networks}}.
In: \beditor{\bsnm{Fritzsche}, \binits{P.}} (ed.)
\bbtitle{Tools in Artificial Intelligence}.
\bpublisher{IntechOpen},
\blocation{Rijeka}
(\byear{2008}).
\bcomment{Chap. 18}.
\doiurl{10.5772/6078}
\end{bchapter}
\endbibitem

%%% 49
\bibitem[\protect\citeauthoryear{Ashby et~al.}{2022}]{Ashby2022a}
\begin{bchapter}
\bauthor{\bsnm{Ashby}, \binits{A.E.}},
\bauthor{\bsnm{Meister}, \binits{J.A.}},
\bauthor{\bsnm{Nguyen}, \binits{K.A.}},
\bauthor{\bsnm{Luo}, \binits{Z.}},
\bauthor{\bsnm{Gentzke}, \binits{W.}}:
\bctitle{{Cough-based COVID-19 detection with audio quality clustering and confidence measure based learning}}.
In: \beditor{\bsnm{Johansson}, \binits{U.}},
\beditor{\bsnm{Bostr{\"{o}}m}, \binits{H.}},
\beditor{\bsnm{Nguyen}, \binits{K.A.}},
\beditor{\bsnm{Luo}, \binits{Z.}},
\beditor{\bsnm{Carlsson}, \binits{L.}} (eds.)
\bbtitle{Proceedings of Machine Learning Research},
vol. \bseriesno{179},
pp. \bfpage{129}--\blpage{148}.
\bpublisher{PMLR},
\blocation{Norfolk}
(\byear{2022}).
\burl{https://proceedings.mlr.press/v179/ashby22a.html}
Accessed 14/10/2024
\end{bchapter}
\endbibitem

%%% 50
\bibitem[\protect\citeauthoryear{L{\"{o}}fstr{\"{o}}m et~al.}{2015}]{Lofstrom2015}
\begin{botherref}
\oauthor{\bsnm{L{\"{o}}fstr{\"{o}}m}, \binits{T.}},
\oauthor{\bsnm{Bostr{\"{o}}m}, \binits{H.}},
\oauthor{\bsnm{Linusson}, \binits{H.}},
\oauthor{\bsnm{Johansson}, \binits{U.}}:
{Bias reduction through conditional Conformal Prediction}.
Intelligent Data Analysis
\textbf{19}(6)
(2015)
\doiurl{10.3233/IDA-150786}
\end{botherref}
\endbibitem

%%% 51
\bibitem[\protect\citeauthoryear{Meister et~al.}{2023}]{Meister2023}
\begin{barticle}
\bauthor{\bsnm{Meister}, \binits{J.A.}},
\bauthor{\bsnm{Nguyen}, \binits{K.A.}},
\bauthor{\bsnm{Kapetanakis}, \binits{S.}},
\bauthor{\bsnm{Luo}, \binits{Z.}}:
\batitle{{A novel Deep Learning approach for one-step Conformal Prediction approximation}}.
\bjtitle{Annals of Mathematics and Artificial Intelligence}
(\byear{2023})
\doiurl{10.1007/s10472-023-09849-y}
\end{barticle}
\endbibitem

%%% 52
\bibitem[\protect\citeauthoryear{Ndiaye}{2022}]{Ndiaye2022}
\begin{bchapter}
\bauthor{\bsnm{Ndiaye}, \binits{E.}}:
\bctitle{{Stable Conformal Prediction sets}}.
In: \bbtitle{Proceedings of Machine Learning Research},
vol. \bseriesno{162},
pp. \bfpage{16462}--\blpage{16479}.
\bpublisher{PMLR},
\blocation{Norfolk}
(\byear{2022}).
\burl{https://proceedings.mlr.press/v162/ndiaye22a.html}
Accessed 14/10/2024
\end{bchapter}
\endbibitem

%%% 53
\bibitem[\protect\citeauthoryear{Meister}{2020}]{Meister2020}
\begin{botherref}
\oauthor{\bsnm{Meister}, \binits{J.A.}}:
{Conformal Predictors for detecting harmful respiratory events}.
Msc dissertation,
Royal Holloway, University of London
(2020).
\doiurl{10.13140/RG.2.2.28575.02728/1}
\end{botherref}
\endbibitem

%%% 54
\bibitem[\protect\citeauthoryear{Sesia and Romano}{2021}]{Sesia2021}
\begin{bchapter}
\bauthor{\bsnm{Sesia}, \binits{M.}},
\bauthor{\bsnm{Romano}, \binits{Y.}}:
\bctitle{{Conformal Prediction using conditional histograms}}.
In: \bbtitle{Advances in Neural Information Processing Systems},
vol. \bseriesno{34},
pp. \bfpage{6304}--\blpage{6315}.
\bpublisher{Curran Associates, Inc.},
\blocation{New York}
(\byear{2021}).
\burl{https://proceedings.neurips.cc/paper_files/paper/2021/hash/31b3b31a1c2f8a370206f111127c0dbd-Abstract.html}
Accessed 14/10/2024
\end{bchapter}
\endbibitem

%%% 55
\bibitem[\protect\citeauthoryear{Vovk et~al.}{2016}]{Vovk2016}
\begin{bchapter}
\bauthor{\bsnm{Vovk}, \binits{V.}},
\bauthor{\bsnm{Fedorova}, \binits{V.}},
\bauthor{\bsnm{Nouretdinov}, \binits{I.}},
\bauthor{\bsnm{Gammerman}, \binits{A.}}:
\bctitle{{Criteria of efficiency for Conformal Prediction}}.
In: \bbtitle{Lecture Notes in Computer Science},
vol. \bseriesno{9653},
pp. \bfpage{23}--\blpage{39}.
\bpublisher{Springer},
\blocation{Cham}
(\byear{2016}).
\doiurl{10.1007/978-3-319-33395-3_2}
\end{bchapter}
\endbibitem

%%% 56
\bibitem[\protect\citeauthoryear{Nouretdinov et~al.}{2020}]{Nouretdinov2020}
\begin{barticle}
\bauthor{\bsnm{Nouretdinov}, \binits{I.}},
\bauthor{\bsnm{Gammerman}, \binits{J.}},
\bauthor{\bsnm{Fontana}, \binits{M.}},
\bauthor{\bsnm{Rehal}, \binits{D.}}:
\batitle{{Multi-level Conformal Clustering: A distribution-free technique for clustering and anomaly detection}}.
\bjtitle{Neurocomputing}
\bvolume{397},
\bfpage{279}--\blpage{291}
(\byear{2020})
\doiurl{10.1016/j.neucom.2019.07.114}
\end{barticle}
\endbibitem

%%% 57
\bibitem[\protect\citeauthoryear{Jung et~al.}{2021}]{Jung2021}
\begin{barticle}
\bauthor{\bsnm{Jung}, \binits{S.}},
\bauthor{\bsnm{Park}, \binits{K.}},
\bauthor{\bsnm{Kim}, \binits{B.}}:
\batitle{{Clustering on the torus by Conformal Prediction}}.
\bjtitle{The Annals of Applied Statistics}
\bvolume{15}(\bissue{4}),
\bfpage{1583}--\blpage{1603}
(\byear{2021})
\doiurl{10.1214/21-AOAS1459}
\end{barticle}
\endbibitem

%%% 58
\bibitem[\protect\citeauthoryear{Ding et~al.}{2023}]{Ding2023}
\begin{bchapter}
\bauthor{\bsnm{Ding}, \binits{T.}},
\bauthor{\bsnm{Angelopoulos}, \binits{A.N.}},
\bauthor{\bsnm{Bates}, \binits{S.}},
\bauthor{\bsnm{Jordan}, \binits{M.I.}},
\bauthor{\bsnm{Tibshirani}, \binits{R.J.}}:
\bctitle{Class-conditional {Conformal Prediction} with many classes}.
In: \bbtitle{Proceedings of the 36th Advances in Neural Information Processing Systems Conference (NeurIPS)}.
\bpublisher{Curran Associates, Inc.},
\blocation{New York}
(\byear{2023}).
\burl{https://papers.nips.cc/paper_files/paper/2023/hash/cb931eddd563f8d473c355518ce8601c-Abstract-Conference.html}
Accessed 14/10/2024
\end{bchapter}
\endbibitem

%%% 59
\bibitem[\protect\citeauthoryear{Papadopoulos et~al.}{2007}]{Papadopoulos2007}
\begin{bchapter}
\bauthor{\bsnm{Papadopoulos}, \binits{H.}},
\bauthor{\bsnm{Vovk}, \binits{V.}},
\bauthor{\bsnm{Gammerman}, \binits{A.}}:
\bctitle{{Conformal Prediction with Neural Networks}}.
In: \bbtitle{19th IEEE International Conference on Tools with Artificial Intelligence (ICTAI 2007)},
vol. \bseriesno{2},
pp. \bfpage{388}--\blpage{395}.
\bpublisher{IEEE},
\blocation{New York}
(\byear{2007}).
\doiurl{10.1109/ICTAI.2007.47}
\end{bchapter}
\endbibitem

%%% 60
\bibitem[\protect\citeauthoryear{Renkema et~al.}{2024}]{Renkema2024}
\begin{botherref}
\oauthor{\bsnm{Renkema}, \binits{Y.}},
\oauthor{\bsnm{Visser}, \binits{L.}},
\oauthor{\bsnm{AlSkaif}, \binits{T.}}:
{Enhancing the reliability of probabilistic PV power forecasts using Conformal Prediction}.
Solar Energy Advances
\textbf{4}
(2024)
\doiurl{10.1016/j.seja.2024.100059}
\end{botherref}
\endbibitem

%%% 61
\bibitem[\protect\citeauthoryear{Hern{\'{a}}ndez-Hern{\'{a}}ndez et~al.}{2022}]{Hernandez2022}
\begin{bchapter}
\bauthor{\bsnm{Hern{\'{a}}ndez-Hern{\'{a}}ndez}, \binits{S.}},
\bauthor{\bsnm{Vishwakarma}, \binits{S.}},
\bauthor{\bsnm{Ballester}, \binits{P.J.}}:
\bctitle{{Conformal Prediction of small-molecule drug resistance in cancer cell lines}}.
In: \bbtitle{Proceedings of the 11th Symposium on Conformal and Probabilistic Prediction with Applications (COPA)}.
\bsertitle{Proceedings of Machine Learning Research},
vol. \bseriesno{179}.
\bpublisher{PMLR},
\blocation{Norfolk}
(\byear{2022}).
\burl{https://proceedings.mlr.press/v179/hernandez-hernandez22a.html}
Accessed 14/10/2024
\end{bchapter}
\endbibitem

%%% 62
\bibitem[\protect\citeauthoryear{Toccaceli and Gammerman}{2019}]{toccaceli2019combination}
\begin{barticle}
\bauthor{\bsnm{Toccaceli}, \binits{P.}},
\bauthor{\bsnm{Gammerman}, \binits{A.}}:
\batitle{{Combination of inductive Mondrian Conformal Predictors}}.
\bjtitle{Machine Learning}
\bvolume{108}(\bissue{3}),
\bfpage{489}--\blpage{510}
(\byear{2019})
\doiurl{10.1007/s10994-018-5754-9}
\end{barticle}
\endbibitem

%%% 63
\bibitem[\protect\citeauthoryear{Johansson et~al.}{2017}]{johansson2017model}
\begin{bchapter}
\bauthor{\bsnm{Johansson}, \binits{U.}},
\bauthor{\bsnm{Linusson}, \binits{H.}},
\bauthor{\bsnm{Lofstrom}, \binits{T.}},
\bauthor{\bsnm{Bostr{\"{o}}m}, \binits{H.}}:
\bctitle{{Model-agnostic nonconformity functions for conformal classification}}.
In: \bbtitle{Proceedings of the International Joint Conference on Neural Networks (IJCNN)}.
\bpublisher{IEEE},
\blocation{New York}
(\byear{2017}).
\doiurl{10.1109/IJCNN.2017.7966105}
\end{bchapter}
\endbibitem

%%% 64
\bibitem[\protect\citeauthoryear{Norinder et~al.}{2021}]{Norinder2021}
\begin{barticle}
\bauthor{\bsnm{Norinder}, \binits{U.}},
\bauthor{\bsnm{Spjuth}, \binits{O.}},
\bauthor{\bsnm{Svensson}, \binits{F.}}:
\batitle{{Synergy Conformal Prediction applied to large-scale bioactivity datasets and in federated learning}}.
\bjtitle{Journal of Cheminformatics}
\bvolume{13}(\bissue{1}),
\bfpage{77}
(\byear{2021})
\doiurl{10.1186/s13321-021-00555-7}
\end{barticle}
\endbibitem

%%% 65
\bibitem[\protect\citeauthoryear{Campagner et~al.}{2024}]{Campagner2024}
\begin{botherref}
\oauthor{\bsnm{Campagner}, \binits{A.}},
\oauthor{\bsnm{Barandas}, \binits{M.}},
\oauthor{\bsnm{Folgado}, \binits{D.}},
\oauthor{\bsnm{Gamboa}, \binits{H.}},
\oauthor{\bsnm{Cabitza}, \binits{F.}}:
{Ensemble predictors: Possibilistic combination of Conformal Predictors for multivariate time series classification}.
IEEE Transactions on Pattern Analysis and Machine Intelligence,
1--12
(2024)
\doiurl{10.1109/TPAMI.2024.3388097}
\end{botherref}
\endbibitem

%%% 66
\bibitem[\protect\citeauthoryear{Rainio et~al.}{2024}]{Rainio2024}
\begin{botherref}
\oauthor{\bsnm{Rainio}, \binits{O.}},
\oauthor{\bsnm{Teuho}, \binits{J.}},
\oauthor{\bsnm{Kl{\'{e}}n}, \binits{R.}}:
{Evaluation metrics and statistical tests for Machine Learning}.
Scientific Reports
\textbf{14}(1)
(2024)
\doiurl{10.1038/s41598-024-56706-x}
\end{botherref}
\endbibitem

%%% 67
\bibitem[\protect\citeauthoryear{Salazar et~al.}{2023}]{Salazar2023}
\begin{botherref}
\oauthor{\bsnm{Salazar}, \binits{A.}},
\oauthor{\bsnm{Vergara}, \binits{L.}},
\oauthor{\bsnm{Vidal}, \binits{E.}}:
{A proxy learning curve for the Bayes classifier}.
Pattern Recognition
\textbf{136}
(2023)
\doiurl{10.1016/j.patcog.2022.109240}
\end{botherref}
\endbibitem

%%% 68
\bibitem[\protect\citeauthoryear{McInnes et~al.}{2018}]{McInnes2018}
\begin{barticle}
\bauthor{\bsnm{McInnes}, \binits{L.}},
\bauthor{\bsnm{Healy}, \binits{J.}},
\bauthor{\bsnm{Saul}, \binits{N.}},
\bauthor{\bsnm{Gro{\ss}berger}, \binits{L.}}:
\batitle{{UMAP: Uniform manifold approximation and projection}}.
\bjtitle{Journal of Open Source Software}
\bvolume{3}(\bissue{29}),
\bfpage{861}
(\byear{2018})
\doiurl{10.21105/joss.00861}
\end{barticle}
\endbibitem

%%% 69
\bibitem[\protect\citeauthoryear{LeCun et~al.}{1998}]{Lecun1998}
\begin{barticle}
\bauthor{\bsnm{LeCun}, \binits{Y.}},
\bauthor{\bsnm{Bottou}, \binits{L.}},
\bauthor{\bsnm{Bengio}, \binits{Y.}},
\bauthor{\bsnm{Haffner}, \binits{P.}}:
\batitle{{Gradient-based learning applied to document recognition}}.
\bjtitle{Proceedings of the IEEE}
\bvolume{86}(\bissue{11}),
\bfpage{2278}--\blpage{2323}
(\byear{1998})
\doiurl{10.1109/5.726791}
\end{barticle}
\endbibitem

%%% 70
\bibitem[\protect\citeauthoryear{Lincoff}{1983}]{Knopf1981}
\begin{barticle}
\bauthor{\bsnm{Lincoff}, \binits{G.H.}}:
\batitle{{The Audubon Society field guide to North American mushrooms}}.
\bjtitle{Mycologia}
\bvolume{75}(\bissue{3}),
\bfpage{574}
(\byear{1983})
\doiurl{10.2307/3792705}
\end{barticle}
\endbibitem

%%% 71
\bibitem[\protect\citeauthoryear{Cortez et~al.}{2009}]{Cortez2009}
\begin{barticle}
\bauthor{\bsnm{Cortez}, \binits{P.}},
\bauthor{\bsnm{Cerdeira}, \binits{A.}},
\bauthor{\bsnm{Almeida}, \binits{F.}},
\bauthor{\bsnm{Matos}, \binits{T.}},
\bauthor{\bsnm{Reis}, \binits{J.}}:
\batitle{{Modeling wine preferences by data mining from physicochemical properties}}.
\bjtitle{Decision Support Systems}
\bvolume{47}(\bissue{4}),
\bfpage{547}--\blpage{553}
(\byear{2009})
\doiurl{10.1016/j.dss.2009.05.016}
\end{barticle}
\endbibitem

%%% 72
\bibitem[\protect\citeauthoryear{Hull}{1994}]{Hull1994}
\begin{barticle}
\bauthor{\bsnm{Hull}, \binits{J.J.}}:
\batitle{{A database for handwritten text recognition research}}.
\bjtitle{IEEE Transactions on Pattern Analysis and Machine Intelligence}
\bvolume{16}(\bissue{5}),
\bfpage{550}--\blpage{554}
(\byear{1994})
\doiurl{10.1109/34.291440}
\end{barticle}
\endbibitem

%%% 73
\bibitem[\protect\citeauthoryear{LeCun et~al.}{2021}]{lecun2021mnistDataset}
\begin{botherref}
\oauthor{\bsnm{LeCun}, \binits{Y.}},
\oauthor{\bsnm{Bottou}, \binits{L.}},
\oauthor{\bsnm{Bengio}, \binits{Y.}},
\oauthor{\bsnm{Haffner}, \binits{P.}}:
{MNIST database of handwritten digits}
(2021).
\doiurl{10.24432/C53K8Q}
\end{botherref}
\endbibitem

%%% 74
\bibitem[\protect\citeauthoryear{Schlimmer}{1987}]{schlimmer1987mushroomDataset}
\begin{botherref}
\oauthor{\bsnm{Schlimmer}, \binits{J.}}:
{Mushroom dataset}
(1987).
\doiurl{10.24432/C5959T}
\end{botherref}
\endbibitem

%%% 75
\bibitem[\protect\citeauthoryear{Cortez et~al.}{2009}]{Cortez2009WineDataset}
\begin{botherref}
\oauthor{\bsnm{Cortez}, \binits{P.}},
\oauthor{\bsnm{Cerdeira}, \binits{A.}},
\oauthor{\bsnm{Almeida}, \binits{F.}},
\oauthor{\bsnm{Matos}, \binits{T.}},
\oauthor{\bsnm{Reis}, \binits{J.}}:
{Wine quality dataset}
(2009).
\doiurl{10.24432/C56S3T}
\end{botherref}
\endbibitem

%%% 76
\bibitem[\protect\citeauthoryear{Hull}{1994}]{hull1994uspsDataset}
\begin{botherref}
\oauthor{\bsnm{Hull}, \binits{J.J.}}:
{USPS dataset}
(1994).
\url{https://www.kaggle.com/datasets/bistaumanga/usps-dataset}
Accessed 2023-11-08
\end{botherref}
\endbibitem

\end{thebibliography}

\end{document}